\def\year{2022}\relax
\documentclass[letterpaper]{article} 
\usepackage{aaai22}  
\usepackage{times}  
\usepackage{helvet}  
\usepackage{courier}  
\usepackage[hyphens]{url}  
\usepackage{graphicx} 
\usepackage{natbib}  
\usepackage{caption} 
\DeclareCaptionStyle{ruled}{labelfont=normalfont,labelsep=colon,strut=off} 
\usepackage{algorithm}
\usepackage{algorithmic}

\usepackage{newfloat}
\usepackage{listings}
\floatstyle{ruled}
\newfloat{listing}{tb}{lst}{}
\floatname{listing}{Listing}
\usepackage{xcolor}
\usepackage{subcaption}
\usepackage{booktabs}
\usepackage{multirow,}
\usepackage{siunitx}
\usepackage{amsmath}
\usepackage{amssymb}
\usepackage{svg}

\title{Searching for Structure in Unfalsifiable Claims}

\author {
    Peter Ebert Christensen,\textsuperscript{\rm 1}
    Frederik Warburg,\textsuperscript{\rm 2}
    Menglin Jia,\textsuperscript{\rm 3}
    Serge Belongie\textsuperscript{\rm 1} \\
}
\affiliations {
    \textsuperscript{\rm 1} University of Copenhagen \& Pioneer Centre for AI \\
    \textsuperscript{\rm 2} Technical University of Denmark \\
    \textsuperscript{\rm 3} Cornell University \\
    pec@di.ku.dk, frwa@dtu.dk, mj493@cornell.edu, s.belongie@di.ku.dk
}

\begin{document}

\maketitle

\begin{abstract}

Social media platforms give rise to an abundance of posts and comments on every topic imaginable. Many of these posts express opinions on various aspects of society, but their unfalsifiable nature makes them ill-suited to fact-checking pipelines. 
In this work, we aim to distill such posts into a small set of narratives that capture the essential claims related to a given topic. Understanding and visualizing these narratives can facilitate more informed debates on social media.
As a first step towards systematically identifying the underlying narratives on social media, we introduce \textsc{papyer}\includegraphics[height=1em]{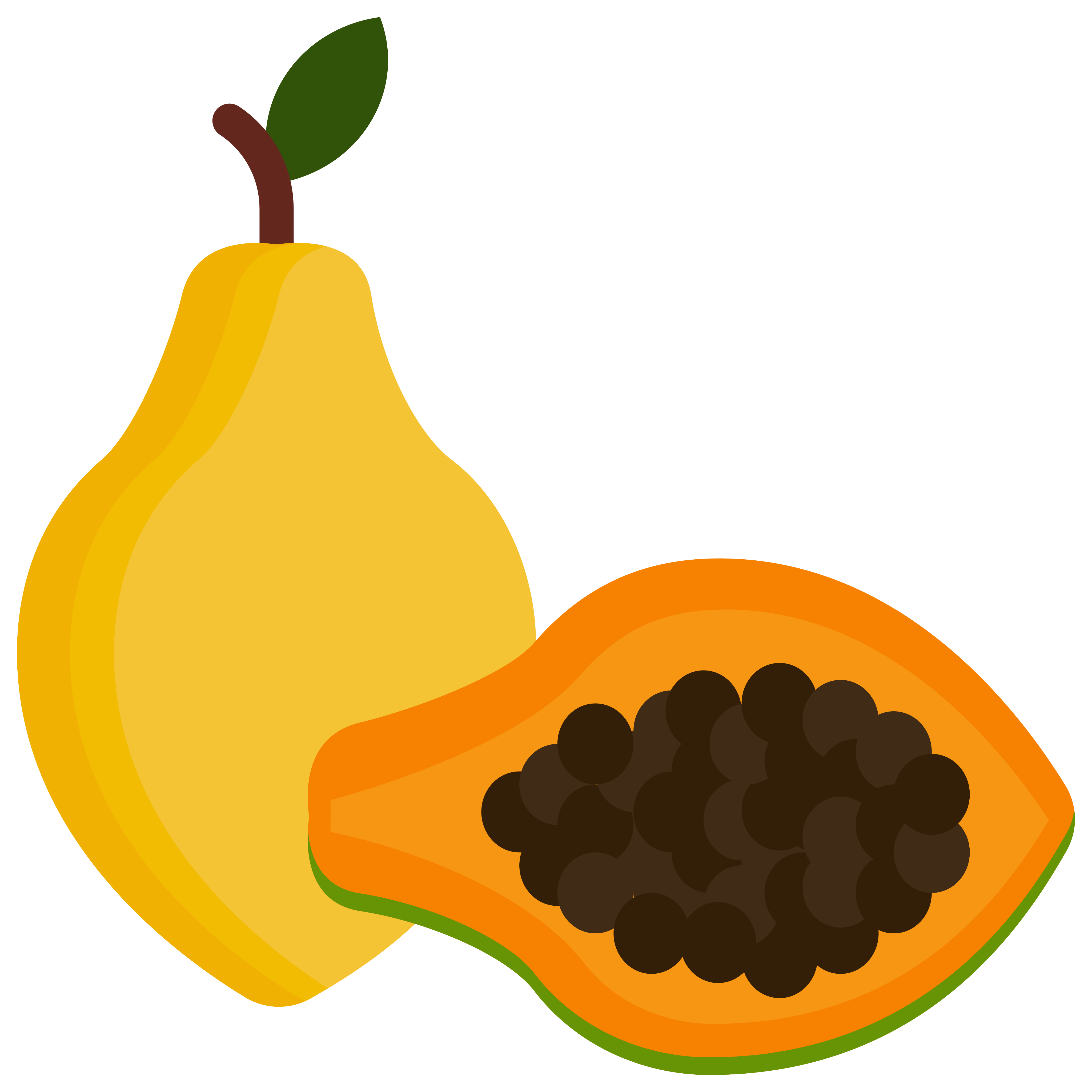}, a fine-grained dataset of online comments related to hygiene in public restrooms, which contains a multitude of unfalsifiable claims. We present a human-in-the-loop pipeline that uses a combination of machine and human kernels to discover the prevailing narratives and show that this pipeline outperforms recent large transformer models and state-of-the-art unsupervised topic models. 

\end{abstract}

\section{Introduction}

Social media platforms have changed the ways information is produced, disseminated, and consumed, creating new opportunities along with complex challenges. One of these challenges is how to grasp, use, and interpret a large corpus of text from online discussion. 

Several works~ \cite{LDA,topicmodellingevolution,mimno-thompson-lda,Moody2016MixingDT,sia-etal-2020-tired} aim to distill large documents either through topic modeling or document summarisation. 
Our work falls into this category, however, we focus on identifying narratives in fine-grained topic-specific discussions.

\begin{figure}[t]
    \centering
    \includegraphics[width=\linewidth]{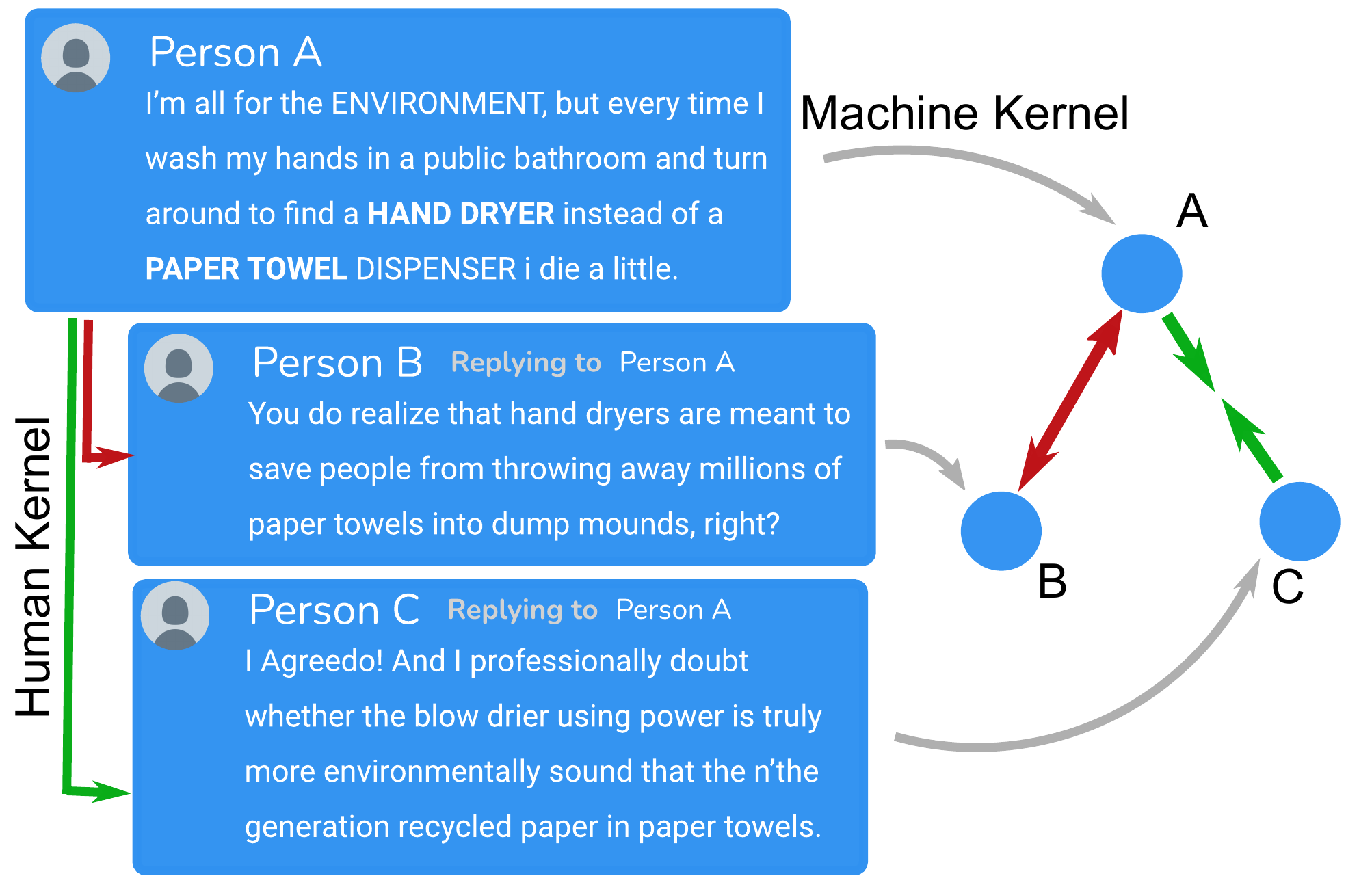}%
    \caption{
        Using a combination of machine and human kernels, we strive to discover the underlying narratives in online discussions. Our human kernel acts on triplets, posing the question of who would likely take the same side in a debate. In this example, while person $A$ and $C$ express different ideas, they both prefer paper towels over air dryers, making them better suited to team with one another rather than person $B$. These claims illustrate the complexity of the task at hand that requires an understanding of content, context, slang, humor and sarcasm.  
        In this work, we demonstrate that human annotated triplets enable us to learn a low-dimensional representation of such claims, which reveals clusters of narratives from online discussions. 
    }
    \label{fig:narrative_illustration}
\end{figure}

Our use of the term \emph{narrative} follows the vulgar sense found in arguments in social media, as opposed to the literary sense that refers to connected events in a story involving a protagonist, villain, transformation, etc. The former sense
frequently appears in accusations of the form \emph{you don't mention X because it doesn't fit narrative Y,} where $X$ is a check-worthy claim and $Y$ is an unfalsifiable claim. Consider the following tweet: 
\begin{itemize}
\item[] 
\texttt{The FBI wants to push the narrative that white nationalism is the biggest domestic threat we face today. Here's the problem: the facts don't fit that narrative.}
\end{itemize}
Whether the FBI would agree that they espouse the above narrative is not within the scope of our current work. We focus instead on the following problems, which we pursue in a human-in-the-loop framework: (1) inferring  narratives from comments and (2) computing distances between text excerpts with respect to narrative alignment.

With such a distance function in hand, one can obtain a valuable signal that is more fine-grained compared to traditional topic modeling, illuminating cases in which differently-worded content resolves to the same narrative. 
In organic discussions on social media, such narrative-based analysis could facilitate more informed debate, in the sense that participants can marshal facts more efficiently to support a compact set of distilled, indexed narratives. In less organic settings, as in cases of astroturfing or sock-puppet infiltration, we envision that such narrative based analysis could point to the existence of latent, manufactured talking points. Furthermore, such narrative-based analysis can complement and extend current fact-checking pipelines, which only consider falsifiable claims.

There are a number of reasons why it is difficult to label narratives present in tweets and other social media comments. $(1)$ Doing so requires insight into the topic, for which the set of potentially relevant narratives is, in practice, not known beforehand. $(2)$ The appropriate level of label granularity is not obvious. $(3)$ The number of labels per topic varies. We, therefore, propose to cast narrative discovery as a triplet-based metric learning problem. This allows us to ask annotators whether comment $A$ belongs with comment $B$ or $C$ without depending on ground truth class annotations, simplifying the annotation process.

Directly presenting annotators with randomly sampled triplets, however, is impractical in terms of the human effort required. In the computer vision literature, SNaCK~\cite{Wilber2015LearningCE} offers an effective means of combining visual similarity with triplet constraints to increase the information gain per new annotation. In that work, Wilber et al.~applied SNaCK to a set of food images, presenting human annotators with Human Intelligence Tasks (HITs) tapping into their perception of the taste of the depicted food, ultimately producing a low-dimensional embedding of the meals as points in flavor space, without appealing to ground truth labels of cuisine types. In the present work, we adapt this method for the annotation of text, with \textit{narrative} used in place of \textit{taste}. \\

In our experiments, we focus on a discussion topic related to hygiene and present \textsc{papyer}\includegraphics[height=1em]{images/papaya.png}, a dataset contains narratives related to the use of hand drying in public restrooms (i.e., \textit{pap}er vs.~air dr\textit{yer}). We select this topic as it (a) gives rise to vigorous discussions in social media, (b) is widely relatable, (c) is manageable in scope, and (d) possesses elements analogous to a variety of other domains involving human decision making.

Our main contributions are as follows: (1) the introduction of a new human-in-the-loop machine learning problem of social media narrative discovery, (2) a workflow for narrative annotation based on SNaCK, and (3) a dataset for quantitative narrative analysis.

\section{Related Work}

We first review topic modeling and fact-checking, as our work can be considered an instance of fine-grained topic modeling adjacent to conventional fact-checking workflows. We examine connections to other approaches such as document summarization, online discourse, and sentiment analysis.
Lastly, we review the relations between our workflow, crowd kernel learning, and  human-in-the-loop annotation for modeling abstract narrative similarity.

\textbf{Topic modeling.} 
aims to discover groups of words corresponding to subcategories in a collection of documents in an unsupervised manner.
The best-known approach is Latent Dirichlet Allocation (LDA)~\cite{LDA}, which uses Dirichlet priors to characterize distributions of topics and words. 
Since the introduction of word2vec \cite{w2v} newer models have incorporated word embeddings to help reduce the sparsity of the word co-occurrence space \cite{Moody2016MixingDT, gui-etal-2019-neural,  Arruda2016TopicSV}. This has enabled models to handle data from a variety of sources including tweets, scientific articles, short texts, and books \cite{topicmodellingevolution}. However, we find that these methods tend to not discover the prevailing narratives, as this task requires an understanding of context, humor, sarcasm, etc. \\

 \textbf{Fact checking.} 
 One of our long-term motivations to discover narratives is to complement fact-checking pipelines that currently do not have a use for unfalsifiable claims.
 
 Concretely, a statement such as “\textit{Queen Elizabeth II was born in 1926.}” is considered falsifiable and hence check-worthy \cite{jaradat-etal-2018-claimrank, conf/ranlp/GenchevaNMBK17,Hassan}, while “\textit{The royal family is a waste of taxpayers' money.}” is an unfalsifiable claim. A common fact-checking pipeline would discard the latter type of claims being not check-worthy nor easily verifiable \cite{augenstein2021explainable}. In case of a check-worthy claim, it then proceeds to retrieve evidence from documents drawn from a database curated by annotators. The claims under consideration could be numerical properties, quotes, or event participation.  Our approach is not applied in any fact-checking pipeline, but can be seen as complementary, as annotators no longer need to consider the veracity of claims, and instead only think about underlying similarity in terms of views held by the individuals making the claims. 
 In doing so, we aim to discover a summary of claims that can subsequently describe all facets of a debate.
 This brings us to our next subarea of related work. \\ 
 
\textbf{Document summarization.} Discovering the prevailing narratives in a text corpus can also be viewed as an instance of document summarization. Document summarization has been studied extensively in two major ways; one that rearranges the content of the document to produce a summary (extractive) and another that generates a summary given context (abstractive) \cite{carenini-cheung-2008-extractive}.
In the latter, a summary is provided as a target, while in the former, one selects sentences directly using scoring functions such as the Jaccard distance between the sentence and intermediate key phrases \cite{jadon-pareek-2016-method}.
Recently, \citet{tan-etal-2020-summarizing} made it possible to generate abstractive summaries using any aspect, such as ``sports'' or ``health.'' The method allows for fine-grained controllability of the text generation by incorporating knowledge through weak supervision using ConceptNet \cite{conceptnet} or BART \cite{lewis-etal-2020-bart}. 
Despite not directly using any summarization methods, our approach can be seen as attempting to find sentences that would cluster around an abstractive summary, i.e., a latent narrative that shares many similarities with statements in online discourse. \\

\textbf{Online discourse analysis.} Our work is closely intertwined with mediated narrative analysis, which explains how characters share stories on social media and how tellers position themselves compared to the narratives \cite{narrativeonline}, be it through hashtags or other means \cite{hashtag}. There has been remarkable progress in mapping out different types of storytelling on different social media platforms \cite{progress_discourse}. 
Additionally, what is considered acceptable social behavior drives storytelling and hence the narratives surrounding the story \cite{forbes-etal-2020-social}. 
This is complementary to our work, as we only focus on narratives that have gone viral, not finding the causes behind surpassing a certain virality threshold, how they were shared, or the affective state in which people might view them. Instead, our proposed approach is conditioned on having a sufficiently large body of comments to investigate for narratives. \\

\textbf{Sentiment analysis.} Narratives often contain subjective unfalsifiable information coming from social media and one could therefore study it with sentiment analysis. Sentiment analysis, which shares similarities with stance detection \cite{ALDAYEL2021102597}, has been applied to a variety of topics, but relies on a discrete classification of sentiments \cite{li-caragea-2019-multi}. An important application of sentiment analysis is hate speech detection, which can be considered a fine-grained category of the former, as it can be incorporated as an auxiliary classification task \cite{schmidt-wiegand-2017-survey}.
In this work, we do not classify the sentiment of claims, but let humans decide
how sentences align in an online debate. Thus, the sentiment is decoupled as sentences with different sentiments can belong to the same narrative. \\

\textbf{Human-in-the-loop} approaches
leverage both machine and human intelligence in an AI pipeline. For example, \citet{Perona2010VisionOA} started the Visipedia project to integrate human visual knowledge into a searchable and organized format, initially as a GUI for annotating images, helping people capture and share visual expertise. Multiple works leveraging human knowledge through crowdsourcing have since appeared in the same or other formats \cite{Jia2021ExploringVE, Wilber2015LearningCE, Horn2018b, Jia2021IntentonomyAD, Branson2010VisualRW}. 
Related initiatives in arts and entertainment include TV-tropes \cite{tvt} and the Periodic Table of Storytelling \cite{p_table}, which enable community members both to submit and query narratives in modern pop culture.

According to \citet{MILLER20191}, people are interested in contrastive explanations -- why $X$ instead of $Y$? -- and selective explanations; only the most important information for decision making is shown. Accordingly, we base our crowdwork annotation interface on triplet-based relative preferences.
Our long-term vision is analogous to Visipedia: we wish to capture and share human narratives in online discussions across a wide array of topics. The present work represents our first foray in this direction, with a deep dive into a single topic. Due to the volume of topics and discussions in online forums, our proposed approach must tap into the complementary strengths of humans and machines, as described next. \\

\textbf{Crowd Kernel Learning}
is a strategy for capturing human notions of similarity or dissimilarity that remain elusive to state-of-the-art machine learning based representations. 
For instance, \citet{pmlr-v2-agarwal07a} investigate how humans perceive light from surfaces by presenting annotators image triplets depicting the Stanford Bunny with varying material properties. 
The annotators were asked which bunnies were more similar, revealing a perceptual space for reflectance. Analogously, CKL~\cite{CKL} presents annotators for triplets of necktie images and asked them whether they would purchase $b$ or $c$ if $a$ was sold out. With these triplet annotations, they uncovered a necktie space where nearest neighbors are explicable in terms of glossiness, pattern, and color. 
Similarly, \citet{tste} use \textit{t}-STE (stochastic triplet embedding) to produce a genre embedding for musical artists. \citet{Wilber2015LearningCE} introduced SNaCK with the motivation of capturing the taste-based similarity of food dishes, addressing cases such as guacamole vs.~wasabi that, despite their visual similarity, are far apart in taste space. 
In this paper, we take a similar approach as illustrated in Figure \ref{fig:narrative_illustration}. We provide annotators with triplets of text snippets, and ask the annotators ``who would be on the same side of a debate on this topic?'' to uncover a latent narrative space.


\section{Method}

Assume a collection of $N$ comments extracted from an online discussion. 
Our approach iteratively applies SNaCK \cite{Wilber2015LearningCE} to learn a low dimensional representation $Y \in \mathbb{R}^{N \times d}$ of these comments. We first run SNaCK, and use the obtained embedding to select informative triplets to annotate. We then update the embedding with the newly annotated triplets. We repeat this process till convergence. 
We show that this iterative optimization clusters the underlying narratives, when enough human domain knowledge has been supplied. 

\textbf{Formulation.} The objective of SNaCK is the weighted sum of the \textit{t}-SNE and \textit{t}-STE losses
\begin{equation}
    C_{SNaCK} = \lambda C_{tSNE} + (\gamma)C_{tSTE}.
\end{equation}
The loss function for \textit{t}-SNE is given by 
\begin{equation}
    C_{tSNE} = KL(P || Q) = \sum_{j \neq i} p_{ij}\mathrm{log} \frac{p_{ij}}{q_{ij}},
\end{equation}
Similarly to \citet{Wilber2015LearningCE}, we use a Gaussian kernel $K \in \mathbb{R}^{N \times N}$, such that
\begin{align}
    p_{i j} &=\frac{1}{2 N}\left(p_{j \mid i}+p_{i \mid j}\right) \\
    p_{j \mid i} &=\frac{\exp \left(-K_{i j}^{2} / 2 \sigma_{i}^{2}\right)}{\sum_{k \neq i} \exp \left(-K_{i k}^{2} / 2 \sigma_{i}^{2}\right)}.
\end{align}
This loss function can be interpreted as finding a low-dimensional distribution of points that maximizes the information gain from the original high-dimensional space \cite{tsne}. 
The bandwidth of the Gaussian kernel $\sigma_i$ is set such that the perplexity of the conditional distribution $p_{j \mid i}$ equals a predefined perplexity $u$.

The embedding similarity $ q_{i j}$ between the two points $y_i$ and $y_j$ is computed as a normalized Student's $t$ kernel with a single degree of freedom
\begin{equation}
    q_{i j} =\frac{\left(1+\left\|y_{i}-y_{j}\right\|^{2}\right)^{-1}}{\sum_{k \neq l}\left(1+\left\|y_{k}-y_{l}\right\|^{2}\right)^{-1}}.
\end{equation}

The loss function for t-STE is given by
\begin{equation}
    C_{tSTE}=\sum_{(i, j, k) \in T} \log p_{(i, j, k)}^{tSTE},
\end{equation}
and can be interpreted as the joint probability of independently satisfying all triplet constraints \cite{tste}. We use a Student's $t$ kernel with $\alpha$ degrees of freedom
\begin{equation}
p_{(i, j, k)}^{tSTE}=\frac{\left(1+\frac{\left\|y_{i}-y_{j}\right\|^{2}}{\alpha}\right)^{-\frac{1+\alpha}{2}}}{\left(1+\frac{\left\|y_{i}-y_{j}\right\|^{2}}{\alpha}\right)^{-\frac{1+\alpha}{2}}+\left(1+\frac{\left\|y_{i}-y_{k}\right\|^{2}}{\alpha}\right)^{-\frac{1+\alpha}{2}}}.
\end{equation}
In all experiments, we set $\lambda=0.1$ and $\gamma=5$, which makes the gradient norm of $C_{tSTE}$ and $C_{tSNE}$ equal.


\section{PAPYER}
We construct a new dataset to conduct our narrative analysis.\footnote{http://github.com/captainE/Searching-for-Structure-in-Unfalsifiable-Claims} The dataset focuses on the topic of hand drying in public restrooms. As discussions on this topic largely center on the \textit{pap}er vs.~air dr\textit{yer} debate, we name the dataset \textsc{papyer}\includegraphics[height=1em]{images/papaya.png}.

\begin{figure}
    \centering
    \vspace{-6mm}
    \includegraphics[width=\linewidth]{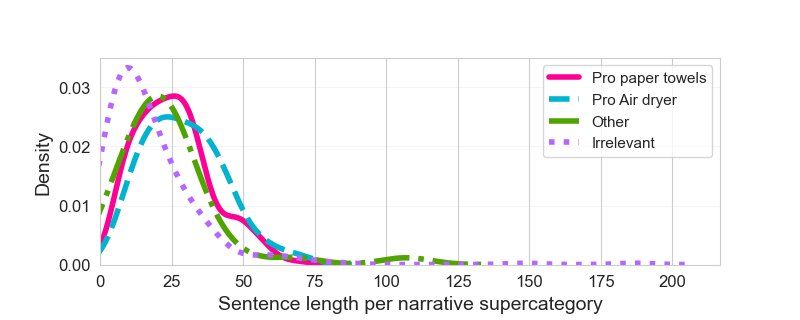}%
    \caption{
        Sentence length distribution by supercategory. 
    }
    
    \label{fig:sent_dist}
\end{figure}

\begin{figure*}[t]
    \centering
    \includegraphics[width=\textwidth]{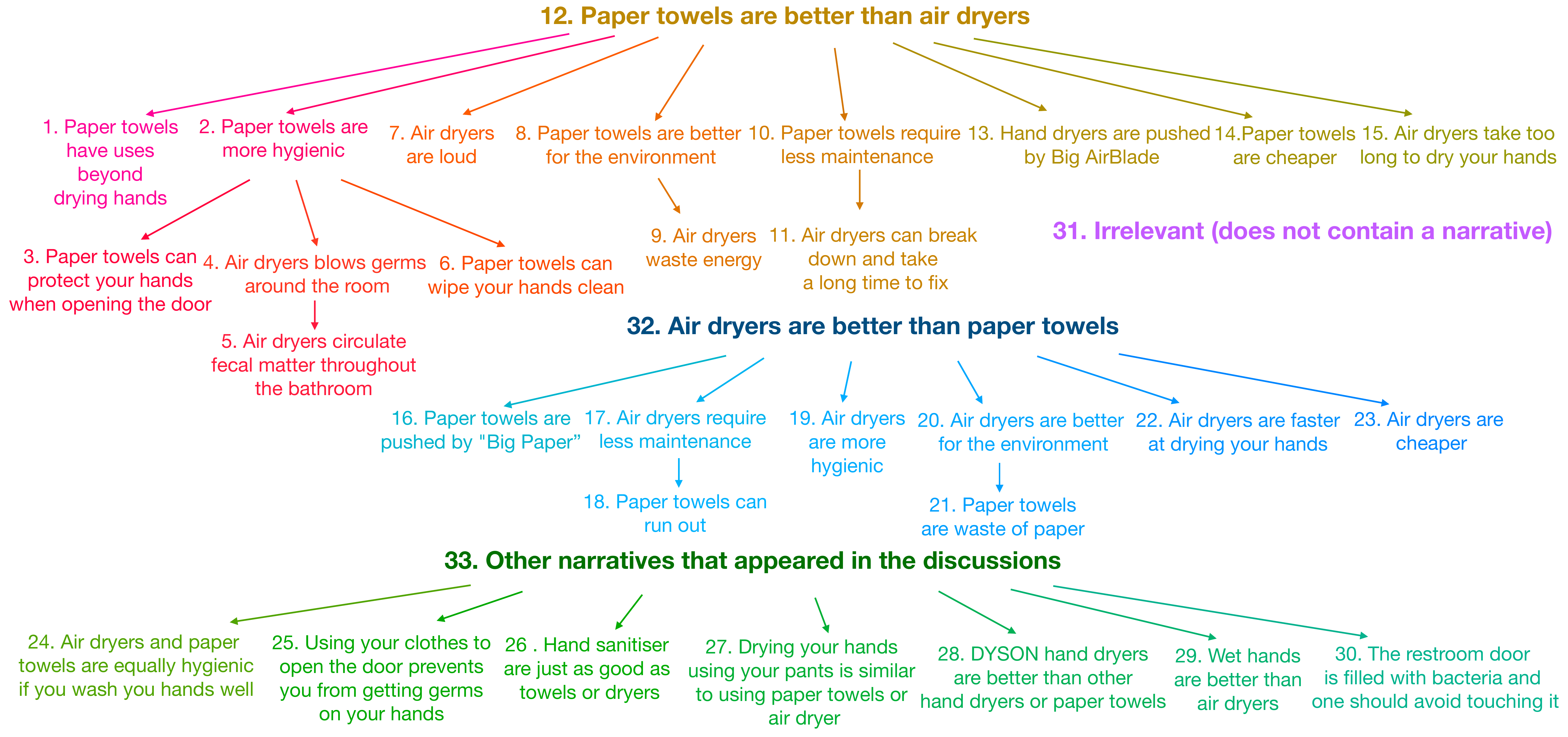}
    \caption{\textbf{Overview of the prevailing narratives.} The narratives are grouped into supercategories: pro-paper towels (in red, orange, yellow), pro-air dryers (in blue), other (green), and irrelevant (in purple). Each sub-narrative has a unique color, which we will use in the rest of the paper. The tree structure highlights the different levels of granularity that may exist within the landscape of narratives.
    }
    \label{fig:latent_narrative}
\end{figure*}

We first scrape Reddit for posts related to hygiene in public restrooms. We manually filter the comments and split them into short text excerpts ($1$-$2$ sentences). Based on these excerpts, we manually define $31$ narratives across $4$ supercategories: $15$ pro-paper towel, $8$ pro-air dryer, $7$ other (related to hand drying), and $1$ for irrelevant (not related to hand drying). We illustrate the narratives in a tree structure in Figure ~\ref{fig:latent_narrative}, which highlights the granularity of narratives in online discussions. Finally, we assign a label to each excerpt by selecting the best match from the list of 31 narratives, which we dub the 31 crystallised narratives.
The dataset consists of $600$ excerpts.
We report the dataset summary statistics, such as Token Type Ratio (TTR) and the number of examples in Table~\ref{tab:quick_text_stats}. Figure~\ref{fig:sent_dist} shows the sentence lengths of the four narrative supercategories.

\begin{table}[ht]
\footnotesize
\center
\begin{tabular}{lllll}
\toprule
            & Pro Paper & Pro Dryer & Other & Irrelevant \\
\midrule
Avg. Length & 26        & 28             & 26      & 18         \\
Word Types  & 1444      & 1019           & 612     & 1959       \\
TTR         & 0.33      & 0.39           & 0.47    & 0.37       \\
Examples    & 169       & 92             & 49      & 290      \\
\bottomrule
\end{tabular}
\caption{Statistics for the \textsc{papyer}\includegraphics[height=1em]{images/papaya.png} dataset. Sentences in across supercategories have roughly equal length, except the irrelevant category, which indicates that length can only be use to distinguish the sentences in the irrelevant category from the rest. The TTR for the Other category is higher than rest of the categories since it has fewer word types. 
}\label{tab:quick_text_stats}
\end{table}

We note that such a manually driven process of collecting and labeling presents practical scaling challenges. It requires access to curators with a complete understanding of all the prevailing narratives before one can begin the annotation process, which is especially challenging because of the multiple levels of granularity and the continuous evolution of narratives in online debates. Neither is it expected that models trained on a specific topic will generalize to narratives for other topics. To use an analogy from our earlier discussion on related work, musical artists continue to produce music that defies existing genre delineations, and the same goes for chefs exploring new culinary directions. 
This means that (1) we cannot simply train a static supervised classifier on these topics, (2) we must strive for more efficient labeling methods that do not require comprehensive knowledge of topic-specific nomenclature, and (3) we need methods that can accommodate multi-level granularity in the data.
That being said, the only manner to evaluate more efficient methods in a quantitative manner  involves going through this tedious, hand-crafted labeling process. We highlight that the purpose of this dataset is to address the above-mentioned scaling problems and therefore only use the labels for evaluation. We believe that the presented dataset is a challenging, real-world example of an online discussion with a number of unfalsifiable claims, and that methods that can efficiently discover the prevailing narratives for this dataset will perform well across other topics. 

\section{Playback Simulation}

We propose to use triplet labeling to improve the efficiency and scalability of collecting and annotating data for narrative discovery. In triplet labeling, an annotator is asked to judge \emph{if text a should be associated with text b or text c}. The main advantages of triplet labeling in our setting are that (1) the annotators are not required to know nor consider all the underlying narratives to label the data, and (2) it organically handles the multi-level granularity as we study similarities rather than class probabilities. This makes data labeling easier and more scalable. To validate if triplet labeling can be used to discover the underlying narratives, we first conduct a playback simulation, that is, we present a computer (a.k.a.,~a synthetic worker) with triplets of text excerpts and use the ground-truth labels to simulate what a human annotator would select as the best match. This playback simulation allows us to explore multiple hyper-parameters before we embark an annotation campaign using Amazon Mechanical Turk (mturk).
We explore three hyper-parameters: (1) sentence embedding network, (2) triplet sampling strategy, and (3) number of positive/negative examples. Lastly, we describe how we simulate human annotators. \\

\textbf{Sentence Embedding Networks.}
Recent, large transformer models understand grammatical and semantic information. We investigate embeddings of several of these models to examine whether they may allow one to discover narratives and which model is most suitable as the machine kernel in SNaCK. More specifically, we explore several sizes of BERT~\cite{devlin-etal-2019-bert}, RoBERTa~\cite{Liu2019RoBERTaAR}, GPT2~\cite{Radford2019LanguageMA}, and T5~\cite{T5}. \\

\textbf{Sampling Strategy.} Randomly presenting five text excerpts per anchor will generally result in choices that are very different from the anchor, thus tasking the annotator to label triplets that will lead to a low information gain. Therefore, we investigate sampling strategies that maximize the information gain per annotation. We explore the following strategies:

\begin{itemize}
    \item \texttt{Random}: Randomly sample $5$ sentences. 
    \item \texttt{Top-k}: Retrieve the $5$ closest text excerpts based on embedding distance for each anchor.
    \item \texttt{Distance}: Retrieve the $20$ closest text excerpts for each anchor, and sample among these sentences with a probability proportional to their distance to the anchor.
    \item \texttt{Distance-Rnd}: The same as Distance except for the last excerpt which is a randomly (Rnd) sampled sentence, not being the anchor nor the $20$ nearest neighbors. 
    \item \texttt{Oracle}: Randomly retrieve $2$ text excerpts from the same narrative as the anchor and randomly select $3$ from different narratives using the ground-truth labels.
\end{itemize}

\textbf{Number of positive and negative examples.}
During training, we force workers to choose $k$ positives (examples most similar to the anchor) from a list of $n$ examples. Earlier studies \cite{Wilber_Kwak_Belongie_2014} have shown the effectiveness of choosing large values of $k$ and $n$ to generate more triplet constraints. 
However, these studies assumed image stimuli, which humans can process efficiently in parallel. Text, on the other hand, is processed sequentially, thus in practice does not enjoy the same scaling properties. Therefore, we limit $n \leq 5$ so as not to overwhelm the annotators with text.\\

\textbf{Simulated Annotations.} 
We simulate human decisions with the following selection procedure. We use the ground-truth labels to select positives that have the same crystallised narrative as the anchor sentence. If the number of sentences with the same narrative as the anchor is not $k$, we select the sentences that are closest in embedding space. 

These synthetic experiments are intended to provide insights into designing our experiment and hyper parameters before collecting human-annotated data with mturk.

\begin{table*}[htp!]
    \centering
        {\fontsize{9pt}{9pt}\selectfont
    \begin{tabular}{lllllll}
        \toprule
        \multirow{2}{*}{\shortstack[*]{}}&
        \multicolumn{1}{c}{} & 
        \multicolumn{2}{c}{Triplet generalization ratio} & \multicolumn{2}{c}{KNN generalization ratio} \\
        & \# Parameters & Embedding & \textit{t}-SNE & Embedding & \textit{t}-SNE \\ \midrule
        T5-base & 220 M & $58.36 \pm 1.58$ & $60.41 \pm 1.93$ & $27.46 \pm 3.62$  & $25.12 \pm 3.79$ \\
        T5-3B &  3 B & \boldmath{$62.48 \pm 1.29$}  & \boldmath{$63.63 \pm 1.37$}  & \boldmath{$33.84 \pm 3.42$} &  \boldmath{$29.84 \pm 3.26$} \\
        T5-11B & 11 B & $62.01 \pm 1.14$  & $62.32 \pm 1.51$  & $33.19 \pm 4.30$ & $29.11 \pm 4.31$ \\
        bert-base & 110 M & $55.89 \pm 1.75$  & $55.18 \pm 1.83$ & $20.82 \pm 3.76$  & $14.42 \pm 3.11$ \\
        bert-large & 130 M & $54.73 \pm 1.76$  & $54.62 \pm 1.58$ & $19.54 \pm 3.18$  & $12.35 \pm 2.52$ \\
        roberta-base & 125 M & $57.39 \pm 1.42$ & $56.47 \pm 1.36$ & $21.69 \pm 3.73$ & $11.14 \pm 3.52$ \\
        roberta-large & 355 M & $59.08 \pm 2.11$ & $58.12 \pm 1.93$ & $24.12 \pm 3.73$ & $14.43 \pm 3.24$ \\
        gpt2-base & 117 M & $53.65 \pm 1.63$  & $53.61 \pm 1.73$ & $12.94 \pm 2.27$  & $9.67 \pm 2.58$ \\
        gpt2-medium & 345 M & $54.28 \pm 1.87$  & $54.31 \pm 1.86$ & $13.75 \pm 2.71$  & $9.54 \pm 2.60$ \\
        gpt2-large & 774 M & $61.56 \pm 1.49$  & $60.38 \pm 1.87$ & $23.32 \pm 2.95$  & $21.80 \pm 3.71$ \\
        \bottomrule
    \end{tabular}
    }
    \caption{\textbf{Embedding network.} The effect of using raw embedding or their \textit{t}-SNE projection for different transformer models. Models are evaluated 10 times using 1 trained model
    (ratios multiplied by 100). Larger transformer models achieve a higher triplet gen. ratio than smaller models and applying \textit{t}-SNE does not change this performance much, except on T5 where it increases. This is not the case for the KNN ratio. 
    We adapt T5-3B as our machine kernel, since it shows the best performance. 
    }
    \label{tab:snack_backbones}
\end{table*}

\begin{table*}[htp!]
    \centering
        {\fontsize{9pt}{9pt}\selectfont
    \begin{tabular}{lllllllll}
        \toprule
          & TGR ($\uparrow$) & KNNGR ($\uparrow$) & SNR ($\downarrow$) & Agreements ($\uparrow$) & Disagreements ($\downarrow$) & Precision ($\uparrow$)& Recall ($\uparrow$)\\ \midrule
        \texttt{Random}             & $75.47 \pm 1.10$ & $16.98 \pm  4.53$ & 2.98 & 11132 & 13110 & 7.62 & 8.53 \\
        \texttt{Distance}          & $68.41 \pm 1.56$ & $31.18 \pm 3.78$ & 0.47 & 16302 & 7998 & 37.81 & 40.85 \\
        \texttt{Top-k}              & $59.86 \pm 1.64$ & $20.00 \pm 4.54$ & 0.68 & 16187 & 8181 & 37.64 & 43.58  \\
        \texttt{Distance-Rnd}   & $77.13 \pm 1.43$ & $40.86 \pm 4.27$ & 0.69 & 15599 & 8703 & 32.13 & 35.35 \\
        \texttt{Oracle}            & $91.71 \pm 1.59$ & $58.17 \pm 2.51$ & 0.17 & 24368 & 0 & 100.0 & 100.0 \\
        \bottomrule
    \end{tabular}
    }
    \caption{\textbf{Sampling strategies.} 
    All models apply the T5-3B embedding network and are evaluated 10 times (Ratios multiplied by 100). 
    Note that the \texttt{Oracle} has access to the ground truth labels. The \texttt{Top-k}, \texttt{Distance}, and \texttt{Distance-Rnd} sampling methods all achieve high precision/recall, however, the \texttt{Distance-Rnd} has higher TGR and KNNGR than the other methods, suggesting that the embeddings space better captures the local structure. Hence we select this strategy for human annotators.}
    \label{tab:synthetic}
\end{table*}
\subsection{Evaluation Metrics}
We use similar metrics as \citet{tste} to measure the quality of the learned embedding.
The Triplet Generalization Ratio (\textbf{TGR}) describes the fraction of ground truth triplets that are not violated by the learned embedding. Given the number of possible triplet combinations are very large we instead sample a subset of 1000 possible ground truth triplets for this metric. The K Nearest Neighbour Generalization Ratio (\textbf{KNNGR}) captures how well local structure is preserved. Since the number of local clusters is unknown after training, we sample 70\% of the data and train a KNN on it and measure how many of the last 30\% fall into the correct clusters to get a measure of locality. The Signal-to-Noise Ratio Distance (\textbf{SNR}) \cite{Yuan_2019_CVPR} measures the similarity of comment embeddings.
Similarly, we compute statistics about the triplets gathered from the workers and the ground-truth annotation of the comments through three metrics: \textbf{Triplet agreement}, \textbf{precision}, \textbf{recall}, \textbf{weighted precision}, \textbf{weighted recall}. The Triplet agreement computes how many of the sampled text excerpts belong to the narrative of the anchor. The precision and recall measure the consistency with the ground truth narratives. These are computed per narrative and then averaged. In their weighted counterparts, the average is weighted by the number of text excerpts in each narrative.
\subsection{Results from playback simulation}
Across our synthetic experiments, we found three key ingredients to obtain embeddings of higher quality. Firstly, we found that using a modern sentence transformer, such as the 3B parameter T5 model \cite{T5}, was crucial for initially embedding the sentences in \textsc{papyer}\includegraphics[height=1em]{images/papaya.png}, before running SNaCK. Table \ref{tab:snack_backbones} compares embeddings with different networks. We found that a T5-3B Transformer gave the best embedding by measuring the triplet generalization loss. T5-3B performs better than T5-11B, despite being trained on the same data. A similar observation can be made in other benchmarks \cite{t5_3_better_11_worse} on which this model has been evaluated, indicating that size alone may be insufficient to determine the best embedding network. 

Secondly, the sampling strategy to gather the triplets used in SNaCK is important. Table \ref{tab:synthetic} shows SNaCK with different sampling techniques. We found that \texttt{Distance-Rnd} worked best, since it has a good compromise between exploiting the text similarities and exploring the solution space. This means it is able to present annotators with the more relevant triplets, than a method like \texttt{Random}. Furthermore, this sampling strategy was more robust to the hyperparameters found in SNaCK. We refer to the Appendix for ablation results on SNaCK hyperparameters.   

\begin{figure}
    \centering
    \vspace{-7mm}
    \includegraphics[width=\columnwidth]{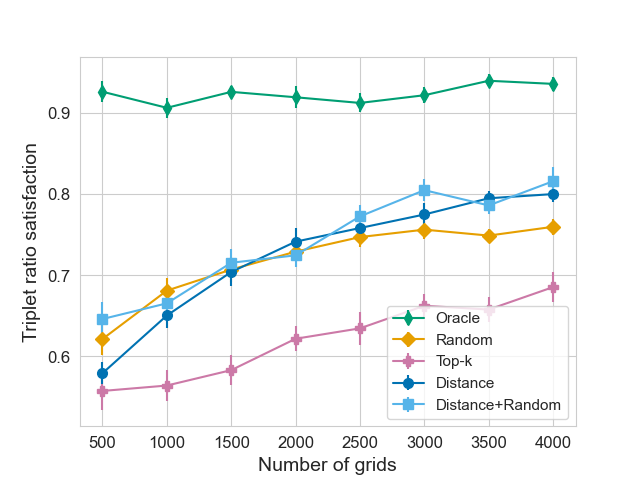}
    \caption{\textbf{Sampling strategy.} The triplet ratio satisfaction as a function of training data for different sampling strategies. We find that \texttt{Distance-Rnd} performs slightly better than \texttt{Distance}.}
    \label{fig:performance_training_data}
\end{figure}

\begin{figure}
    \centering
    \vspace{-5mm}
    \hspace{1.2mm}
    \includegraphics[width=0.98\columnwidth]{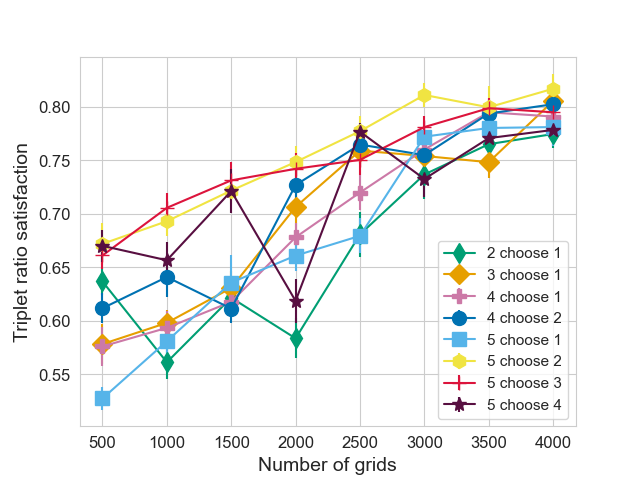}
    \caption{\textbf{Number of positive / negative examples.} Triplet ratio satisfaction as a function of the amount of text available and the number of clicks. The best setup is $5$ choose $2$.}
    \label{fig:numclicks}
    \vspace{-3.5mm}
\end{figure}

We also investigate how the amount of labeled data affects the sampling strategies. Figure \ref{fig:performance_training_data} shows the trained embedding improves when increasing the size of training data. This is the case for all sampling strategies. In all experiments the synthetic worker has access to both the ground-truth label of the sentences and using the distance between their embeddings, when selecting the most similar sentences to the anchor.
Lastly, we explored how the number of sentences $n$ and number of forced choices $k$ affects the performance. Figure \ref{fig:numclicks} shows that with the \texttt{Distance-Rnd} sampling strategy, presenting the synthetic worker with $5$ sentences and asking it to choose $2$ worked the best. Given that human workers cannot process text in parallel, increasing the number of sentences would not have the same benefits as in visual annotations \cite{Wilber_Kwak_Belongie_2014,psycho1,psycho2}. 
With these consolidated observations of the hyper-parameters and design choices from the play-back simulation in hand, we can proceed to collect labels from human annotators.


\begin{figure*}[ht]
\centering
\includegraphics[width=\textwidth]{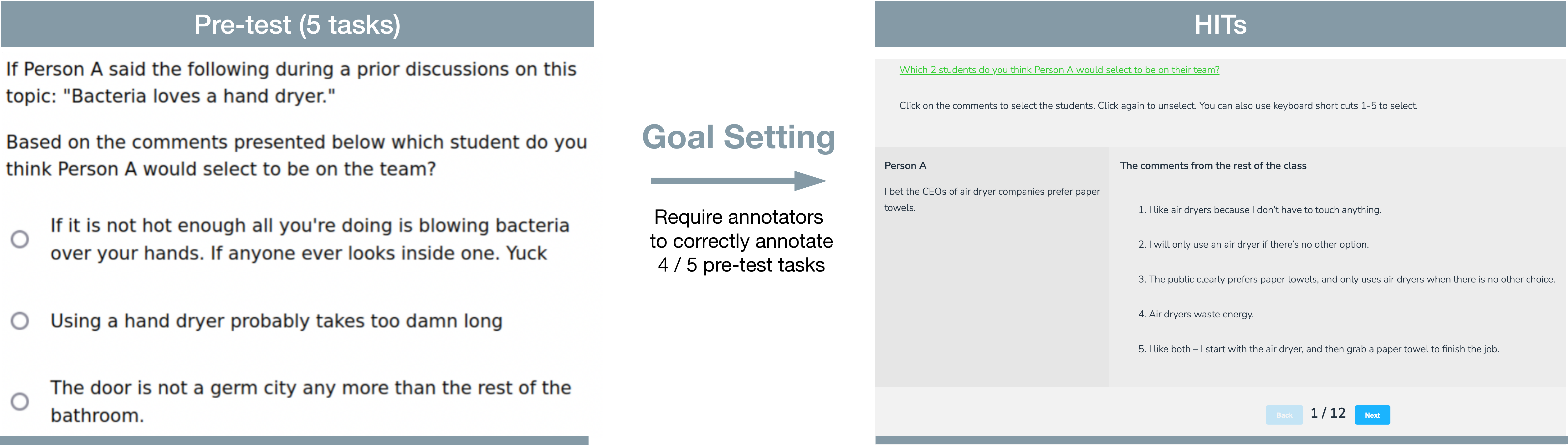}
\caption{ \textbf{Workflow for annotators.} Annotators are selected based on a multiple-choice pre-test that consists of five questions of which they must answer a minimum of four correct. The selected annotators are then asked to select the two best matching statements for the anchor. We use these selections to form triplets.}
\label{fig:flow_dia}
\end{figure*}

\section{Human Annotators}

The play-back simulation revealed several insights, but the synthetic workers differ from human annotators in several aspects. We now proceed to test our pipeline using human annotators. 

Figure \ref{fig:flow_dia} displays the overall flow of the data collection. Before the workers can accept our HIT, we ask them to take a pre-test, consisting of five multiple-choice questions where one sentence is shown and they have to select the best match from two sentences. If the worker passes four out of the five questions they are permitted to work on our HITs.
A worker is shown grid consisting of an anchor and a list of five text excerpts (see Figure \ref{fig:flow_dia}). The anchor is randomly sampled from the $600$ sentences and the five comments are sampled according to the Distance-Rnd sampling strategy, which we found to work the best in the play-back simulation. 
We ask the worker to select the two persons who would likely be on the same debate team as the anchor person, resulting in two debate teams with three persons each. The worker was paid \$1 per HIT, where each HIT contains 12 grids. For each HIT we include one \textit{catch trial}, i.e., a grid designed to be particularly easy to solve as we compose the five sentences  two with the same narrative as the anchor and three with irrelevant narratives. In order to get a richer similarity representation, and to examine the quality of the annotators, we also deploy \textit{sentinel examples}. These examples helps us track whether an anchor belongs to a specific narrative. We insert these as to verify if the annotator agrees with our list of narratives. We refer to the Appendix for the catch trial and sentinel example agreement results.
After gathering the data, we train SNaCK for 100k epochs. Across all experiments, we collected $2880$ grids, yielding around $20,000$ triplets. Collecting this data cost $\$480$. In our setup with one anchor, five candidates of which two must be selected, we found that the average worker spends around 3 minutes and 36 seconds to complete a HIT. However, the time to complete a HIT varies widely between workers: the fastest worker answered a HIT in 2 minutes, while the slowest used 23 minutes.  

\begin{figure*}
     \centering
     \begin{subfigure}[b]{0.45\textwidth}
         \centering
         \includegraphics[width=\linewidth]{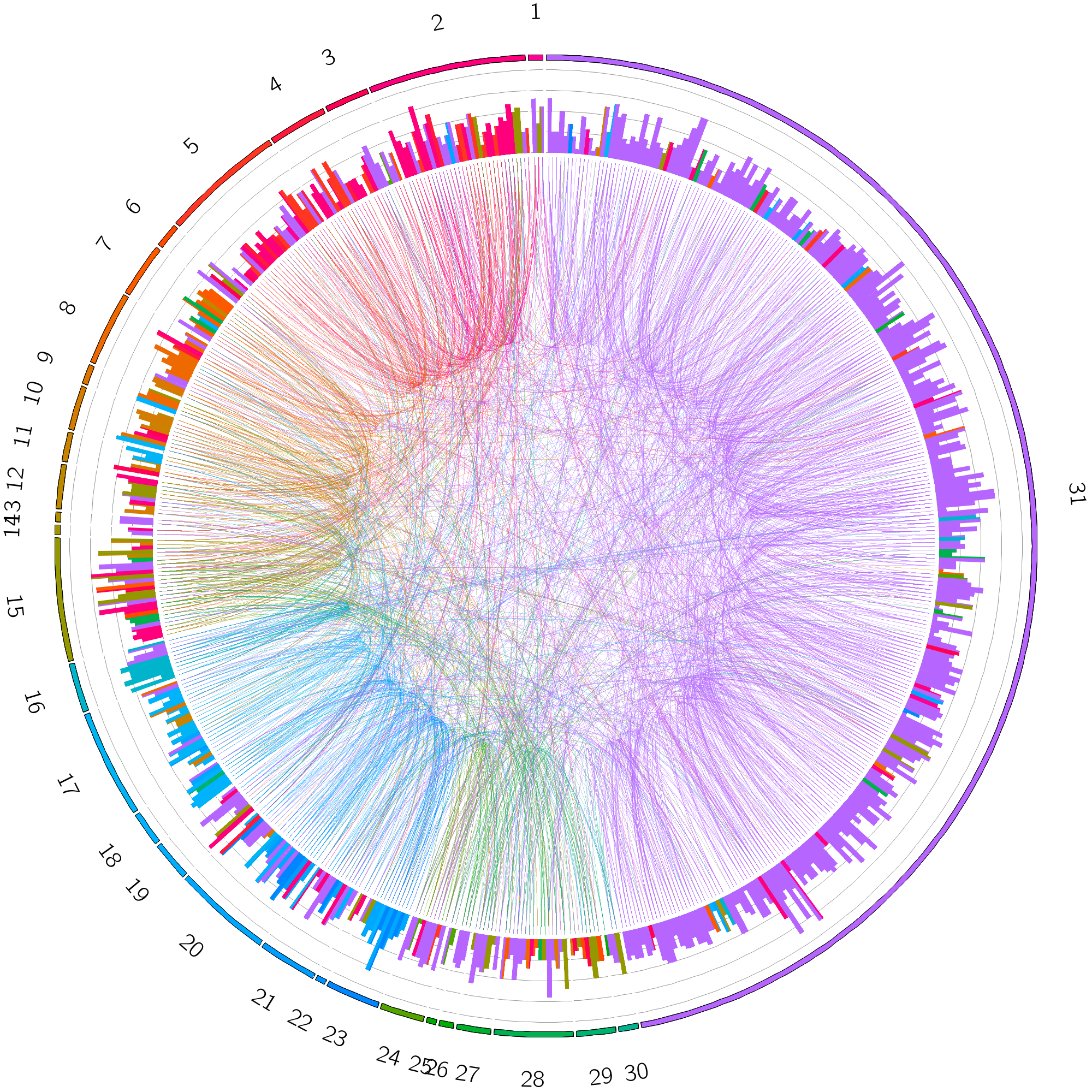}
    \caption{Positive pairs}
         \label{fig:circos_pos}
     \end{subfigure}
     \begin{subfigure}[b]{0.45\textwidth}
         \centering
         \includegraphics[width=\linewidth]{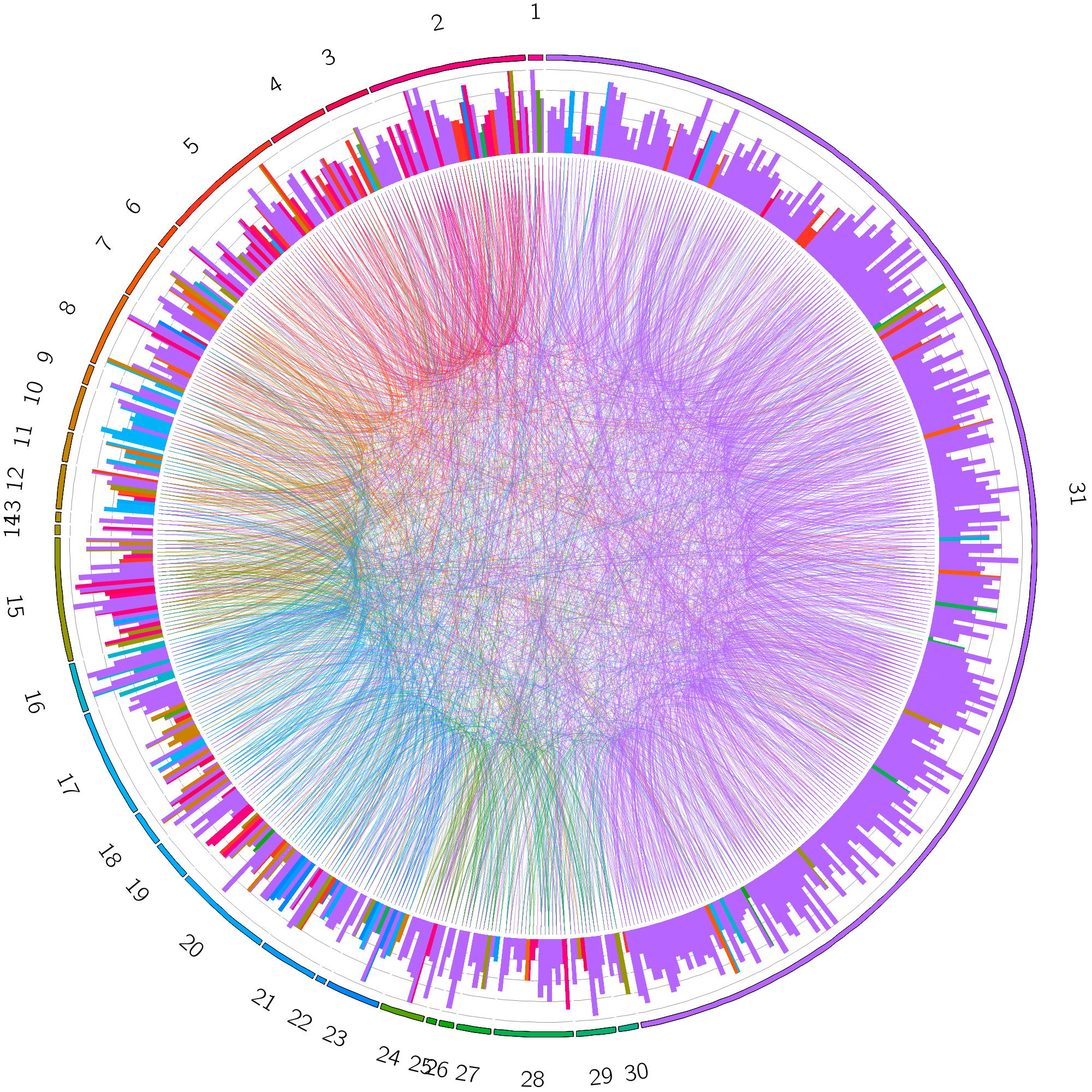}
    \caption{Negative pairs}
         \label{fig:circos_neg}
     \end{subfigure}
    \caption{\textbf{Visualisation of human annotations.} Each line represents a positive (a) or negative (b) human annotation for an anchor. The histograms in the circumference describe the number of incoming connections. The color of the histogram describes the class of the majority of incoming classes and the color of the lines describes the ground truth class of the anchor. If the color of the histogram matches the above class color, then the pair belongs to the ground truth class, and if the colors differ they do not. The numbers above the class color indicate the individual narrative classes.
    In (a) the red and blue links show that the anchor and selected answer sometimes share the same class. More generally speaking a trend can be identified in which narratives revolving around favoring paper towels (red colors) are linked together, which is similarly true for narratives favoring the air dryer (blue colors). In (b) the colors of the histogram and an above color circle do not match indicating a mismatch between the class of the anchor and answers that were not selected. This is reflective of the workers' ability to distinguish text originating from different classes from each other.}
    \label{fig:circos}
\end{figure*}

\textbf{Triplet data visualization.} The collected triplet annotations that have passed the catch trials are shown in two \textit{circos plots} \cite{Krzywinski18062009}. Figure \ref{fig:circos_pos} illustrates the connection from the anchor to the selected answers (i.e., anchor to positive) and Figure \ref{fig:circos_neg} shows the connections from the anchor to sentences that were not clicked (i.e anchor to negative). In both figures, we show a histogram of the number of times a connection is made to the specific anchor, and the color identifies the narrative of the most popular connection to that anchor. As such, the ideal scenario would be that the colors in the histogram match the narrative colors in Figure \ref{fig:circos_pos} and that they do not in Figure \ref{fig:circos_neg}. Figure \ref{fig:circos_pos} shows that some of the histogram colors match the class colors and there is a trend that they similarly fall within their four super categories, thus revealing the polarisation between the pro-paper towel and pro-hand dryer contingents.
Table \ref{tab:human_annotators} shows several statistics of the selected HITs that passed the catch trials in our mturk experiment. We notice that despite requiring catch trials to ensure annotations of high quality, there is still a large proportion of triplets where annotators select an answer that does not share the same narrative as the anchor. This is also reflected in the precision and recall ratios. This highlight how challenging the task at hand actually is, where even human annotators produce noisy labels.

\begin{table*}[t]
\centering
        {\fontsize{9pt}{9pt}\selectfont
    \begin{tabular}{lllllll}
        \toprule
         & Agreements ($\uparrow$) & Disagreements ($\downarrow$) & Precision ($\uparrow$)& Recall ($\uparrow$)  & Weighted Precision ($\uparrow$)& Weighted Recall ($\uparrow$)\\ \midrule
         Random Annotator   & 3093  & 4455 & 9.75  &   9.75    &  38.91     & 40.71  \\
         Human Annotator    & 3447  & 4107  & 15.97 & 15.73 & 43.61 & 47.93 \\
        \bottomrule
    \end{tabular}
    }
    \caption{\textbf{Statistics of annotations}. Overall there is a trend of having a low agreement, precision, and recall indicating that the annotations are noisy. The triplets were selected based on annotators that passed all catch trials. This highlights that the task at hand is challenging even for human annotators.  
    }
    \label{tab:human_annotators}
\end{table*}

\section{Experiments}

\textbf{Baselines.} We compare the SNaCK embeddings with several popular unsupervised topic models, namely LDA~\cite{LDA}, methods using pre-trained embeddings networks~\cite{sia-etal-2020-tired,mimno-thompson-lda}, such as BERT and T5, and a mixture of these inspired from \cite{context_topic_identify}. 
For LDA and mixed models, we follow the standard protocol~\cite{sia-etal-2020-tired} and tokenize the text using NLTK~\cite{bird2009natural}, lowercase the tokens, remove stopwords, punctuation, digits, and URLs, and transform words into their stem versions and fix typos. We then convert these tokenized sentences into a document-term matrix to be used for LDA with $32$ topics. To visualize in 2D, we run \textit{t}-SNE on top of the document-term matrix. Recent work on topic models~\cite{sia-etal-2020-tired} suggests that traditional topic models can be replaced with clustering of 2D projections of deep sentence embeddings. We explore this using different transformers followed by different projections such as \textit{t}-SNE \cite{tsne} and UMAP \cite{2018arXivUMAP}.

\section{Results}

\begin{figure*}
     \centering
     \begin{subfigure}[b]{0.24\textwidth}
         \centering
        
         \includegraphics[width=\linewidth]{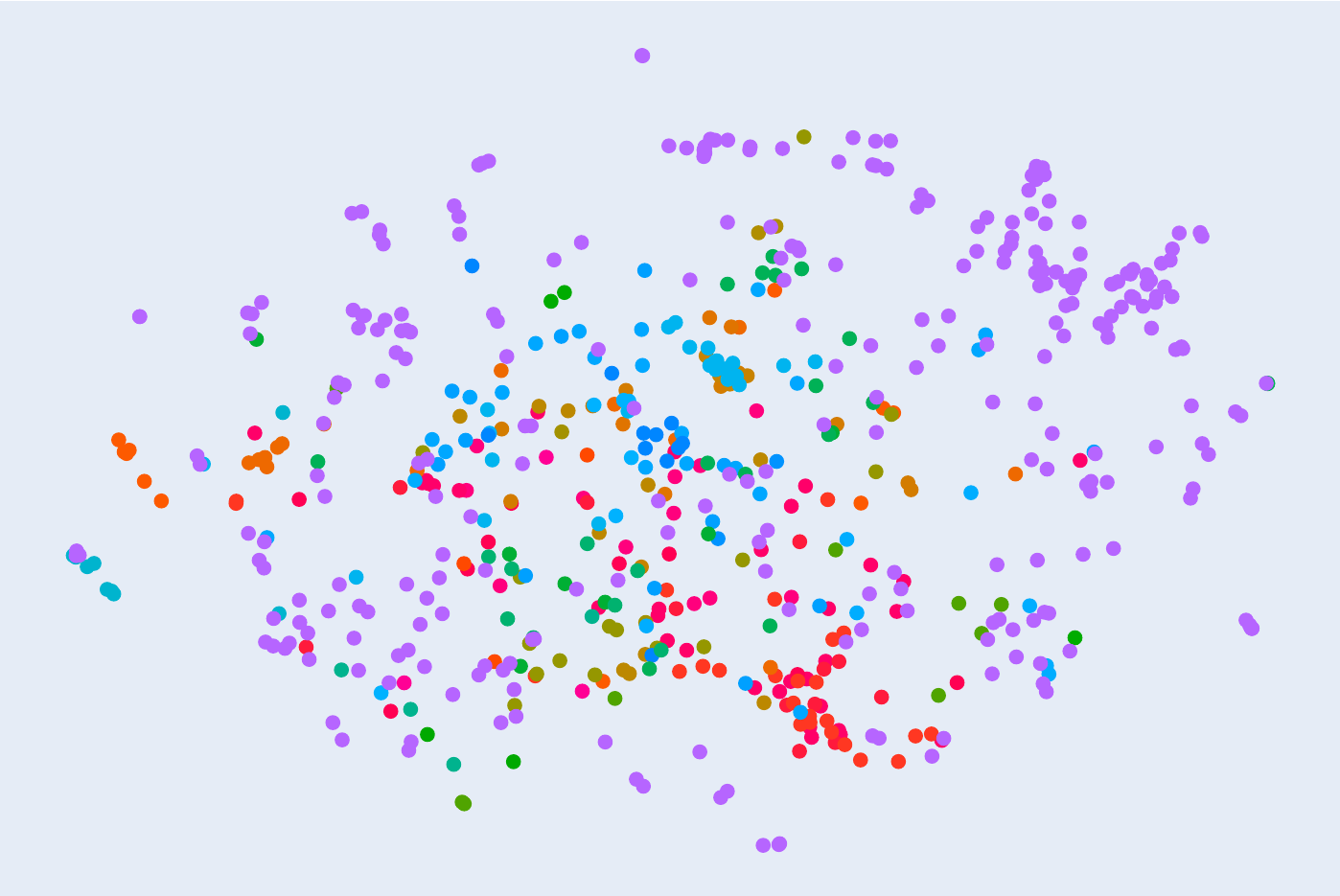}
          \caption{\textit{t}-SNE T5}
         \label{fig:TSNE_T5}
     \end{subfigure}
     \begin{subfigure}[b]{0.241\textwidth}
         \centering

         \includegraphics[width=\linewidth]{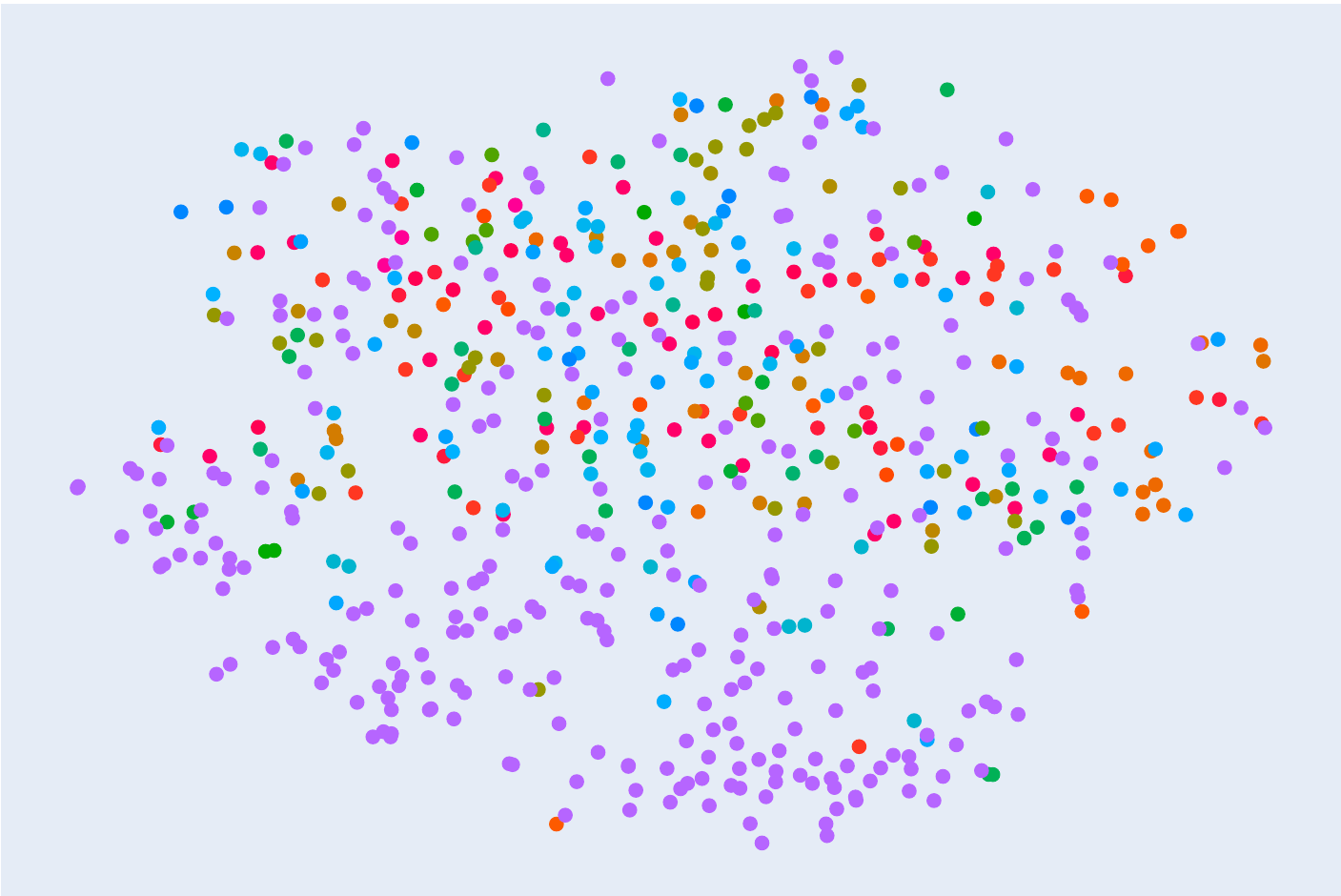}
        \caption{\textit{t}-SNE BERT}
         \label{fig:TSNE_BERT}
     \end{subfigure}
     \begin{subfigure}[b]{0.2385\textwidth}
         \centering

         \includegraphics[width=\linewidth]{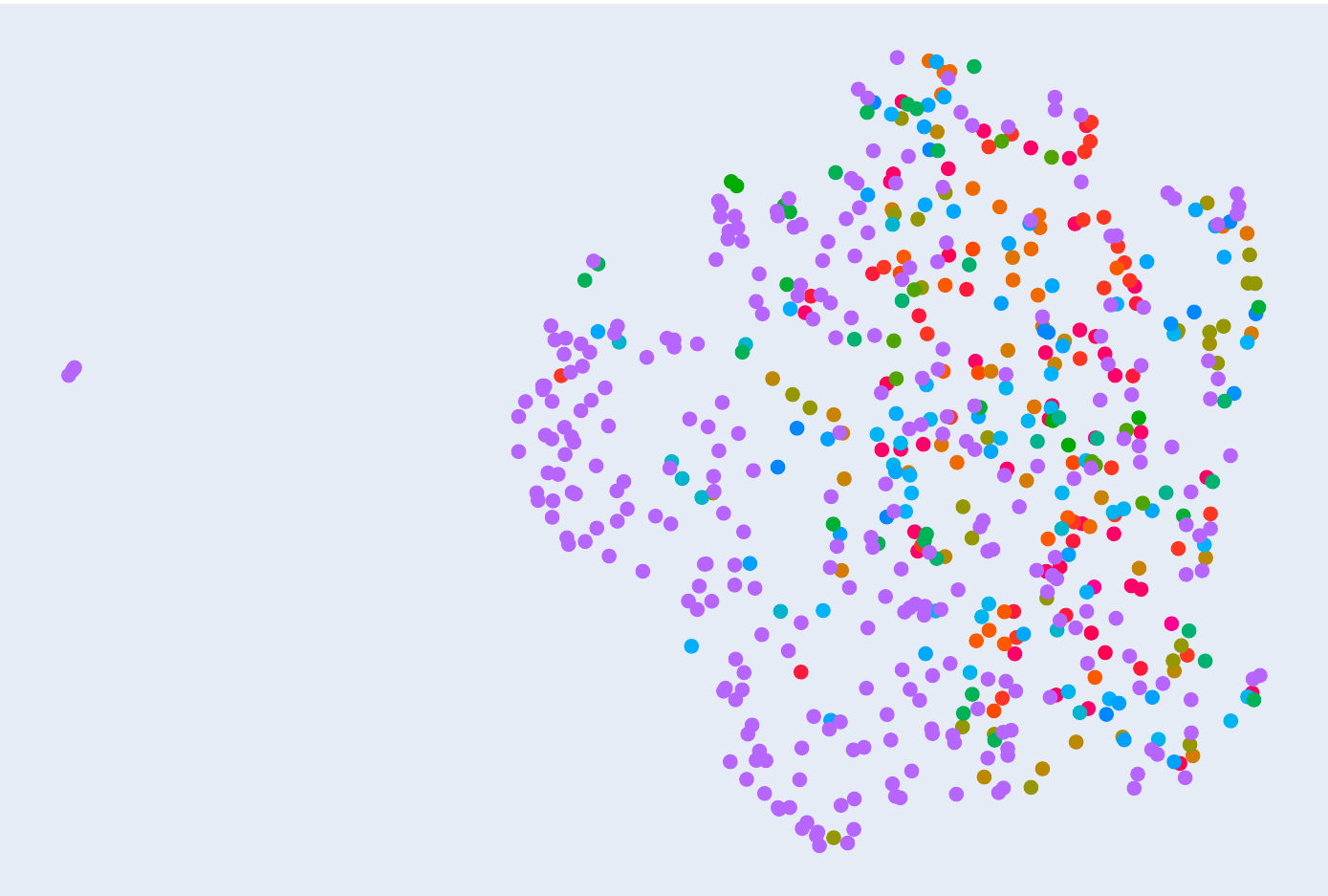}
         \caption{UMAP BERT}
         \label{fig:UMAP_BERT}
     \end{subfigure}
     \begin{subfigure}[b]{0.245\textwidth}
         \centering
         \includegraphics[width=\linewidth]{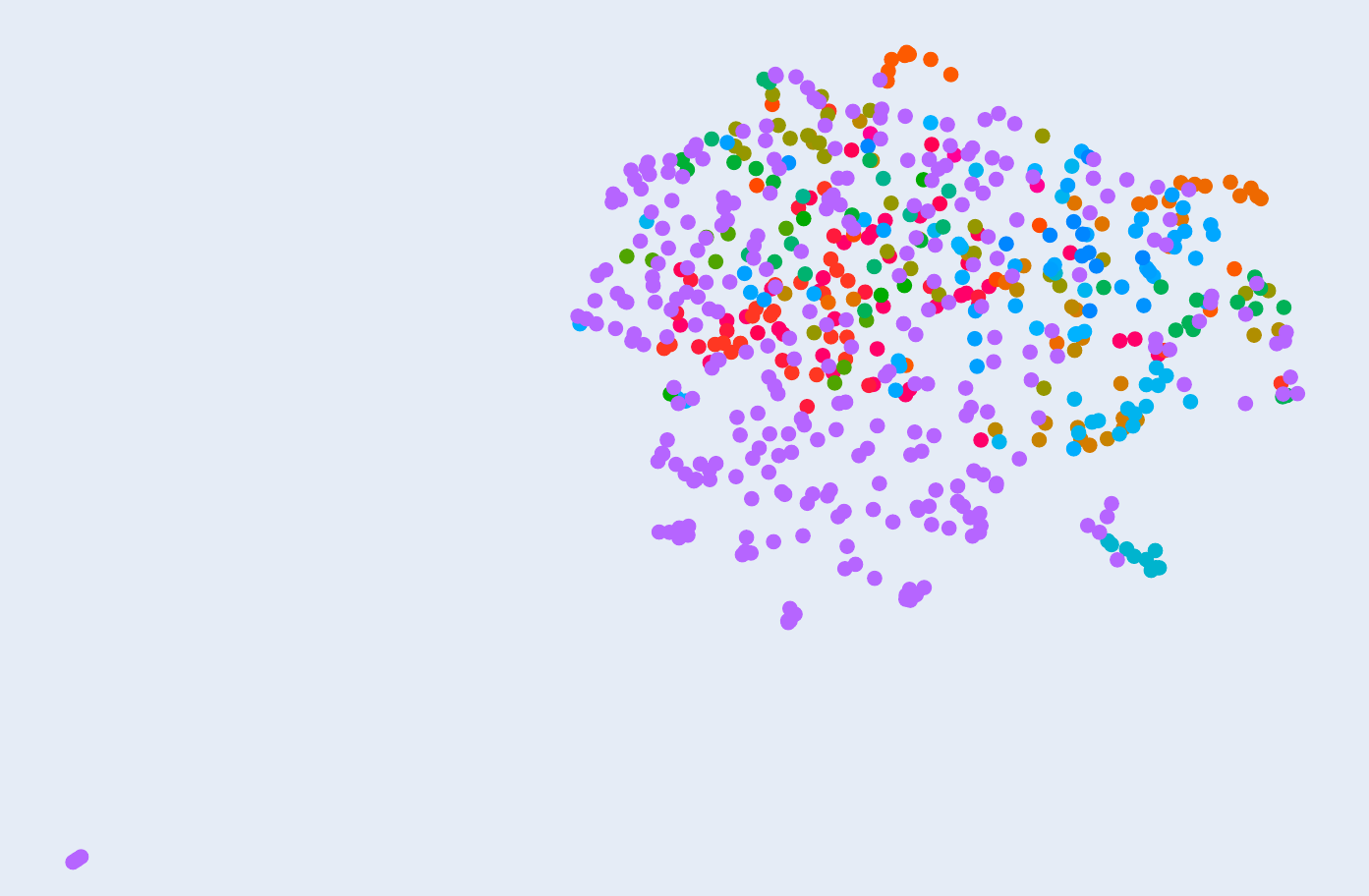}
         \caption{UMAP T5}
         \label{fig:UMAP_T5}
     \end{subfigure}
    
     \centering
     \begin{subfigure}[b]{0.241\textwidth}
         \centering
         \includegraphics[width=\linewidth]{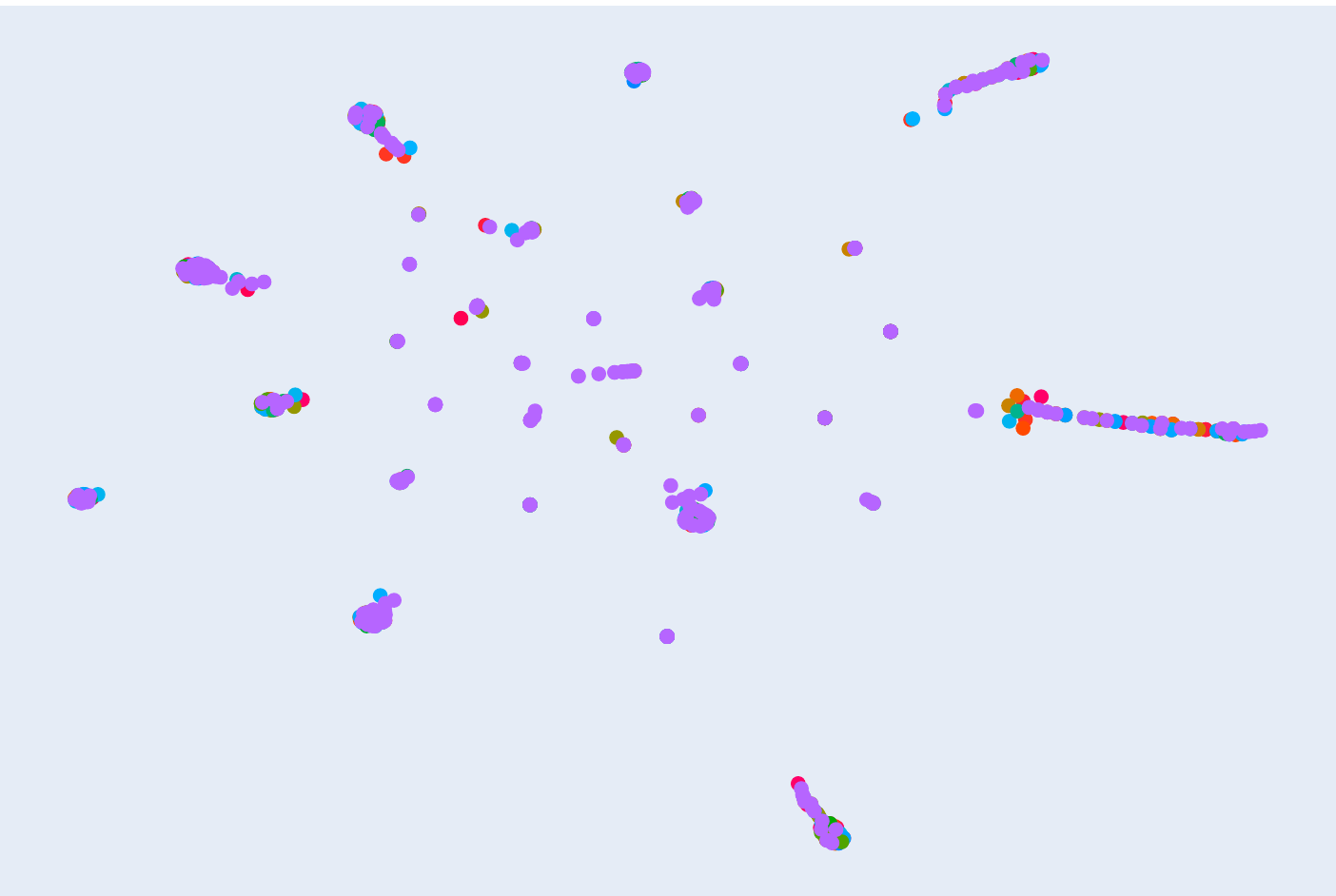}
         \caption{LDA}
         \label{fig:LDA}
     \end{subfigure}
     \begin{subfigure}[b]{0.2425\textwidth}
         \centering
         \includegraphics[width=\linewidth]{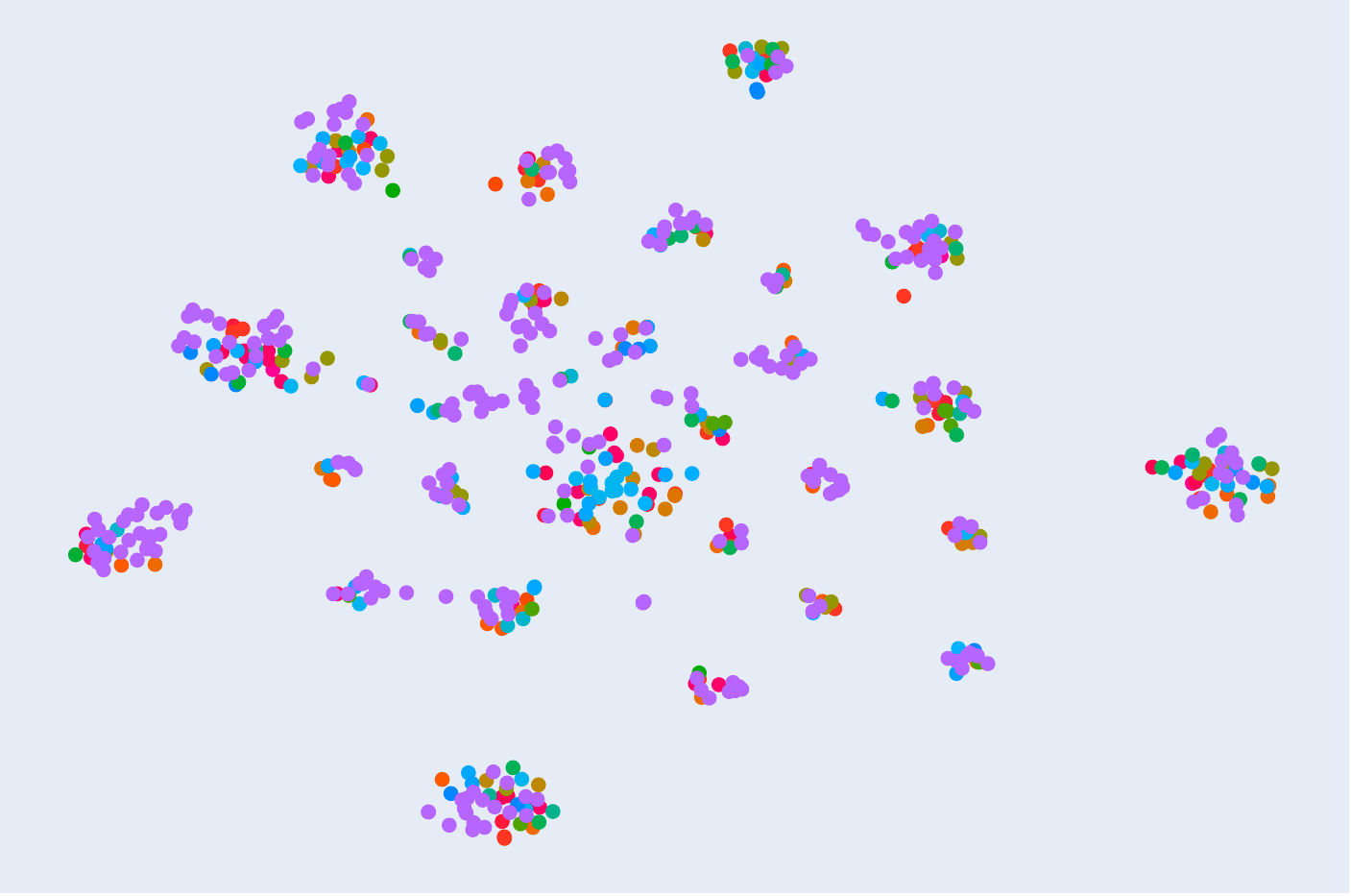}
         \caption{\textit{t}-SNE BERT + LDA}
         \label{fig:LDA_BERT}
     \end{subfigure}
     \begin{subfigure}[b]{0.2395\textwidth}
         \centering
         \includegraphics[width=\linewidth]{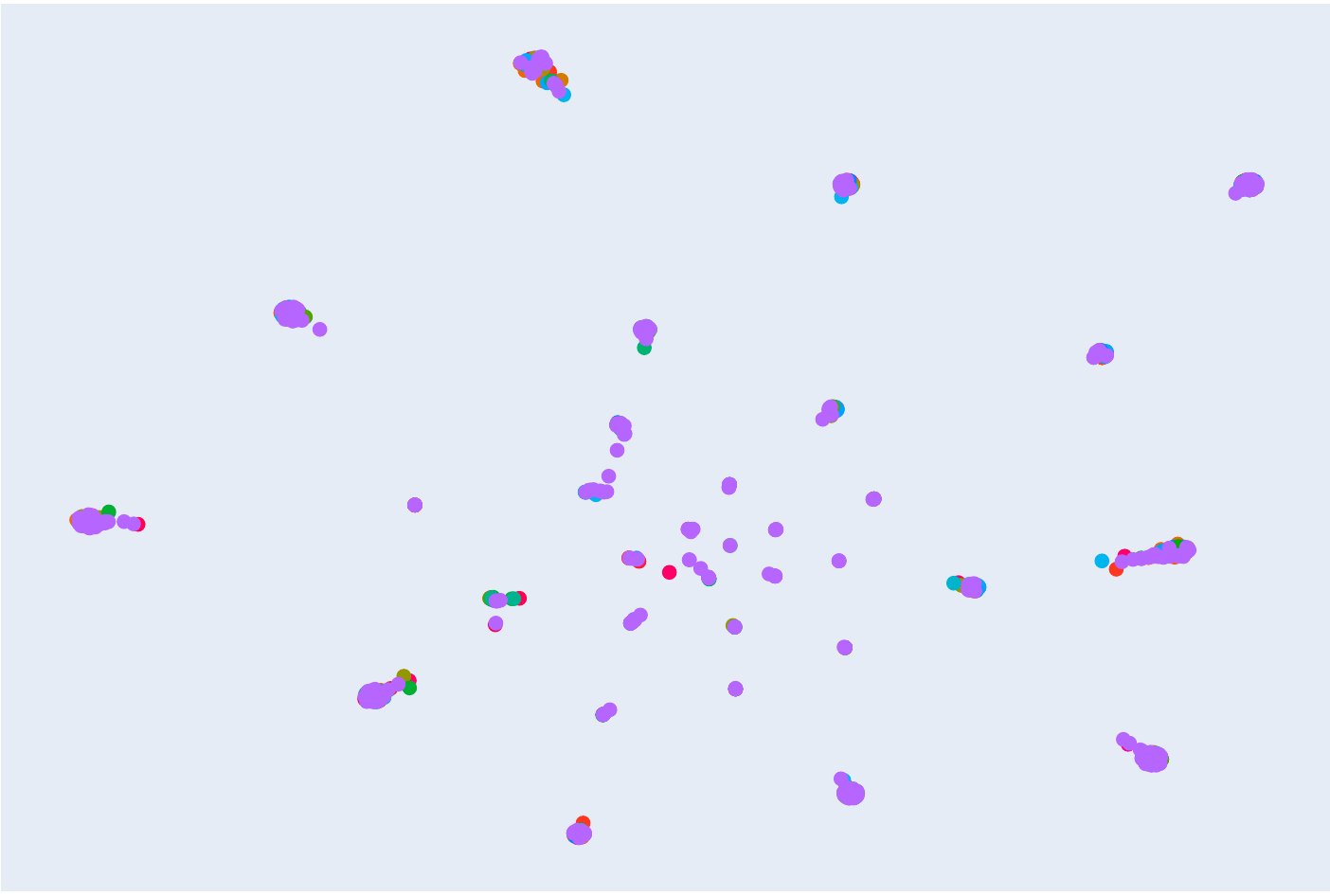}
         \caption{\textit{t}-SNE T5 + LDA}
         \label{fig:LDA_T5}
     \end{subfigure}
     \begin{subfigure}[b]{0.2408\textwidth}
         \centering
         \includegraphics[width=\linewidth]{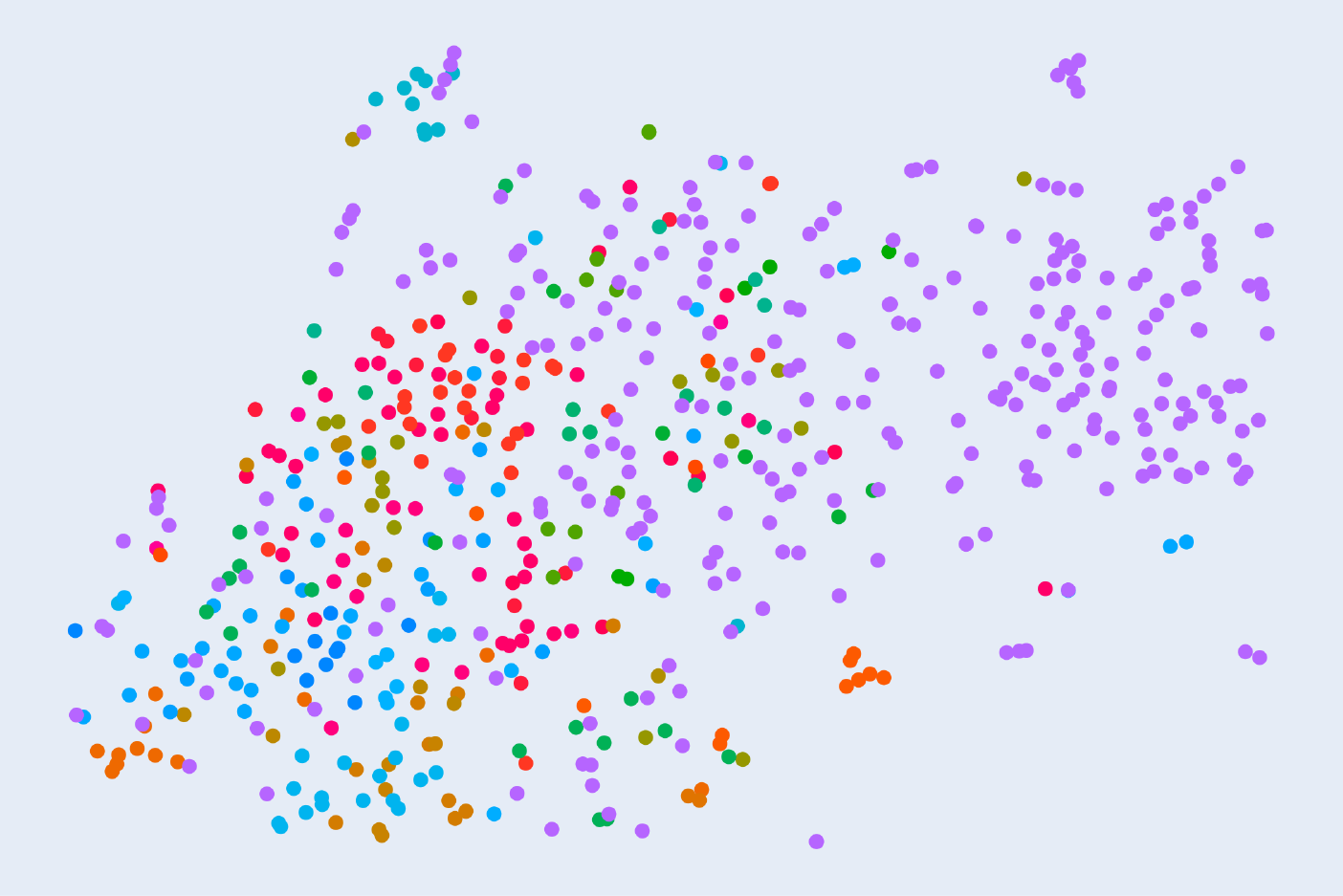}
         \caption{SNaCK T5}
         \label{fig:snack}
     \end{subfigure}
    \caption{2D visualizations using different encoders and embedding projections. Sentences are coloured according to the crystallised narrative they belong to using the colors from Figure \ref{fig:latent_narrative}.  Despite showing more local structure in Figure e-g the quality of the embedding is lower as shown in Table \ref{tab:real}. Figure h illustrates the transition from the initial \textit{t}-SNE embedding as shown in Figure a to the SNaCK embedding by supplying human annotations. }
    \label{fig:embedding_spaces}
\end{figure*}

\begin{figure*}
    \centering
    \includegraphics[width=\textwidth]{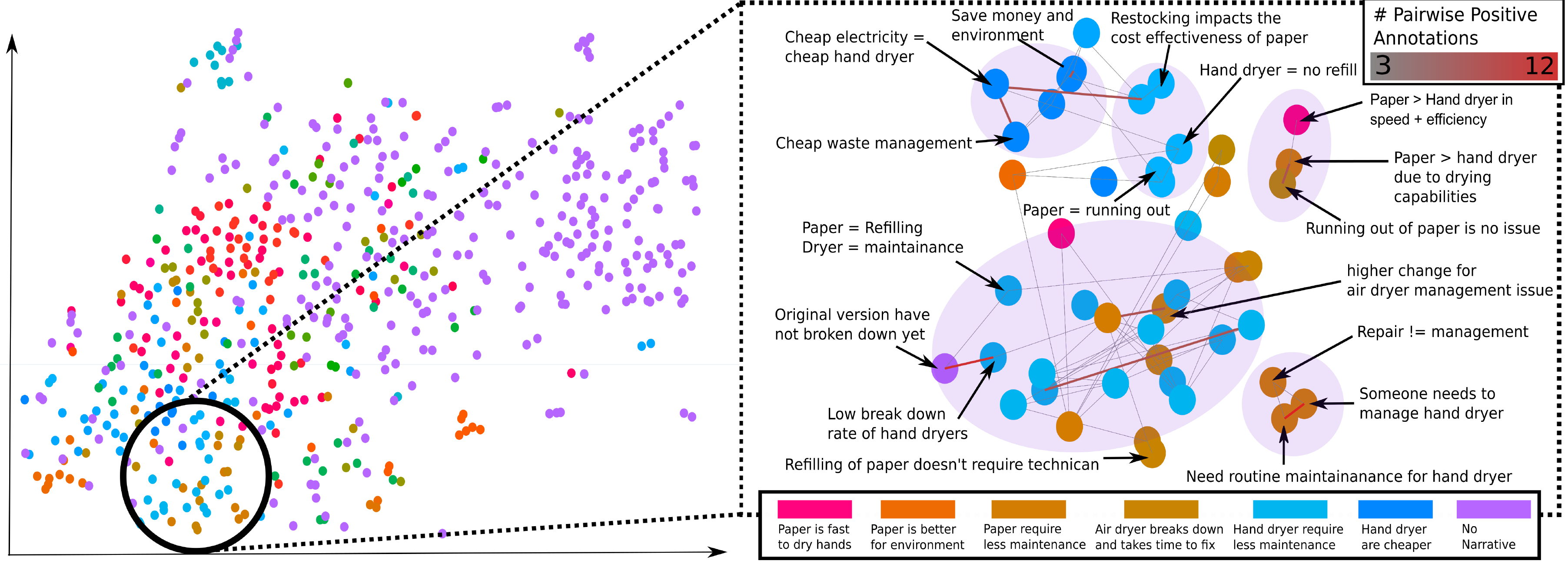}
    \caption{Zoom in on the SNaCK embedding. Relations chosen by human annotators are visualized as edges, which are colored according to the number of annotations (gray = 3, red = 12). Related terms are grouped, either due to similar class or discussion families within yellow circles. Six such families represent the complicated nature and are shown in the inset on the right.}
    \label{fig:snack_circle}
\end{figure*}

Figure \ref{fig:LDA} shows that the simple LDA baseline does not discover the prevailing narratives. This is not surprising, since LDA is known to struggle when the amount of data per topic is limited. Furthermore, the prevailing narratives are heavily context-based, which word co-occurrence based methods such LDA do not handle. On top of that comes the plenitude of sarcasm, irony, and humor that language models are known to struggle with, thus doing LDA on top of deep features from language models remains ineffective (Figure \ref{fig:embedding_spaces} (f, g)). This is also evident when projecting the T5 or BERT embeddings into the low dimensional embeddings space using either UMAP or \textit{t}-SNE (Figure \ref{fig:embedding_spaces} (a)-(d)). In contrast, Figure \ref{fig:snack} shows that incorporating human annotated triplets into the representation highlights interesting clusters that obey the crystalized narratives. 

To better visualize this, Figure \ref{fig:snack_circle} shows the obtained SNaCK embedding (left) and zooms into a region in the embedding space (right), where we display the input sentences for several embeddings. 
We highlight that SNaCK manages to find local clusters, and although some clusters have mixed narratives, the text emphasizes that they are related. 
For instance, the largest cluster in the zoomed region revolves around maintenance, whereas the cluster at the top of the zoom discusses management. Similarly, the cluster to the right focuses on the environmental costs. These clusters highlight that the combination of human and machine kernels can lead to embeddings that discover the underlying narratives from online discussions.  

These visual observations are backed up with quantitative experiments. Table \ref{tab:real} shows that the SNaCK and UMAP-T5 achieve the highest triplet generalization and k-NN ratio compared to the other baselines. The SNaCK embeddings further achieve a lower SNR than UMAP-T5, suggesting a better representation. We note that both SNaCK and UMAP-T5 have higher SNR than \textit{t}-SNE-T5, suggesting that on average positive pairs are closer for \textit{t}-SNE-T5 than for SNaCK or UMAP. This is caused by the projection of \textit{t}-SNE-T5 which maps sentences to a plane with a range from -5 to 5 for both axes, while the SNaCK method project points to a plane with an axis range from -40 to 40 for both axes, thus yielding greater distance between points, resulting in higher SNR.

\begin{table}[ht]
\setlength\tabcolsep{5pt}
\footnotesize
\center
    \begin{tabular}{llllllll}
        \toprule
        Method  & TGR($\uparrow$) & KNNGR ($\uparrow$) & SNR ($\downarrow$) \\ \midrule
        \textit{t}-SNE-BERT  & $55.33 \pm 1.55$ & $14.30  \pm 2.63$  & 2.59  \\
        \textit{t}-SNE-T5 & $58.93 \pm 2.28$ & $31.05 \pm 3.24$ & $\boldmath{0.47}$  \\
        UMAP-BERT  & $54.39 \pm 1.32$ & $15.91 \pm 2.71$ & 2.49  \\
        UMAP-T5 & $61.44 \pm 2.61$ & $\boldmath{33.44 \pm 4.25}$ & 1.48  \\
        \textit{t}-SNE-LDA  & $53.34 \pm 0.51$ & $7.31 \pm 1.42$ & 3.82  \\
        \textit{t}-SNE-BERT-LDA  & $54.01 \pm 2.47$ & $8.17 \pm 3.01$ & 3.93  \\
        \textit{t}-SNE-T5-LDA  & $52.56 \pm 1.14$ & $9.56 \pm 3.54$ & 3.75  \\
        SNaCK-T5   & $\boldmath{67.61 \pm 1.13}$ & $33.11 \pm 3.07$ & 1.17 \\
        \bottomrule
    \end{tabular}
    \caption{\textbf{Discovery of prevailing narratives.}  All models are evaluated 10 times using 1 trained model (Ratios multiplied by 100).}\label{tab:real}
    \vspace{-3mm}
\end{table}


\section{Conclusion and Discussion}

In this work, we advocate for the necessity to model the unfalsifiable claims that weave through online discussions on social media. We cast this problem as a metric learning task, where we aspire to cluster unfalsifiable claims into a subset of crystalized narratives. 
To study this task, we present \textsc{papyer}\includegraphics[height=1em]{images/papaya.png}, a dataset, based on hand drying in public restrooms, suitable to study and evaluate methods for narrative discovery. 
We find that recent, large transformer models are unable to discover the prevailing narratives. We demonstrate that by combining machine and human kernels, we can learn a representation that better captures the structure of the narratives.
We emphasize that finding the prevailing narratives is a very challenging task, requiring an understanding of context, humor, and sarcasm, which is  exemplified by low precision among human annotators. We, therefore, hope that our dataset will facilitate future research to better understand and discover the prevailing narratives in online discussions.

\textbf{Limitations and Future Work.} In this paper, we focused on narratives related to hand drying in public restrooms. We emphasize that a similar procedure could be performed on other topics, such as cryptocurrencies, maternity leave of female sports stars, or elections in the US to reveal interesting, prevailing narratives on these topics. 
We highlight that the presented sampling strategies are simple and the results are biased due to using heuristic based workers. Since we found that the sampling strategies are important for high information gain per triplet annotation we believe more sophisticated sampling methods should be explored, such as using a classifier \cite{Jia2021IntentonomyAD} to recommend sentences to annotators. These methods should recommend sentences that are related to the anchor to distinguish the difficult cases from each other and ensure that we avoid sampling trivial or redundant examples for a worker to read.

Caution should be used about the implication of our results as the current metrics do not take into account the hierarchical structure of the labels, e.g., clustering two sentences with ``Air dryers blow germs around the room'' and ``Paper towels are more hygienic'', as labels will be incorrect in our evaluation, although the first is a subset of the latter. Thus, our evaluation metric is too conservative and evaluation methods that take this hierarchical structure into consideration should be explored to consolidate our findings.

A direction of future work is to explore the narratives in a geographical setting. Narratives related to sports stars or cryptocurrencies would likely show different distributions of narratives depending on culture and governmental policies. 
We believe associating narratives with geomarkers that directly guide us to selecting specific crowd workers with cultural understanding would  improve the discovery of new narratives. In addition to geographical information, modeling the users with collaborative filtering or recommender systems would likely also improve performance as narratives supported by certain users will correlate. 

\section{Acknowledgements}

We would like to thank Sagie Benaim, Stella Frank, and Vésteinn Snæbjarnarson for their comments on an earlier draft. PEC and SB are supported by the Pioneer Centre for AI, DNRF grant number P1. FW is funded by The Technical University of Denmark. MJ is supported by a Facebook AI research grant.

\bibliography{references}

\appendix
\section*{Appendix}

The aim of our work is to investigate the complex narrative landscape hidden behind social media posts, and to lay the groundwork for the research in this domain. 
Such research can foster the development of systems to identify harmful posts and to reduce social media abuse and misinformation. 
In our work we proposed to explore narratives revolving around hygiene in public restrooms introducing a new textual dataset.
In the supplemental material, we provide the following items that shed further insight on these contributions: 

\begin{itemize}
    \item Details for reproduce our results (Experimental details)
    \item Information about data collection process (Dataset Creation Details)
    \item Data analysis (Dataset Analysis)
    \item Illustrative annotation examples (More Examples from \textsc{papyer}\includegraphics[height=1em]{images/papaya.png}) 
\end{itemize}

\section{Experiment Details}
\label{supsec:model}

\subsection{Experimental setup}
\paragraph{Model details}
In this section, we describe our Sentence transformer model T5, as
well as their training procedures in detail. We use the same hyperparameters as in the study of Raffel et al. (2019). 

\paragraph{Architecture}
To extract an embedding from our sentences, we use a T5~\cite{T5} transformer model~, while similar to BERT \cite{devlin-etal-2019-bert} but T5 was pretrained on the Colossal Clean Crawled Corpus dataset (750 GB) and trained using a variety of tasks including translation, question answering, and classification. \\

The text input consists of a sequence of tokens, provided by the wordpiece tokenizer. \cite{wu2016google, sennrich-etal-2016-neural} These tokens are surrounded by two special tokens, ${[CLS], w_1 , . . . , w_T , [SEP]}$ .

The body of our Transformers consist of a encoder and decoder, each with 24 attention layers, where each layer consists of a self-attention mechanism, optional encoder-decoder attention, and a feed-forward network. The feed-forward networks in each layer consist of a dense layer with an output dimensionality of $dff = 16,384$ followed by a ReLU nonlinearity and another dense layer. \\

The “key” and “value” matrices of all attention
mechanisms have an inner dimensionality of $dkv = 128$ and all attention mechanisms have 32 heads. All other sub-layers and embeddings have a dimensionality of $dmodel = 1024$.

Finally, the global representation for a sentence is obtained by the pooled representations for the text modality, i.e we extract processed token with the same index as the [CLS] token.

\begin{figure*}[t]
\centering
\includegraphics[width=\textwidth]{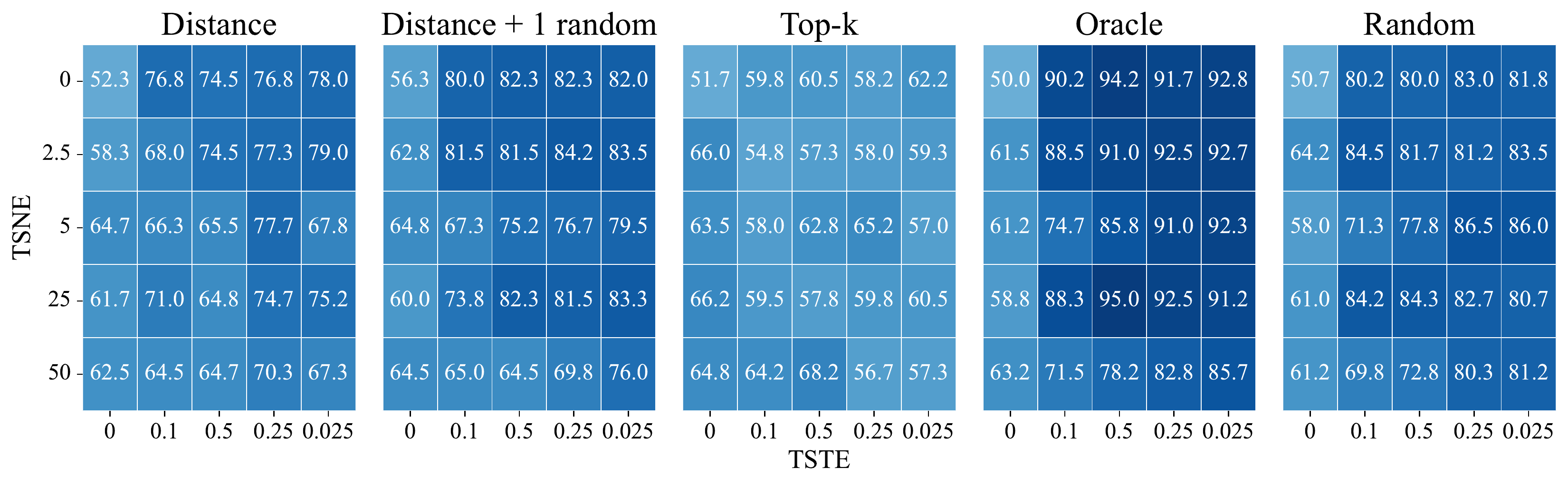}
\caption{Ablation study measure triplet generalization ratio of 25 different hyperparamters and sampling techniques using synthetic workers. The Oracle method outperforms all other methods as it is only trained on triplets from our ground truth. Surprisingly both the Random and distance + random based method seem to behave the same way, despite using different heuristics.   }
    \label{fig:ablation}
\end{figure*}

\paragraph{Training details} We train our models on a single NVIDIA 3060 GPU.
We reimplement the SNaCK algorithm in Pytorch~\cite{NEURIPS2019_9015} as the older version was written in Cython or python 2 and wasn't capable of using modern GPU support.

We follow SNaCK and use SGD using the initial momentum of $\beta=0.5$ and final momentum of $\beta=0.8$. We switch to the final momentum parameter after 20 gradient steps. The learning rate is set to 1. \\

We follow SNaCK and \textit{t}-STE and implement early exaggeration, additional weighting of the \textit{t}-STE loss increasing with the number of triplets, we also employ the momentum hack as introduced in the original \textit{t}-SNE paper.

We make our own optimizer as the weighting of the losses as done in SNaCK is made on the gradient level and not on the losses, however unlike the original implementation of SNaCK we leave the option of switching between this special SGD optimizer and the standard ones in PyTorch but where we do the weighting on the losses.  \\

Running  SNaCK algorithm using 10k triplets for 100k iterations takes around 15 minutes.
The parameter sets giving the best performance in our synthetic setup is used for our experiments involving MTurks.

\subsection{Hyper parameters}
\paragraph{SNaCK loss}
Given that our loss function $C_{SNaCK} = \lambda C_{SNE} + (\gamma)C_{STE}$ consist of two hyper-parameters, we conducted grid search for $\lambda$ with the range $\{0.0, 2.5, 5, 10, 25\}$ and $\gamma$ with the range $\{0.5, 0.25, 0.1, 0.025, 0.0\}$ as shown in Figure \ref{fig:ablation}.

We set $\lambda = 5$ and $\gamma=0.1$ in the end, in which the order of magnitude is consistent with the parameters used in previous work~\cite{Wilber2015LearningCE}. We set the perplexity equal to 30 after initial experimentation.

\subsection{Identifying sampling technique}

To quantify which type of sampling strategy should be used for human annotators we investigate the effect of utilising 4 different sampling strategies (Random, Oracle, Distance, Top-k and Distance-random). \\

We analyse the sampling strategies based on the triplet generalisation ratio with a given hyper parameter pair ($\gamma$, $\lambda$) as well as triplets made from 300 synthetic HITs (around 4k triplets). The experimental results are shown in Figure \ref{fig:ablation}.  

All results should however be carefully considered as they are intertwined with the synthetic workers, thus the worker always knows the ground truth label of every sentence. A real MTurker doesn't have such information and as such we will stick to the Distance + Random approach for the real experiments.

\paragraph{By hyperparamter free methods}
Of the sampling strategies that themselves doesn't have any hyperparameters are: Random, Oracle and Top-k. Although $k$ in Top-k might seem like a hyperparameter it is always set to be the maximum number of sentences in a HIT. 

As such the method relies on a KD-tree (similar to the other distance methods), but will however only consider the k most similar sentences, according to the KD-tree, and will therefore not change as we only consider 1 HIT configuration (5 choose 2 for instance).
The other strategies; random and oracle, either sample triplets completely at random or sample a positive and negative example according to the class of the anchor.

\paragraph{Neighbouring strategies}
Based on creating neighbouring information by applying a kd-tree on the learning embedding we can design different strategies that uses this information.
First we simply pick the top 5 nearest neighbours.

Secondly we sample from the top $k$ nearest neighbours based on the distance from the anchor. As such $k$ can be much larger than the number of sentences shown in the HIT and we wish to investigate if there is a benefit of choosing one value of $k$ over another. \\

Formally, given a the number of nearest neighbours $k$ which resulting embedding will give the best clusters? The larger the value of $k$ the more samples can be considered and the less likely it will be to select nearby datapoints as shown in Figure \ref{fig:topk}. \\


Additionally one can also apply a mix of multiple strategies. We tried the KNN using distance based sampling with 1 random element, we find that including one random sampling will ensure that we visiting far away datapoints that may be outliers of the anchor class. 
Similarly we also ensure that if a cluster of datapoints of the same class exists not all of the datapoints will be shown and thereby ruin the already formed cluster.

\begin{center}
\includegraphics[width=\linewidth]{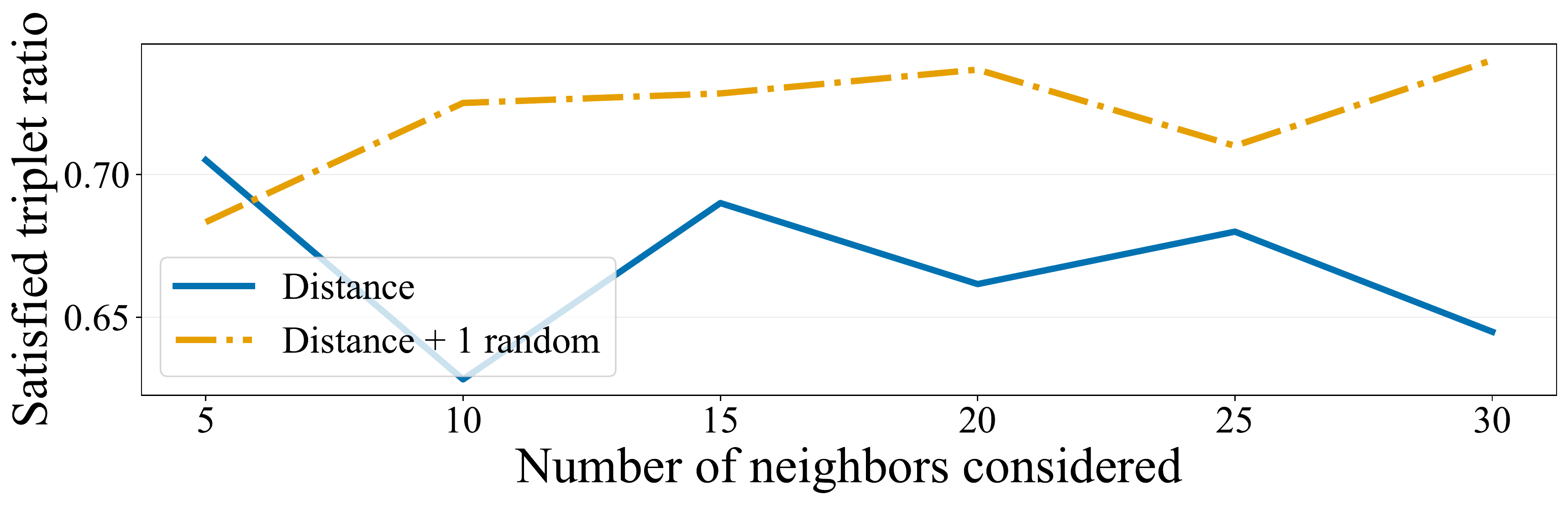}
\captionof{figure}{Effect of increasing the number of neighbours considered when sampling points proportionally to their distance from the anchor. In general, the triplet ratio peaks at $k = 30$, $k \in \{5, 10, 15, 20, 25, 30\}$. Y-axis is a ratio from 0 to 1.}\label{fig:topk}
\end{center}

\paragraph{Recall and Retrieval scores}
As mentioned in the main text a single average across classes is reported, however since the recall and retrieval scores was calculated per class we report these numbers in Table \ref{tab:pr_table} as other ways of aggregating the numbers could have been used (such as weighting the numbers by elements in the class).

\begin{table}[ht]
\setlength\tabcolsep{5pt}
\footnotesize
\center
\begin{tabular}{lll}
\toprule
            & Precision & Recall \\
\midrule
Class 1  & 0       & 0        \\
Class 2  & 34.57   & 31.46    \\
Class 3  & 3.12    & 3.12     \\
Class 4  & 13.89   & 8.77     \\
Class 5  & 25      & 19       \\
Class 6  & 0       & 0        \\
Class 7  & 38.89   & 46.67   \\
Class 8 & 22.22      & 21.05  \\
Class 9  & 20      & 19.05 \\
Class 10 & 22.73   & 31.25    \\
Class 11 & 4.55    & 4.76     \\
Class 12 & 13.64   & 7.14    \\
Class 13 & 12.5    & 12.5     \\
Class 14 & 0    & 0     \\
Class 15 & 16.15   & 17.65    \\
Class 16 & 48.21    & 58.70    \\
Class 17 & 37.93    & 26.19    \\
Class 18 & 21.43    & 23.68    \\
Class 19 & 3.57     & 2.44    \\
Class 20 & 15     & 13.95    \\
Class 21 & 2.38     & 1.75    \\
Class 22 & 0     & 0    \\
Class 23 & 25     & 16.67    \\
Class 24 & 12.5     & 10.42    \\
Class 25 & 0     & 0    \\
Class 26 & 5     & 14.29    \\
Class 27 & 5.56     & 2.38    \\
Class 28 & 16.67     & 10.96    \\
Class 29 & 5.56     & 3.85    \\
Class 30 & 0     & 0    \\
Class 31 & 69.11     & 80.02    \\
\bottomrule
\end{tabular}
\caption{Precision recall score for SNaCK in the mturk experiment. Class 32 and 33 excluded as they are only supercategories but didn't manifest in the actual comments.}\label{tab:pr_table}
\end{table}


\begin{figure*}[t]
\centering
\includegraphics[width=\textwidth]{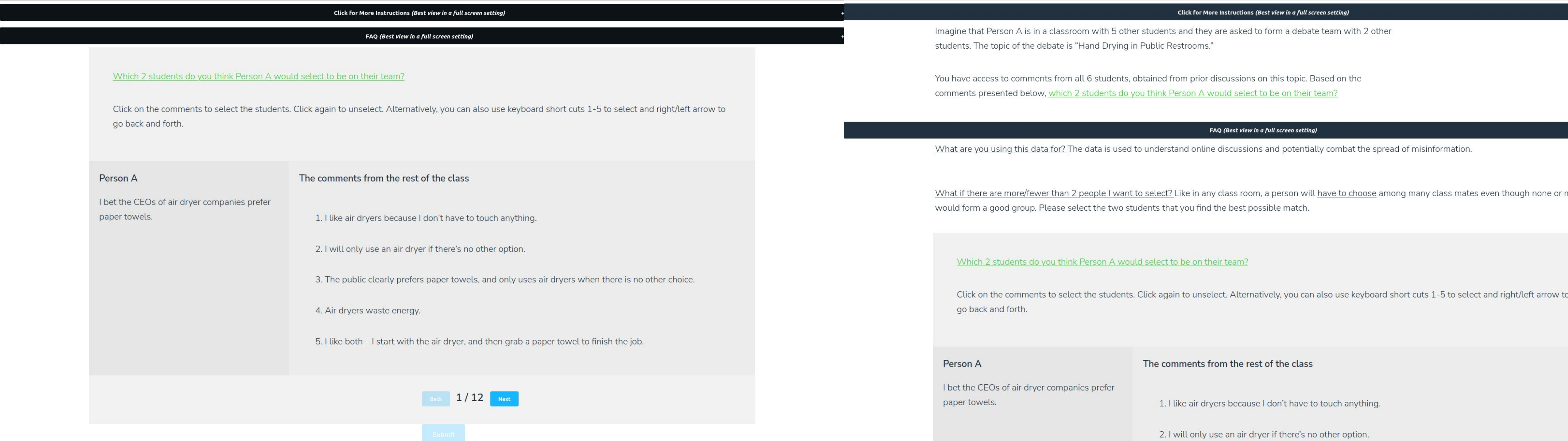}
\caption{Annotation interface. present a setting in which the workers imagine that they are high school students. The workers are presented We present a story to the workers to put them into the mindset of the imagined user who want to post the image presented. left: Main annotation page, with probe text and 5 texts displayed side-by-side. (b) Collapsible instruction on the top of the interface.}
    \label{fig:annotation_faq}
\end{figure*}

\section{Dataset Creation Details}
\label{supsec:data}

Given the inherent fine grained nature of narratives, one challenge we are facing is that how to collect reasonable labels in an effective manner. 

The authors first search for posts on reddit regarding \textit{hand dryer paper towels} on \url{http://www.pushshift.io} and read through the posts and select those that are related to discussions on hygiene in public restrooms.
Given the limited size of the dataset the authors read through all of the text and select the narrative of the sentences if they have any. \\

This approach is both time-consuming and highly dependent on the expertise of our annotators. However given the universality of the topic the authors read through all of the text and selected text fragments of comment which would appear at least twice in the dataset and that expresses a narrative within the specified taxon.

Given the sentences we adopt a \emph{game with a purpose} approach to keep annotators engaged and let them focus on the "compatability" of text pairs regarding a fictional debate team.

For our HIT we use relative similarity comparison in batch using a list or one column grid format following~\cite{Wilber_Kwak_Belongie_2014}.

The annotation task is to select which two sentences on the right would align with the statement to the left. Note that the resulting labels represent the \emph{perceived} shared narrative amount the sentences: the viewer's opinion of the shared narrative of the sentences. 
This section provide more details on the dataset acquisition process.






\subsection{Annotation interface}
\label{suppsusec: interface}
Similar to \cite{Jia2021IntentonomyAD,Horn2015,von2008recaptcha} we try to design our annotation approach to keep users engaged. 

We follow \cite{Jia2021IntentonomyAD} and design an interface that displays a probe sentence and a list of 5 sentences.

The Amazon Mechanical Turk workers are asked to select two sentences 
that would be in agreement with the probe statement on the left.

Notably the subject also encounters such a setup as a qualification test before processing to the actual HIT as shown in Figure \ref{fig:flow_dia}. 

In both the pre-test and the HIT is a splash screen, where the subject followed a 3-step procedure before starting the annotation as illustrated in Figure \ref{fig:splash_screen}:

\begin{itemize}
    \item \textit{Step 1}: The subject are introduced to the main idea of the task and are shown a comic to show the concept of the debate team and selection of students (Figure \ref{fig:splash_screen} A)
    \item \textit{Step 2}: Then, the subject is introduced to the notion of similar and different statements from different students, as well as an explanation as to why these are different.(Figure \ref{fig:splash_screen} B)
    \item \textit{Step 3}: Finally, the subject is being shown an example of an answer to a HIT using real data.(Figure \ref{fig:splash_screen} C)
\end{itemize}

\begin{center}
\includegraphics[width=\linewidth]{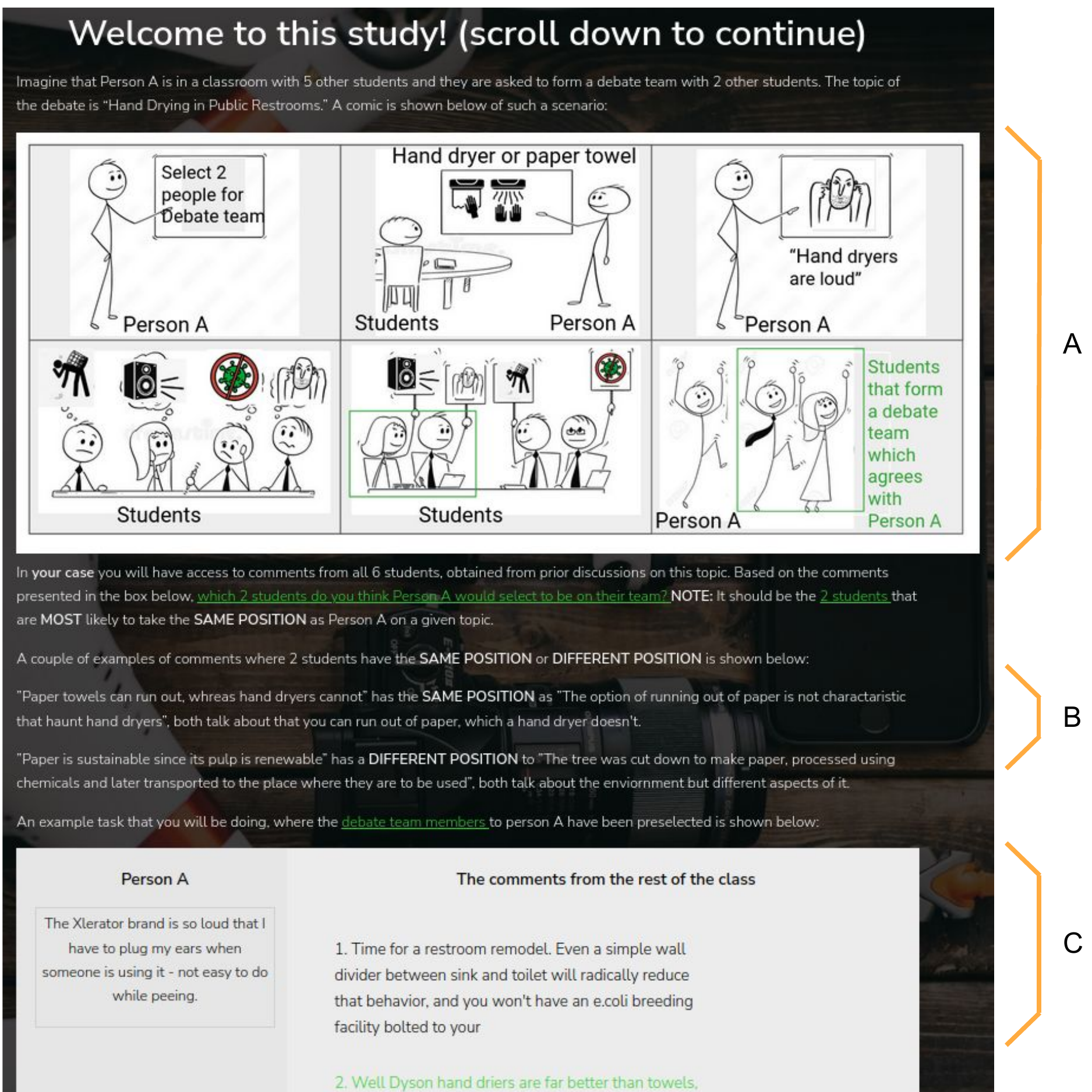}
\captionof{figure}{Splash screen. A:
subjects are shown a comic of student A making a statement and trying to find students who are likely to take the same side as student A. 
B: based on the above description and comic, textual examples of what makes statements similar and different are illustrated.
C: The actual HIT is finally shown with real data as well as an answer to which students student A should pick}\label{fig:splash_screen}
\end{center}

Continuing from (Figure \ref{fig:splash_screen} C) is the main annotation interface as shown in Figure \ref{fig:annotation_faq} .  This page has a collapsible section on top of the interface that display instructions.
The probe text on the left is always kept shown on the screen throughout scrolling up and down the page.

Since human motives are inherently abstract to understand, we provide a narrative, which is shown below, for the annotators so they could focus on the swapability of images.
The narrative presents a story for the workers, which bring them to the scenario of the imagined user who want to post the image presented on the left. 
We also provided example selections inside the collapsible instructions and the welcome splash page (see Fig.~\ref{fig:annotation_faq}(b)).

\begin{quote}
\small{Annotation narratives:}
\small{Imagine that Person A is in a classroom with 5 other students and they are asked to form a debate team with 2 other students. The topic of the debate is “Hand Drying in Public Restrooms. In your case you will have access to comments from all 6 students, obtained from prior discussions on this topic. Based on the comments presented in the box below, \textit{which 2 students do you think Person A would select to be on their team?}” NOTE: It should be the 2 students that are MOST likely to take the SAME POSITION as Person A on a given topic.}
\end{quote}

We used 5 sentences per text anchor, 12 grids per HITs, including 1 catch trials, and sometimes 1 sentinel example. We only use annotation results that pass at least 1 part of the catch trial.

\subsection{Sentence selection} 

\paragraph{Candidate sentences} 
Our goal is to fetch relevant sentences from online discussions forums like Reddit.
Each sentence comes from a subset of posts that are found using the pushshift search engine \url{http://www.pushshift.io}  on Reddit regarding the query \textit{hand dryer paper towels}, and have been read by the authors to ensure that we do not select any NSFW, banned, or quarantined subreddits.
Each sentences of pushshift has a list of associated keywords, produced by an online API \textit{praw}.
We use the above mentioned query to query posts from pushshift.
A total of $200$ posts were fetched using this query and 5 posts where selected based on their appropriate content.

\subsection{Annotators management}
\label{suppsubsec:anno_management}
To ensure quality, we restrict access to MTurks who pass our qualification task as shown as the first and second step in Figure \ref{fig:flow_dia}. Additionally we design two tasks to check the performance of the MTurks. The first is catch trials that is in a similar format to that of the qualification test. Second we create sentinel examples to check if the annotators agree on the narrative of the anchor.\\

Every annotator is obliged to take our qualification test in order to get access to our annotation task. The purposes of qualification test are two folds: firstly, to help us to select qualified workers who understand that we are creating debate teams based on similar viewpoints on the same topic. This setup is chosen specifically to avoid creating unintentional polarization by calling it narratives. Secondly, we wish to help workers get familiar with the content that they are likely to encounter during our annotation task.\\

We show a total of 5 questions during our qualification test to the potential annotator, as illustrated the first step in Figure \ref{fig:flow_dia}. Aside from passing at least 4 out of the five questions, it is necessary to have completed 100 HITs before as well as having a quality score of above 95 \%.
Additionally we carefully curate a triplet pair (a probe text and two options) where the answer is shown and an explanation given as to why these match.  We specifically selected texts that are within the same topic, such as environmentalism, but have different narratives.

\subsection{Sentinel example annotation}
We additionally inserted some sentinel examples for the MTurks to select, when asking them to identify similar sentence. The sentinel examples are inserted according to a poisson process and are always chosen to be the textual description of the anchor class. 
We found that MTurks were more likely to select the sentinel examples than the sentences belonging to the same class of the anchor.
This further demonstrate the MTurks are able to identify correct narratives using our \emph{game with a purpose} approach.
Yet in general MTurks tend to miss some of the labels.

\subsection{Catch trial and sentinel example statistics}

A total of 240 HITs were completed on our mturk campaign. Each Hits consists of 12 grids of which 1 was always a catch trial and 58 grids was assigned to our sentinel example experiments. The answers from these grids doesn't appear in our triplet training data. 

Of the 240 catch trials only 49.16 \% answers passes both catch trials in the grid and 90.82 \% passes at least one. All of the grids was answered by persons who passed the pre-test.

Similarly of the 58 grids that where assigned a sentinel example 44.82 \% of responses included the actual narrative, indicating that the workers doesn't always think that the crystallised narrative is most similar to the anchor and possibly that the anchor isn't always enough to decipher the actual narrative.

\section{Hygiene Taxonomy}
\label{supsec:taxon}

Table~\ref{supptab:taxon} lists the detailed taxonomy and explanation for each intent class.
We also note that these classes are always a subset of their 4 respective superclasses: \textit{For paper towel}, \textit{Against paper towel}, \textit{others} and \textit{irrelevant}.

\begin{table*}
\small
\begin{center}
\resizebox{0.98\textwidth}{!}{%
\begin{tabular}{ l  l }
\textbf{Class }      & \textbf{Descriptions}   \\
1&P: Paper towels have uses beyond drying hands \\ \hline\noalign{\smallskip}
2&P: Paper towels are more hygienic \\ \hline\noalign{\smallskip}
3&P: Paper towels can protect your hands when opening the door \\ \hline\noalign{\smallskip}
4&P: Air dryers circulate fecal matter throughout the bathroom \\ \hline\noalign{\smallskip}
5&P: Air dryers blows germs around the room \\ \hline\noalign{\smallskip}
6&P: Paper towels can wipe your hands clean \\ \hline\noalign{\smallskip}
7&P: Air dryers are loud \\ \hline\noalign{\smallskip}
8&P: Paper towels are better for the environment \\ \hline\noalign{\smallskip}
9&P: Air dryers waste energy \\ \hline\noalign{\smallskip}
10&P: Paper towels require less maintenance \\ \hline\noalign{\smallskip}
11&P: Air dryers can break down and take a long time to fix \\ \hline\noalign{\smallskip}
12&P: Paper towels are better than air dryers \\ \hline\noalign{\smallskip}
13&P: Hand dryers are pushed by Big AirBlade \\ \hline\noalign{\smallskip}
14&P: Paper towels are cheaper \\ \hline\noalign{\smallskip}
15&P: Air dryers take too long to dry your hands. \\ \hline\noalign{\smallskip}
16&H: Paper towels are pushed by Big Paper \\ \hline\noalign{\smallskip}
17&H: Air dryers require less maintenance \\ \hline\noalign{\smallskip}
18&H: Paper towels can run out \\ \hline\noalign{\smallskip}
19&H: Air dryers are more hygienic \\ \hline\noalign{\smallskip}
20&H: Air dryers are better for the environment \\ \hline\noalign{\smallskip}
21&H: Paper towels are waste of paper \\ \hline\noalign{\smallskip}
22&H: Air dryers are faster at drying your hands \\ \hline\noalign{\smallskip}
23&H: Air dryers are cheaper \\ \hline\noalign{\smallskip}
24&O: Air dryers and paper towels are equally hygienic if you wash you hands well \\ \hline\noalign{\smallskip}
25&O: Using your cloths to open the door prevents you from getting germs on your hands \\ \hline\noalign{\smallskip}
26&O: Hand sanitizers are just as good as towels or dryers \\ \hline\noalign{\smallskip}
27&O: Drying your hands using your pants is similar to using paper towels or air dryer \\ \hline\noalign{\smallskip}
28&O: DYSON hand dryers are better than other hand dryers or paper towels \\ \hline\noalign{\smallskip}
29&O: Wet hands are better air dryers \\ \hline\noalign{\smallskip}
30&O: The restroom door is filled with bacteria and one should avoid touching it \\ \hline\noalign{\smallskip}
31&N: Irrelevant \\ \hline\noalign{\smallskip}
32&H: Air dryers are better than paper towels \\ \hline\noalign{\smallskip}
33&O: Other narratives that appeared in the discussions \\ \hline\noalign{\smallskip}

\end{tabular}
}
\caption{The taxonomy for our \textsc{papyer}\includegraphics[height=1em]{images/papaya.png} dataset. P stands for paper towel narrative, H for Hand dryer narrative, O for Other narrative and N for No narrative. }
\label{supptab:taxon}
\end{center}
\end{table*}

\section{Dataset Analysis}
\label{supsec:data_analysis}

In this section, we analyze the properties of the dataset, as shown in table 8 and forward, in more detail.

\subsection{Dataset statistics}
Fig.~\ref{fig:circos_pos} shows the label distribution of whole training data, over 30 classes and 4 super categories.
It shows there is class imbalance in our dataset, which is the property of datasets in the real world~\cite{Horn2018b}.

\subsection{Lexical statistics}
We fetch the accompanying text description with the images found on the website. These descriptions are generated by a deep-learning based API and verified by human. 
We report the lexical (word-level) statistic of the dataset.
Specifically, the top words occurred in the descriptions of validation images are presented.
Table~\ref{supptab:lexical} shows frequent non-stopping words per class, shedding light on the properties of the images.
Although the descriptions can be heavily biased, Table~\ref{supptab:lexical} illustrates that, as they should, the occurrences of image objects and properties are relatively balanced across all the classes, indicating that most of the frequent words are not necessarily directly predictive of the intent label.

\begin{table*}
\scriptsize
\begin{center}
\resizebox{0.9\textwidth}{!}{%
\begin{tabular}{ l  l }
\toprule
\textbf{Class}       & \textbf{Sentences}   \\
\toprule
0  & Nah. The autoflush toilets can flush while you are sitting on them\\
\hline
1  &Drape a piece of toilet paper over the sensor and remove when finished' \\
             
\hline
                  
2  &Connoisseur de fÃ©calisation. \\

\hline

3  &i always do this, i hate when my ass gets sprayed with toilet water. why not just wire up a foot pedal you could tap to flush. \\
\hline
                
4  &Because sadly people have no aim of sense of right and wrong and it would get peed on constantly\\
\hline
                                
5  &And they wouldn't use it anyways. The point of the sensor is to flush when those people don't.\\
\hline
          
6  &The automated sinks never want to run water for me. I get so mad standing there with soap in my flailing hands trying to get it to run.\\
 \hline
         
7  &My boss (black) told me that they only see colored folks as he laughed exiting the bathroom \\
 & after easily washing his hands while I tried endlessly to wash my hands. \\
\hline

8  &hat was a good first day at work.\\
\hline

9  &man i miss this show \\
\hline

10 &HA! It all makes sense now. ;p That would be hilarious. \\
\hline

11  &I'm pretty partial to the floor switches for sinks\\
\hline

12  &I have no idea what that is. Is it like the pump sinks where you have to push on it?\\
\hline

13  &It's just a lever on the floor that turns the water on when you step on it. \\
\hline

14  &I used to work at a place that had those. It was great. \\
\hline

15  &You have to curse at it for a while. \\
\hline

16  &Aw damn. So you're telling me there is a password?\\
\hline

17  &Yeah, the password is 'FUCKING FUCK! GOD DAMMIT! *{kick}* FUCK! \\
\hline

18  &Yeah, but then he's going to touch the bathroom door and enter your place of work with you, \\
 & touching everything from keyboards to coffee pots to pizza. \\
\hline

19 &You can't avoid Ol' Nasty Joe \\
\hline

20  &I saw foot pedal operated faucets. That's awesome. \\
\hline

21  &It's to reduce the cleaning and stocking necessary. \\
\hline

22 &Basically, I hereby declare that no one is allowed to complain about the lack of attended toilets \\
 & unless they are seriously able to consider whether they would do toilet cleaning as a job themselves. \\
\hline

23  &While you're all planning your careers as astronauts and engineers - don't forget that you think toilet cleaning to be beneath you, \\
 & and that your talents would be wasted doing it, yet you want someone to do it \\
\hline

24  &Depends what type of dryers. Anything made before the Dyson Airblade was shit. \\
\hline

25 & The Dyson Airblade is still shit, it may "dry" your hands better than some traditional hand dryers, \\
 & but it's still very unsanitary. \\
\hline

26  &But I was responding to a comment about waste and restocking. \\
 & They save a lot of paper. There is a trade off I agree \\
\hline

27  &The trade off is the same as with traditional hand blowdriers, trading sanitation for paper waste \\
& (no one ever accounts for the wasted energy), but it's just not a justified trade off when the point is to clean your hands \\
\hline

28 &Airblades are just as unsanitary as traditional hand blowdriers. \\
\hline

29  &Older dryers are much less efficient, so the traders off is not the same \\
\hline

30  &I recall reading a study referenced in the book How Bad are Bananas regarding carbon and pollution in various product cycles \\ & that a paper towel was more efficient if you used just one, if you used more than one then a dryer would be more efficient. \\
& And this was taking into account the entire life cycle. \\
\hline

31 & if you used more than one then a dryer would be more efficient. And this was taking into account the entire life cycle. \\
\hline

32 & Since Dyson airblades are many more times efficient than existing dryers, then it's clear that using a Dyson is much more efficient\\
\hline

33 & Dyson claims 18 pairs of hands per one sheet of paper. And yes, we do have to factor in environmental aspects \\
& because we have limited and dwindling resources and we're pumping carbon into the atmosphere at frightening rate. \\
\hline

34 & At the end of the day we're always exposed to germs - and whether they are in the air or simply on the door handle when \\
& someone hasn't washed their hands at all, washing your hands frequently is the best technique fit limiting contact. \\
\hline

35 & You wrote that entire rant either intentionally without refuting my point that the efficiency is pointless without sanitation, or you didn't read my response. \\
\hline

36 & All of these articles everywhere are based on one study that Dyson refutes. They cite another study, which I provided links to above.\\
\hline

37 & Of **course** Dyson refutes them. Like Phillip-Morris did for decades with cigarettes. Always trust that less biased party.\\
\hline

38 & Is it more likely that this sole research group is trying to bring Big Airblade, or that Dyson is self-interested? Occam's Razor says number 2. \\
\hline

39 & Ok. But one study that's a bit sketchy is being pushed into every news source there is is. \\
\hline

40 & One could argue that it was promoted by a competitor. I've looked at the Dyson study, do you care to have a look at the other? \\
\hline

41 & Cause no one wants to be walked in on? \\
\hline

42 & .....not the stall doors \\
\hline

43 & I'd rather have a door that opens outward so I can push it open with my foot. \\
\hline

44 & I don't mind drying my hands with the dryers, but how the hell am I supposed to turn off the water \\
& and open the door to get out without a paper towel? \\
\hline

45 & I remember Nikki Sixx from Motley Crue talking about this exact thing on the radio last year. \\
 & Said he would throw paper towels on the ground near the door if there wasn't a trash can near.\\
\hline

46 & That's me. \\
\hline

47 & Health educator here. This is my biggest pet peeve about public bathrooms \\
& there needs to be more sanitary door opening systems. If everything else can be automatic... \\
\hline

48 & If they weren't 30 fucking dollars, I'd buy several dozen of them and install them everywhere I frequent free-of-charge. \\
\hline

49 & This is why the paper towels are replaced with hand dryers. \\
\hline

50 & As the person who used to have to clean your mess, but had no authority to make any change, \\
& yet was yelled at for having dirty bathrooms by other guests, fuck you. \\

\bottomrule
\end{tabular}
}
\caption{Processed textual dataset part 1 }
\label{supptab:lexical}
\end{center}
\end{table*}

\begin{table*}
\scriptsize
\begin{center}
\resizebox{0.9\textwidth}{!}{%
\begin{tabular}{ l  l }
\toprule
\textbf{Class}       & \textbf{Sentences}   \\
\toprule
51  & In my experience there wasn't really any place to put a trash can that wouldn't  \\
& block the door/entry way in some way, unless it was a small trash can like one you'd use in your car.\\
\hline

52 & In which case, it wouldn't have served its purpose long, would have needed to be checked/cleared multiple times during high \\
& traffic times, and so would have pulled manpower away from where we needed it in order to keep the bathrooms clean and safe \\
\hline

53 & I mentioned it to the manager to see if we could determine a solution \\
& and he said the same thing so no trash can was placed near the door since the value didn't outweigh the cost. \\
\hline

54 & So I didn't want to encourage using towels for drying hands, followed by getting more towels to open the door \\
\hline

55 & I'm glad I'm not the only one that does this \\
\hline

56 & Nurse (student) here. We usually use our elbows. If I'm alone and my hands \\
& are full I sometimes use my foot. Or keep paper napkins or something close \\
\hline

57 & I despise this! Now that it's getting to be winter, and the weather is a lot drier where I live, I get bloody noses wearily often. \\
\hline

58 & So when I'm bleeding all over myself, and I rush into a bathroom it pisses me \\
& off so much when there are no paper towels to staunch my bleeding face \\
\hline

59 & Uhhh toilet paper? \\
\hline

60 & Always an option, but it just seems more sanitary grabbing paper towels over tp. \\
\hline

61 & People grab pt after washing their hands, but people grab tp after touching their ass. \\
& Especially considering its an open wound and you're jamming it in a body orifice. \\
\hline

62 & Just pull it along the roll. People don't grab toilet paper and then carefully roll it back up again perfectly. \\
& You know when you get to an untouched portion. Don't be silly" \\
\hline

63 & But what if someone's wet poo soaked in from the side? \\
\hline
64 & Then don't use it. \\
\hline

65 & *But what if the gross germy poo creatures climbed deeper into the tp and spread their spores and reproduced get their icky \\
& ickiness all up in there and then I touch it and die?* This is how these people seem to me \\
\hline

66 & These assholes don't even realise that every time they touch a door handle \\
& they're exposing themselves in god knows what - and they are still surviving \\
\hline

67 & Well, due to the fact I'm in a college dorm and each bathroom has only two stalls, \\
& they're usually occupied and I don't exactly feel comfortable barging in to grab toilet paper \\
\hline

68 & Yup. I don't use them. My hands get shaken off and stay damp. \\
\hline

69 & I am usually wearing a dress that I don't really want to get wet. (I live in a hot climate and \\
& it's all dresses and occasionally Capri pants for me.) Plus then I have handprints on my dress. \\
\hline

70 & Maybe they're cream\\
\hline

71 & This is good advice.. Except if you wear khakis or other light pants like most of the population \\
& - then drying your hands with your pants makes you look like you pee'd yourself... \\
\hline

72 & Why not just use it? Why would you rather have wet hands? \\
\hline

73 & I don't really know. I just hate the way they make my hands feel so much.\\
\hline

74 & Because they'd rather have clean hands, that is the point of washing them, is it not?\\
\hline

75 & That's like saying "I don't flush, because it spreads germs. \\
& You know flushing blows shit all over the place, right?\\
\hline

76 & But who gives a shit? We've been doing it for years. You may \\
& get sick, you'll probably be fine. That's why we have immune systems \\
\hline

77 & Letting my hands dry normally for a whole 2 minutes isn't close \\
& to comparable with not flushing because the flush is not sanitary. \\
\hline 

78 & I flush where I can, and that's why you wash your hands after you flush. \\
& Taking reasonable steps, and avoiding blow dryers is a pretty reasonable step to take. \\
\hline

79 & I have strict views on handwashing due to my work, managing handwashing procedures for major food production facilities and \\
& studying food safety makes me anal for seemingly small details regarding handwashing, \\
\hline

80 & Makes sense. \\
\hline

81 & What you're suggesting is not reasonable by any stretch \\
\hline

82 & No, it's absolutely reasonable when it doesn't require more work, just education. \\
\hline

83 & It's pretty common knowledge that hand dryers in bathrooms are not sanitary. Are you arguing otherwise? \\
\hline

84 & Stop breathing air too, there might be germs.\\
\hline

85 & You're not very bright are you? I'm so far from a germaphobe, \\
& washing your hands doesn't make you a germaphobe, it makes you not disgusting. \\

\bottomrule
\end{tabular}
}
\caption{Processed textual dataset part 2}
\end{center}
\end{table*}

\begin{table*}
\scriptsize
\begin{center}
\resizebox{0.9\textwidth}{!}{%
\begin{tabular}{ l  l }
\toprule
\textbf{Class}       & \textbf{Sentences}   \\
\toprule
86  & But feel proud, you couldn't even be hired for an unskilled worker in one of facilities. That's an impressively low bar to go under. \\
\hline

87 & Typing so many words may injure your fingers. Also make sure you wear gloves, they keyboard may have germs too. Be safe, bro. \\
\hline

88 & Yup. Definitely unhireable. \\
\hline

89 & I have my own small business and a cushy consultant job where I do little while make good money.\\
\hline

90 &Guess I'll do just fine while you are too timid to even dry your hands \\
\hline

91 & I get paid to (in part) tell people smarter than you to wash their hands so companies can produce and prepare safe and wholesome foods\\
\hline

92 & Brags about being a consultant, is triggered someone says the word expert.\\
\hline

93 & Are you sure you're not the one with phobias? \\
\hline

94 & You are welcomed to make fun of me being a consultant. In fact, someone already did. \\
\hline

95 & Oh, and you're asking for basic interviewing techniques and advice 3 months ago. Nice business and consulting job you got there, congrats \\
\hline

96 & Huh? If you gonna troll, at least read ALL my history, and you'd know I have worked multiple jobs, \\
& started many businesses and I interview for EVERY job I get contacted for on LinkedIn.\\
\hline

97 & Try better next time. In fact, if you go far back enough, you'd see I have mentioned I sprinkle wrong info on all my reddit accounts to make it difficult to doxx me. \\
\hline

98 & Good try, guvernah. Try harder next time \\
\hline

99 & I don't want to doxx you, you're just a loser on Reddit who has a high percentage chance of having his own poop on his hands.\\
\hline

100 & And exactly the type of loser who doxxes people which is self evident in your doxxing paranoia to add to your other phobias. \\
\hline

101 & Glad you can be a tough guy and hurl insults on the internet to feel good about yourself. \\
\hline

102 & Even if I am the biggest loser, that won't make you a better person.And oh, if I knew how to doxx, that'd be great. I'll pay you if you can teach me \\
\hline

103 & Clearly you're really above it, starting the insults first because you were upset someone said the commonly known fact that bathroom blowdryers aren't sanitary. \\
\hline

104 & muh triggers Include hand towels, hands free of poop, yourself, and I'm sure much, much more \\
\hline

105 & Yeah 'cause that means I have to use my shirt to open the door instead of a paper towel \\
\hline

106 & Very much so. I avoid the public bathrooms at the building I work in cause they have hand dryers and the staff ones have paper towels \\
\hline
107 & If I can I also avoid public restrooms that I know only have hand dryers. Now if they have both, that's some dedication. \\
\hline
108 & I only like it when its a DYSON brand hand dryer,those things are awesome \\
\hline

109 & The airblades? Those are sweet. \\
\hline

110 & Yup,those airblades are fancy lookin \\
\hline

111 & Nothing worse than the cheap imitation brands tho. \\
\hline

112 & You can stand there for 5 minutes and your hands still wont be dry. \\
\hline

113 &  And it's always a shitty blow dryer that is on its last limb. \\
\hline

114 & I'm so old I remember when those dryers first became a thing. \\
\hline

115 & The equivalent of what would nowadays be a meme, the instruction tags were \\
& often modified: 1. Push button 2. Rub hands under warm air. 3. Wipe hands on pants. \\
\hline

116 & Receive bacon!!!! \\
\hline

117 & As someone who works in a place with hand dryers only. It's so us employees don't have to deal with assholes leaving paper towels everywhere. \\
\hline

118 & Our bathrooms also don't have doors and our faucets get wiped down regularly throughout the day. \\
\hline

119 & Yeah it sucks I've had to go and use the toilet seat cover paper to open doors cause of exactly this \\
\hline

120 & TIL I apparently don't care nearly as much about germs in bathrooms as the average redditor. \\
\hline

121 & The worst thing is that these places tend to have the way-underpowered hand dryers, like an asthmatic baby is blowing on your hands. \\
\hline

122 & Whereas, the places that actually do have paper towel dispensers seem to usually also have those \\
& awesome overpowered hand dryers. You're either getting nothing, or way too much. \\
\hline

123 & I guess the naysayers in this thread have never come across a Dyson Airblade. Better than paper towels \\
\hline

124 & Better than paper towels. False. \\
\hline

125 & I just use half a roll of their toilet paper to simulate paper towels. \\
\hline

126 & Not as good but it makes me feel better by punishing them for forcing my hand. \\
\hline

127 & That'll show 'em \\
\hline

128 & I hate it the other way around because 80\% of the time the paper is empty \\
\hline

129 & Haha! You should visit Japan. Public Restrooms often don't have anything to dry your hands with \\
\hline

130 & I find they are incredibly effective... at transferring any water on my hands directly onto my glasses. \\
\hline

131 & I just wipe my hands on my trousers and leave. \\
\hline

132 & I can never but my hands in those without touching sides, which defeats the purpose of washing \\
\hline

133 & Very much. I'd rather dry my hands on my clothing than use a hand dryer. \\
\hline

134 & It's OK when those really nice ones are there because I like those but any other hand dryer needs paper towels to match it. \\
\hline

135 & I get kinda annoyed, but then I think about the environmental benefits of not using tons of unneccessary paper and just wipe my hands on my pants. \\
\hline

136 & Paper is 100\% sustainable. Paper that's made today is made from young, plantation-grown trees that are re-planted after, just like food crops. \\
\hline

137 & It's actually pretty good for the environment compared to burning fossil fuels to run those hand dryers. \\
\hline

138 & Oh really? Shit, didn't know that. Thanks \\
\hline

139 & The high powered ones are okay but the old ones serve no purpose at all because the air coming out might as well be you blowing on your hands. \\
\hline

140 & Those ones with hand sanitizer only. \\
\hline

141 & Yeah, as a person with chronic sweaty palms this is more than mildly infuriating \\
\hline

142 & I use them to blow fresh air into my pants \\
\hline

143 & Ugh it's the worst. I can't stand not being able to completely dry my hands. \\
\hline

144 & here's only hand dryers on my campus minus a few spots here and there. But going back home to a bathroom with a wash towel is gonna be the best \\
\hline

145 & At my school, all the faucets are automatic, so I can't change the temperatures which are always blistering cold or scalding hot \\

\bottomrule
\end{tabular}
}
\caption{Processed textual dataset part 3}
\end{center}
\end{table*}

\begin{table*}
\scriptsize
\begin{center}
\resizebox{0.9\textwidth}{!}{%
\begin{tabular}{ l  l }
\toprule
\textbf{Class}       & \textbf{Sentences}   \\
\toprule
146  & This in combination with the door swinging into the restroom is my worst nightmare.\\
\hline

147 & Dark, enclosed, warm. Check . Bacteria loves a hand dryer.\\
\hline

148 &If it's not hot enough all you're doing is blowing bacteria over your hands. If anyone ever looks inside one. Yuck \\
\hline

149 & I HATE air driers. First, all they do is make my fingers freezing cold for several hours, while leaving them moist \\
\hline

150 & Second, soap helps get rid of germs by making them very slippery so you can wipe them off *en masse*. \\
& Without the paper towel, you're missing out on the germ removal part of the process." \\
\hline

151 & Using a paper towel is the fastest, most effective way to dry your hands and remove surface bacteria \\
\hline

152 & Third, I need a paper towel *anyway* to open the door, so I might as well dry my hands with it first. \\
\hline

153 & Fourth, the loud noise they make overloads me and freaks me out. \\
\hline

154 & OH MY GOD YES! \\
\hline

155 & I agree with you but doesn't data show that hot air hand dryers are better for sanitation? Or is that just a myth \\
\hline

156 & You are correct in the terms that it is a myth. \\
\hline

157 & On the subject of myths, Mythbusters made an episode and proved that these hand dryers simultaneously spread the harmful microbes into the air while drying your hands. \\
\hline

158 & Using towels also left less microbes on their hands after wash.\\
\hline

159 & Well then. Now I will be even more frustrated when I don't see paper towels \\
\hline

160 & Have you ever seen that gross cloth towel system that is always nasty/dirty in some public restrooms? \\
\hline

161 & I've never seen the cloth towel thing anywhere other than high-end locations, where there's a nice clean pile and then a fancy laundry basket for the used ones. \\
\hline

162 & No. this is like a circle of cloth that is partially inside a dispensing box. I searched "cloth public washroom dispenser"...you will see it. \\
\hline

163 &Huh, interesting. I've never seen those. That seems pretty gross \\
\hline

164 & Very. Very gross. \\
\hline

165 & Here at Eastern Europe have I never seen such things, at public premises we always have the disposable kind of towels. \\
\hline

166 & The cloth towel thing isn't actually a circle. It's a big roll that feeds out into the space you use and back onto an empty roll.\\
\hline

167 & It's like a film reel. I do agree that it's still gross, though. I don't know how clean that cloth \\
& was when they loaded it, unless it's obviously new. (Though it never is.) \\
\hline

168 & Nursing student here: hand dryers are nasty nasty. \\
\hline

169 & This makes more sense than what I thought I previously heard where the dryer spews \\
& bacteria out on your hands, it's not like what comes off your hands is going back into the dryer \\
\hline

170 & Yes, with the rare exception of those dryers that you put your hands *inside* and move them up and down. Those are cool. \\
\hline

171 & No, they aren't. Sometimes called Airblade which blows the water up my sleeves. Also who names something I stick my hands in "*blade"? \\
\hline

172 & 'what about wiping them on my pants? \\
\hline

173 & 'Your pants are a conveniently warm, moist environment for me to wipe my hands on.\\
\hline

174 & "sounds hot, what\'s your number?\\
\hline

175 & 'Taiwan, is that you? \\
\hline

176 & 'Taiwan numbah wan! \\
\hline

177 & 'He was number one... \\
\hline 

178 & '867-5309 Pls rspnd. \\
\hline

179 & 'I got it  \\
\hline

180 &'I got your number on the wall" \\
\hline

181 & 867-5309 \\
\hline

182 & 'My number is "No".\\
\hline

183 & '>to wipe my hands in. FTFY \\
\hline

184 & 'But you have to stick your hands *inside*.\\
\hline

185 & 'Depends on how clean your pants are.  \\
\hline

186 & 'As prescribed, I only wash my jeans once a month or so... \\
\hline

187 & 'Jeans dont actually need to be washed, unless you actually get something on them, in which case you actually want \\
& to spot clean the area than the whole pants. But then I noticed you said "as prescribed" so gg \\
\hline

188 & 'Woosh \\
\hline

189 & 'Oh, I just reply to random comments with "woosh" sometimes. \\
\hline

190 & 'I thought you just had a funny way of telling me to wash them... ;-) \\
\hline

191 & What the fuck, Chris? \\
\hline

192 & What about not pissing on my hands? \\
\hline

193 & It says in the article that that\'s a viable option. \\
\hline

194 & I agree, we should all clean our hands by wiping them on c4stiel's pants \\
\hline

195 & It is quite clear that Reddit is full of wise asses like me. Someone else always posts my idea first. Good to be among like minded friends. \\
\hline

196 & Wasnt the study that discovered that funded by Scott towels? \\
\hline

197 & Next you're gonna tell me Brawny squanches their customers with subliminal propaganda! \\
\hline

198 & They just look in their squanch and squanches whatever they're squanching \\
\hline

199 & I love it \\
\hline

200 & Holy fuck, my high school history teacher had a poster of this on her wall. Always cracked me up. \\
\hline

201 & I've got it in my bathroom. \\
\hline

202 & The stuff left on your hands after either air-drying or towel-drying is the harmless stuff that\'s found everywhere on your skin anyway. \\
 & The point of air dryers was never to reduce the bacterial count on your hands. It was to reduce the use (and waste) of paper. \\
 \hline
 
203 & I was looking for this. Who gives a shit about bacteria on your hands? Wash your hands then \\
& touch your face; congratulations, you've got a ton of bacteria on your hands, and it doesn't matter. \\
 \hline
 
204 & The bacteria floating around a bathroom air dryer are the same type on your face? \\
 \hline
 
205 & Yes. Fecal bacteria are everywhere, including all over you, right now. Hand-washing is about parasites and viruses,  \\
&which are not hiding inside bathroom air dryers.   Oh, and I guess if you actually got poop on your hands you'd want to wash that off too. \\
& Quantity matters with bacteria; a poop smear is kind of a big deal, but not imperceptible bits flying through the air. \\

\bottomrule
\end{tabular}
}
\caption{Processed textual dataset part 4}
\end{center}
\end{table*}

\begin{table*}
\scriptsize
\begin{center}
\resizebox{0.9\textwidth}{!}{%
\begin{tabular}{ l  l }
\toprule
\textbf{Class}       & \textbf{Sentences}   \\
\toprule
206  & >a poop smear is kind of a big deal Not to be confused with a pap smear"\\
\hline

207 & The FDA cares a LOT. \\
\hline

208 & While mostly true, this comment is misleading to the extent it suggests that soap specifically targets "problematic" bacteria while leaving behind harmless bacteria. \\
\hline

209 & It's possible, but there's also the benefit of literally removing any potential left over dirt with a towel (or any medium really) vs a dryer. That's the real benefit. \\
\hline

210 & Mythbusters also did an episode on this, and they got similar results. I don\'t remember if their numbers were the same though \\
\hline

211 & They've found that 1) it will matter if you wash hands without soap. Soap which they used killed everything. \\
& 2) both dryer and towel reduced number of bacteria, but towel did much better. Dryer - by 23\%, towel by 71\% \\
\hline

212 & Jamie noted that when people used hand dryer, their hands were more wet comparing to paper towel.\\
\hline

213 & Would anyone actually ever opt to use the hand dryer over using paper towels if they're available? Shit takes forever and they usually aren't timed long enough.\\
\hline

214 & The Xlerator brand is so loud that I have to plug my ears when someone is using it - not easy to do while peeing.\\
\hline

215 & YES. Those things are louder than the machinery rooms on USN warships. \\
\hline

216 & When I have to use them, I'm constantly getting my hands close enough to trip the sensor, immediately moving them like two feet away, \\
& then they shut off and I move my hands back up to trip the sensor again and then immediately back down again, etc.  Fucking pain in the ass.
 \\
\hline

217 & Totally agree on them being obnoxious even out in the dining area, too. There are two tables in one of my \\
& favorite local joints that I refuse to be seated at, because of those dryers.'
\\
\hline

218 & They are loud in the afternoon as well. \\
\hline

219 & Depends if you're using one of the shitty old-style dryers, or one of the \\
& new high-powered ones that dry your hands in seconds. And people opt to use them because they reduce paper waste.\\
\hline

220 & Looking at you, fucking World-Dryer. \\
\hline

221 & I like those. I think they're called Air Blades or something.  \\
\hline

222 & depends on whether you're using one of the old-style dryers, or one of the new ones that generates \\
& literally ear-destroying decibel levels if you get your hands within 18 inches of the vent"
 \\
\hline

223 & ***FUCK*** those things. They're so damn loud even sitting at a restaurant table NEAR the bathroom is irritating as hell.\\
\hline

224 & I always use the dryer because I love the way my hands feel after. They get all smooth feeling. You have to rub your hands together and wait til they're \\
& actually completely dry. So nice. For a long time I never used the dryers but then one day I discovered what I'd been missing out on and never went back. \\
\hline

225 & I don't use them even if there aren't paper towels. I'll go back to a stall and use a bunch of TP instead. Those air dryers are damn near useless and a waste of time.\\
\hline

226 & You sound like you are too dumb to make it through life.\\
\hline

227 & 're dumb", said "/u/IdontReadArticles".\\
\hline
228 & Well that's gross. Those poor people who only use hand drier to save the environment are getting a ton more bacteria on their hands haha \\
\hline

229 & Obviously Scott towels paid them off too \\
\hline

230 & Fucking Big Paper Towel eh?\\
\hline

231 & Paper companies are huge. GP, for instance, is also owned by the EVIL Koch Brothers.... Scary!\\
\hline

232 & Even if so, that doesn't make it wrong. It just means we have to pay extra attention to their methods and rigor and \\
& preferably have someone else repeat the study (which should be done anyway). Conflicts of interest to not disqualify science.  \\
\hline

233 & It does not, but we don't know if they did other studies that showed results that did not cater to their marketing needs. \\
& Pharmaceutical companies pull that shit all the time, they hide unfavorable studies from publication and put gag orders on scientists in their employ. \\
\hline

234 & >Conflicts of interest to not disqualify science. No, but its reasonable to be more skeptical when its industry funded.\\
\hline

235 & Which is exactly what /u/sixblackgeese said. \\
\hline

236 & The person you're responding to already said that...>It just means we have to pay extra \\
& attention to their methods and rigor and preferably have someone else repeat the study'
 \\
\hline

237 & That was exactly his point. \\
\hline

238 & It's necessary in all situations, to varying degrees.\\
\hline

239 & Also it's interesting to note I have never seen an air dryer in the hospital I was in for a few weeks. In fact when certain procedures were \\
& being done they didn't want a lot of air moving around so the bacteria wouldn't collect on my newly disinfected skin. \\
\hline

240 & > Conflicts of interest to not disqualify science.True, but in the modern world where most people don't have time, \\
& inclination, or knowledge to read science papers and thoroughly disprove them, conflicts of interest like this should be red flags that allow a layperson to make a quick \\
& assessment of the likelihood that the research is solid. In this case such a strong conflict of interest should make people very very wary of this research.  \\
\hline

241 & My brother did an experiment in middle school where they tested bacterial cultures from washed hands 1. with soap 2. without soap 3. dried with paper towels \\
& 4. air dried and the paper towel dried hand samples cultured fewer bacteria. Yes, its anecdotal and it's middle school science but I tend to believe it. \\
\hline

242 & It's a good question, I sort of remember the biggest bacterial difference was based on drying method - \\
& not washing method. It probably depends on the type of soap and how it is dispensed though. \\
\hline

243 & Not Scott towels, but an equally questionable source. There's no research that proves things either way, but most hand dryer manufacturers don't hire idiots \\
& to engineer equipment, so I suspect their internal surfaces are rather inhospitable to any bacteria (copper alloys, super hot coils, etc). \\
\hline

244 & Studies funded by corporations always have an obvious agenda and obviously want the results to favour them. \\
& So they're going to have bias and potential skew the data. I don't trust sponsored studies. Too many bad angles. \\
\hline

245 & Were you expecting it to be funded *and then published* by hand drier manufacturers? \\
& The studies that get published are published by people with vested interest in the study's conclusion. \\
\hline

246 & And yet the observation is both valid and intuitive \\
\hline

247 & Do you have real life hulk hands? \\
\hline

248 & Child of Atom! \\
\hline

249 & It's rare that I laugh while browsing but this got me pretty good. I admire your enthusiasm. \\
\hline

250 & Well I thought it was a funny not meant to be taking serious jab.  \\

\bottomrule
\end{tabular}
}
\caption{Processed textual dataset part 5}
\end{center}
\end{table*}

\begin{table*}
\scriptsize
\begin{center}
\resizebox{0.9\textwidth}{!}{%
\begin{tabular}{ l  l }
\toprule
\textbf{Class}       & \textbf{Sentences}   \\
\toprule
251  & It's neither..\\
\hline

252 & If that comment of his was meant to be serious he's an asshole, If it's not it's just a poor joke Thats slightly offensive. \\
\hline

253 & Nah. It's just dumb. If anyone is offended by something that dumb... They, themselves, are fucking morons.\\
\hline

254 & Am I allowed to be offended for someone else? \\
\hline

255 & That's a useless exercise. How do you know they will be offended? Why can't they be offended or not offended themselves? \\
\hline

256 & Some people are too privileged to realize when they should be offended. only kidding don't hurt me\\
\hline

257 &  Shrug emoji \\
\hline

258 & I just pee on my hands and leave. \\
\hline

259 & Not bad.. you go to the bathroom and you relieve yourself and clean your hands all in the same step. \\
\hline

260 & You could just shove your hands in your pants and let wee-wee run free wherever you happen to be. \\
& Clean hands, clean pants, clean floor, happy bladder. Do it around engineers and they'll applaud your ingenuity. \\
\hline

261 & I bet they'll want to shake my hand in respect \\
\hline

262 & I am not an expert by any means but I do believe this stop's the burning. \\
\hline

263 & This stop is the burning? \\
\hline
264 & Must be from California \\
\hline

265 & At UCLA, they teach us to wash our hands after we piss. At USC, we just piss on our hands.\\
\hline

266 & Why is the dryer a moist environment? \\
\hline

267 & Tests have shown that you'll have more bacteria on your hands from using the air dryer because you don't dry your hands as well as you would a paper towel \\
\hline

268 & Glad I'm not the only one leaving the bathroom while cherishing the taste of the air. \\
\hline

269 & How bad can it be though? It allowed Peter Pan to fly!\\
\hline

270 & faith, trust, and fecal dust\\
\hline

271 & Tap tap tap a Tinker Bell and out comes magic fecal dust! \\
\hline

272 & ...and tongue.  LMAO \\
\hline

273 & You feel self conscious because it's been 30 seconds and your palm is still wet and there's 5 guys waiting in line.\\
\hline

274 & Bathroom air is probably moist.\\
\hline

275 & Because people put their wet hands near it all day \\
\hline

276 & Still wouldn\'t make it moist.  \\
\hline

277 & At least it would make the air directly under your hands moist. And that's the air that it sucks in to blow out. \\
& A family friend works for a air dryer company and showed if to me. If you look into some of them they are disgusting \\
\hline 

278 & Well that makes sense.  Thanks for teaching me something.\\
\hline

279 & Most of the time they blow hot air which can contain more water vapor. \\
\hline

280 & But the bacteria that grow in a conveniently warm, moist environment are probably soft and weak. \\
& Their idea of "roughing it" is probably going without tea and crumpets. So it's easy for your immune system to take them out.\\
\hline

281 & Now on the other hand, any bacteria strong enough to thrive in dry paper towels are probably going to be bad-ass, \\
& Spetznaz-type bacteria that can do a leaping somersault hatchet throw and kick your white blood cells' asses"
 \\
\hline

282 & I'll take my chances with the blower \\
\hline

283 & This sounds wrong but I don\'t know enough about bacteria to refute it \\
\hline

284 & I know bacteria breeds when conditions are good, and can go dormant when they're not. I'm not entirely sure the types of bacteria, and their constitutions, will be different. \\
& It could be the exact same bacteria on the paper towels breeding in the dryer. I could be wrong, but I have no way of knowing.\\
\hline

285 & Guess you'll need to start avoiding those commie napkins as well. \\
\hline

286 & I prefer using hand towels to dry my hands mostly because I can use the towels to turn off the faucets and open doors without getting even more bacteria on my hands. \\
\hline

287 & Hand dryer bacteria = British Paper towel bacteria = Russian Special Forces Got it edit:yayalorde made me learn something \\
\hline

288 & Spetznas is russkie special forces \\
\hline

289 &TIL, I will fix it.\\
\hline

290 & Dont worry brother. Seal Team Six is hiding on the soap dispenser and is antibiotic resistant \\
\hline

291 & What an awesome analogy. The hand sanitizer dispenser must be like the US Navy Seals approaching on speedboats in the middle of the night. \\
\hline

292 & I  feel like the spetznaz bacteria would also survive in the warm moist environment, so you'd be getting both, rather than just the weaker one. \\
\hline

293 & This reminds me of the Buffalo Theory. \\
\hline

294 & But in most cases **it doesn't matter**, because **our immune systems are resilient** enough to keep the numbers of these bacteria low. \\
& Furthermore, **the overwhelming majority of these bugs are fairly harmless.** They don't usually include the real nasties like salmonella, shigella, campylobacter, \\
& hepatitis A or B, the SARS coronavirus, or the virus that causes meningococcal disease." \\
\hline

295 & Thank you.  I was also wondering if you've already reduced the bacteria on your hands by washing them properly then adding \\
& 100 bacteria to the 100 that made it through the washing part would be doubling the bacteria, but it's not worth worrying about.
\\
\hline

296 & And yet noro virus is often best transmitted by fecal matter, which could have been easily cleaned with a \\
& good wash of the hands. Doctors didn't start washing their hands for no reason, it's because it saves lives. ' \\
\hline

297 & Washing your hands is what cleans them.  How you dry them is a distant second.  That\'s all I was trying to say. \\
\hline

298 & What about the newer ones that dry your hands in ten seconds? This says it takes 50... \\
& Also ours at work also have some UV nonsense built in, I assume that is supposed to help.. \\
\hline

299 & Supposedly instead of increasing bacterial counts by 255\% it increases them merely by 100\%.  \\
& So, much better that any hand-dryer, but still significantly worse than paper towels. \\
\hline

300 & According to a towel-manufacturer sponsored study \\

\bottomrule
\end{tabular}
}
\caption{Processed textual dataset part 6}
\end{center}
\end{table*}

\begin{table*}
\scriptsize
\begin{center}
\resizebox{0.9\textwidth}{!}{%
\begin{tabular}{ l  l }
\toprule
\textbf{Class}       & \textbf{Sentences}   \\
\toprule
301  & 'I found the paper towel. \\
\hline

302 & Someone should just make a paper towel that blows.\\
\hline

303 & He says in there that he finds jet hand dryers hard to believe. He also handily pointed out that they make 'enormous noise'. Sounds like good sciencing to me\\
\hline

304 & It's not the quantity of the bacteria that matters, it's what kind of bacteria it is. All bacteria are not harmful and many are very beneficial to us. \\
\hline

305 & conveniently warm, moist environment" = the edges of the Dyson Hand Dryers that \\
& every single person is bumping their dripping, wet, bacteria-coated hands against as they move them clumsily up and down.
\\
\hline

306 & The ones they 'tested' in this study were not the newer dyson dryers that take ~10 seconds, but older dryers that take almost a full minutes to dry. \\
\hline

307 & Take 10 seconds? I guess until I give up and just wipe my hands on my shirt.\\
\hline

308 & They don't take 10 seconds. They get the first bit of water off quicker, but after that they're \\
& arguably worse, since you can't spread out the water by rubbing your hands together.Regardless, death to hand blow dryers.\\
\hline

309 & Look what's sitting at the bottom of that Dyson dryer....bacteria soup\\
\hline

310 & WELCOME TO THE REAL WORLD \\
\hline

311 & Sanitary on a macro level. No over flowing trashcans etc. \\
\hline

312 & I used to think washing my hands in really hot water would help.  Turns out they can take water hotter \\
& than you can for longer than you can. You can handle soap much better than them though. \\
\hline

313 & Her mouth is probably dirtier than the dryer. So she's boosting the dryer's immune system. \\
\hline

314 & Silly, there are no libraries on reddit. \\
\hline

315 & Came into the comments to ask who the savages are that dry their hands on their hair. I'll leave now\\
\hline

316 & I work in construction and regulatory have regularly swap out or move electric hand dryers and paper towel dispensers. \\
& Honestly, they are both always pretty nasty with all types shit growing on the back, inside and hard to clear areas. \\
& If you are so worried about drying your hands I would suggest that you bring your own paper towels and avoid either of the public drying options.   \\
\hline

317 & The energy necessary to actually dry your hands with one of those stupid hand dryers also HAS TO exceed the amount required to manufacture one paper napkin... \\
\hline

318 & You'd also have to factor in shipping and disposal costs.\\
\hline

319 & Well, that's literally the opposite of the case that's made about them. They are marketed as and purchased as a more sustainable alternative to paper towels.\\
\hline

320 & In the long term, yeah, it is more sustainable.\\
\hline

321 & I'm absolutely positive that it's cheaper only because people get frustrated and leave with damp hands. \\
\hline

322 & I've tested this, it takes nearly a minute to get your hands as dry as a paper towel, and there's no way it's cheaper to run that than to just have a paper towel. \\
\hline

323 &You are not using them properly. \\
\hline

324 & Wood pulp contains cellulose, wood fibers, and lignin, biological glue holding the cellulose together. They are separated in pulp mills and the cellulose is used to make the paper towel \\
& and the lignin is burned in a recovery boiler to generate power. The pulp, which is a renewable resource, literally generates the power to turn itself into paper. \\
\hline

325 & The blower on the other hand, uses utility power (mostly coal) to run the heater and the blower.\\
\hline

326 & The paper towel is much better for the environment.\\
\hline

327 & Know of anywhere I can read some stats about this?\\
\hline

328 & The dryers that don't have a heater do have a smaller carbon footprint than paper towels. \\
& But with the bacteria they blow on your hands, you're better off just not drying them anyway.\\
\hline

329 & I don't remember any details, but I read somewhere that the big problem with paper towels (and toilet paper, naturally) \\
& is that they (somehow) take an obscene amount of water to make. \\
\hline

330 & As long as hand dryer bacteria are friendly, they are welcome to join my other trillion or so bacteria already living on me.\\
\hline

331 & Is this why I'm the only one using them? I see people just walk out without drying if there is no paper towels.\\
\hline

332 & Probably because it takes too damn long \\
\hline

333 & I went to a new movie theater in town a few days ago and they have unheated hand dryers with really high pressure. \\
& They worked really well and it only took a few seconds. First time I ever seen them.\\
\hline

334 & Who cares as long as my hands are dry.\\
\hline

335 & Confirmed on Mythbusters. \\
\hline

336 & Well Dyson hand driers are far better than towels, because the sound makes the bacteria deaf and they become harmless. \\
\hline

337 & Mythbusters proved this all false. As long as you use soap there's no bacteria to blow around.  \\
\hline

338 & And this is why I'd rather dry my hands on my own cloths than use one of these things. A huge wad of toilet paper can work as well.\\
\hline

339 & instructions unclear penis stuck in zip.\\
\hline

340 & How to think you're making a difference, while making literally no important difference.  \\
\hline

341 & This is old news now, would be interesting what the results from the newer dyson jet dryers are like. \\
\hline

342 & I have them at work but yep to come up with suitable experiment.\\
\hline

343 & Even beyond the germophobe reasons, I hate not being able to wipe my hands. What if I get something sticky on them? \\
& You can't just rinse it off a lot of times; you need wiping action. \\
\hline

344 & I hate places that only use air dryers, lke they do it for show to pretend they're so environmentally conscious, \\
& but I bet they just want to save a buck not buying paper towels. \\
\hline

345 & I mean, saving paper is saving paper. There's certainly convenience in a paper towel, but air dryers are the way to go moving forward.\\
\hline

346 & Saving a business or school money on paper towels is a plus but it'll certainly take a while for those savings to catch up to the cost of a single hand dryer.\\
\hline

347 & Sustainability is a driving factor for a lot of groups, especially in low traffic areas.\\
\hline

348 & People forget that much of the waste and such generated in the ~~wastern~~ western world originates from public health concerns. \\
\hline

349 & wastern Paging Dr. Freud \\
\hline

350 & Gah! Frickin' mobile typing. I'll fix it.  \\

\bottomrule
\end{tabular}
}
\caption{Processed textual dataset part 7}
\end{center}
\end{table*}

\begin{table*}
\scriptsize
\begin{center}
\resizebox{0.9\textwidth}{!}{%
\begin{tabular}{ l  l }
\toprule
\textbf{Class}       & \textbf{Sentences}   \\
\toprule
351  & Wood pulp contains cellulose (wood fibers) and lignin (biological glue holding the cellulose together). They are separated in pulp mills and the cellulose is used \\
&to make the paper towel and the lignin is burned in a recovery boiler to generate power. The pulp, which is a renewable resource, literally generates the power to turn itself into paper. \\
& The blower on the other hand, uses utility power (mostly coal) to run the heater and the blower. The paper towel is much better for the environment."
 \\
\hline

352 & Any towel manufacturer is going to use outside power as well, which would contain just as \\
& much coal power as anything else in the region. But bringing up coal power is a bit misleading in itself.\\
\hline

353 & lf. The college I work for just made a pretty sweeping change to air dryers, only after making an almost 100\% switch to solar power.\\
\hline

354 & I type paper towels vs air dryers into google and have a seriously difficult time finding anyone willing \\
& to claim or back up a claim that paper towels are somehow better for the environment in the short or long term. \\
\hline

355 & for the next 20+ years, the majority of the power used for air dryers will be from from non renewable resources.\\
\hline

356 & The dryers that don't have a heater do have a smaller carbon footprint than paper towels. \\
\hline

357 & But with the bacteria they blow on your hands, you're better off just not drying them anyway.\\
\hline

358 & The article you just posted compares the dyson airblade to **one single sheet** of *recycled* paper towel \\
& and finds the airblade has 1/5th the carbon footprint. That's extremely significant.\\
\hline

359 & And, because it's making a comparison between air dryers and a single paper towel, that same article is also saying that old \\
& style air dryers are *as* efficient as **2** sheets of paper towels and ***more*** efficient than using 3 paper towel sheets.\\
\hline

360 & I don't know many single paper towel users in this world, and with that in mind, I feel even more confident saying *all* air dryers \\
& are a more sustainable alternative to paper towels. And that is taking into account the environmental cost of energy production. \\
\hline

361 & Number one, you're better of not washing your hands at all than washing them and drying with an air dryer. \\
\hline

362 & If air dryers could even get to close to the level of sanitation of paper towels you could make an argument to install more heaterless air dryers. \\
\hline

363 & And secondly, not all air dryers are dyson airblade models. The majority of them do have heaters, \\
& and they are worse for the environment than using paper towels, even if you need two of them \\
\hline

364 & Number two, now we've come full circle. The two main studies that have made this claim were funded by a company named Scott \\
& (you may be aware of the product they sell) and a group called the European Tissue Symposium (can't imagine what interests they represent). \\
\hline

365 & Now, as I mentioned in my previous comment, old style heated dryers are absolutely unsanitary and plenty of studies confirm that. I personally choose not to use them.\\
\hline

366 & But, specifically, Dyson Airblades are NSF (National Sanitation Foundation) certified as sanitary, and are in fact the only air dryers with this certification.\\
& Most all industrial paper towel dispensers have the same certification. This is the same certification required for all equipment used in an industrial kitchen. \\
\hline

367 & Two recycled paper towels and the air dryer have the same carbon footprint. Paper towels are a renewable resource, \\
& whereas the energy generated for the air dryer is not (mostly).\\
\hline

368 & It's the renewable resource that makes paper better for the environment.\\
\hline

369 & Did you know virgin fiber is better for the environment than recycled?\\
\hline

370 & The trees grown reduce carbon dioxide and the lignin offsets the power to produce the paper.\\
\hline

371 & So what then is the difference between a Dyson airblade and an old fashioned heated dryer as \\
& far as sanitation is concerned? Nothing. They both increase the bacterial count on your hands. \\
\hline

372 & he dyson air dryer and it's equivalent are the most environmentally efficient hand drying method. Period. \\
\hline

373 & Where are your sources? We are commenting on a study that very well documents how unsanitary air dryers are.\\
\hline

374 & Cite sources showing air dryers are hygienically comparable to towels please? Towels are the winner here. \\
\hline

375 & If the air dryer increases the amount of bacteria on your hands it is not even comparable to hand towels. \\
\hline

376 & Carbon footprints are independent of the renewability (did I just make up a word?) of a resource. \\
& Pulp, recycled or not, is renewable and the majority of industrial power generation are not. \\
\hline

377 & Air dryers are not an environmental friendly alternative to paper towels because they are not an alternative at all. They literally worse than not washing your hands. \\
\hline

378 & Sure, I'd love to repeat the same arguments over and over again and have you not respond to a single one of them. Before I begin, I have to \\
& mention how much I love that "Where are your sources" has turned into "do my research for me" or "prove your point with sources so I don't have to do the same for mine." \\
& You have yet to refute anything I'm telling you and yet you keep spouting bullshit. The "best" alternative for the environment is to not dry your hands at all, \\
& but that isn't the world we live in and wet hands are considerably less hygienic (something you seemed to be very concerned with) as a damp breeding ground for bacteria. \\
& I am starting to understand your perspective, though. Once again, you are looking to the study we are commenting on as gospel \\
& (You remember, that same one we agreed was paid for by Scott paper towel company). So I was completely unsurprised when, after a few minutes of digging an... \\
\hline

379 & wait... soap disgusts you? huh? \\
\hline

380 & It's slimy and it smells bad. Especially liquid/foamy soap. I'm fine with a bar in the shower; usually the smell isn't too offensive. \\
& But I hate the idea of my hand smelling soapy for hours. I avoid it when I can. \\
\hline

381 & Gross. \\
\hline

382 & i think you may actually be a dog \\
\hline

383 & I hate dogs more than I hate soap. \\
\hline

384 & I've been downvoted for saying I don't use those gross hand driers for that reason. I go with dripping hands if there's no paper towels. Who's laughing now?\\
\hline

385 & My hands are covered with skin, not cotton. They dry in less than a minute on their own. Dripping hands is what I go with as well. \\
\hline

386 & Not to mention the old pants/shirt wipe expedites the process. \\
\hline

387 & I just wash my dick in the morning, so I don't have to wash my hands.  \\
\hline

388 & Wonder how many hand dryers are in hospitals.\\
\hline

389 & I have never seen one in a hospital around here. Always paper towels.\\
\hline

390 & Most new hospitals now have built in faucets in every room for the nurses to wash their hands, dry with towels, sanitize, and \\ 
& wear gloves before touching the patient. I've never seen one with the hand dryers where it mattered, I have seen one in the public bathrooms though.  \\
\hline

391 & The bathrooms on the other floors with patient care areas have hand towels. Nearly all staff hate using the bathrooms with the dryers because it just feels wrong. \\
\hline

392 & I guess if we're going to go the most germless route, the real question is, "with what do surgeons dry their hands?\\
\hline

393 & Having OR experience, the surgeons I've assisted leave their hands wet or use a sterile towel and toss it away prior to putting on the sterile gloves. \\
\hline

394 & I hate using the cotton paper to wipe my hands, it always breaks off and becomes more of a mess. \\
\hline

395 & You're doing it wrong\\
\hline

396 & Those hand dryers are only there to waste time until I just dry them on my pants anyway. Those things are useless\\
\hline

397 & Anyone have a study that compares new air dryers like the Dyson Airblade?\\
\hline

398 & I couldn't find a link to the scholarly work that inspired this article anywhere in the article. Do you have it OP? \\
\hline

399 & Frankly, I am not surprised.\\
\hline

400 & TIL the inside of hand dryers are moist. \\

\bottomrule
\end{tabular}
}
\caption{Processed textual dataset part 8}
\end{center}
\end{table*}

\begin{table*}
\scriptsize
\begin{center}
\resizebox{0.9\textwidth}{!}{%
\begin{tabular}{ l  l }
\toprule
\textbf{Class}       & \textbf{Sentences}   \\
\toprule
401  & Whatever happened to that cloth role that we used to use in elementary school? Isn\'t that more eco friendly? \\
\hline

402 & Sure, but probably the least good for preventing the spread of colds and stuff.\\
\hline

403 & TIL it is possible to have -155\% bacteria coverage over your hands.\\
\hline

404 & Dyson masterrace \\
\hline

405 & After you wash your hands isn't the bacteria count supposed to be pretty low anyways? So who cares if it\'s reduced by half or doubled?
\\
\hline

406 & I learnt this from Sheldon  \\
\hline

407 &How does the dyson airblade compare?\\
\hline

408 & Sheldon cooper warned us about this years ago\\
\hline

409 & Sounds interesting, but also something that I was probably better off not knowing\\
\hline

410 & I can confirm! In my AP bio class, we swabbed and cultured various locations in the bathroom. \\
& The one with the most bacteria and the most virulent (gram negative) was the inside of the hand dryer. Truly disgusting.'
\\
\hline

411 & So the Dyson Airblade is an overpriced bacteria spreader...fantastic\\
\hline

412 & My workplace is bad for this. The large set of toilets have got paper towel holders but never have paper towels in them and \\
& the single small toilet closer to where I work is filthy with no paper towels,no anti-bacterial soap and a really old hand drier. \\
& I tend to get sick a lot more when I'm working, I wonder if this is why. (I work 1 week on 1 week off) \\
\hline

413 & always wet You're using it wrong.  Go for the double-pull. \\
\hline

414 & Lol...the only time I was ever FORCED to use one was at a high school I was visiting (it was in a really bad area...not a good high school) and that's the only thing they had in the bathroom.  \\
& I would've likely been better off without washing at all, the entire towel was soaking wet.  A rather gross experience.  \\
\hline

415 & Yikes. I always double-pull those, and sometimes have to triple pull, but that takes a lot of patience because you have to wait for the click each time. \\
& Maybe someone lazy just swapped the empty clean roll with the full dirty one in your case. That should be a crime.\\
\hline

416 & The ones I'm speaking of were just a roll of towel...like linen towel.  A big loop of maybe 8-10 feet of towel that was in a constant state of soaked. \\
& Might work well for a non-busy bathroom but not for a busy one.  \\
\hline

417 & I don't think they're widely used anymore...this was probably 15 or so years ago that \\
& I saw these in use.  I'd sooner wipe my hands on my pants before use those again. \\
\hline

418 & So that Dyson guy lied about it working properly.\\
\hline

419 & Did they do this test in an existing bathroom with towels that have been collected floating butt germs? \\
\hline

420 & I don't know if they're that much butter but those Dyson AirBlade dryers have a Hepa Filter built \\
& in and are supposedly much more sanitary.  They even have them know built right into the faucet. \\
\hline

421 & Also, hand dryers suck compared to paper towels. \\
\hline

422 & i effin hate hand dryers. they drive me nuts\\
\hline

423 &What about normal towels? I can\'t stand this paper crap.\\
\hline

424 & Can we just get over the fact that keeping our hands bacteria free is incredibly difficult and not really worth much in the end (albeit a few careers). \\
\hline

425 & Freaking Tree Huggers tryna get us all sick.\\
\hline

426 & What about those 'air blade' dryers?\\
\hline

427 & Sheldon was right\\
\hline

428 & Is the bacteria  left from hand dryer pretty  much harmless  then?\\
\hline

429 & I always hated hand dryers. Dryers are for hair not hands. \\
\hline

430 & I saw this on big bang theory, I had no idea it was actually true...\\
\hline

431 & They never work either. I like to wash my face off every time I wash my hands and it's impossible to dry off with a hand dryer.\\
\hline

432 & Hey...Sheldon was right! \\
\hline

433 & Except for the Dyson ones. Nothing can live in those things cause they blow so hard\\
\hline

434 & I would think installing a simple UV light inside it would eliminate this.\\
\hline

435 & So Sheldon was right?\\
\hline

436 & The worst thing is on a road trip when every single rest stop has nothing but dryers.\\
\hline

437 & Well it makes sense. But the concern for me with most hand dryer blowers is that they are weak and take too long for most people. \\
& I think a lot of folks simply don't wash their hands when that's the only drying option.  \\
\hline

438 & What about the really cool ones in costco? \\
\hline

439 & Why would the inside of the dryer be moist?\\
\hline

440 & Portable size hand sanitizers ftw  \\
\hline

441 & Yeah but how bad are these germs? Do they make you sick? They seem to be everywhere who cares  \\
\hline

442 & After reading this thread I\'m electing to cut off my hands.\\
\hline

443 & I fucking love bacteria. \\
\hline

444 & Med student here. Don't be deceived - paper towels are far superior. \\
\hline

445 & Because you can turn off the tap without touching it and open door handles without touching it - just use the paper towel.\\
\hline

446 & Problem is waste. But in terms of cleanliness, single use paper towel continues to be the gold standard.\\
\hline

447 & Dyson Hand Dryer.\\
\hline

448 & I think Mythbusters showed that the air driers do a better job of spreading that bacteria all around in the air, so fuck other people, I guess. \\
\hline

449 & It's a even warmer, moister bacteria factory than OP's mom. \\
\hline

450 & Paper towels are largely bacteria free. \\

\bottomrule
\end{tabular}
}
\caption{Processed textual dataset part 9}
\end{center}
\end{table*}

\begin{table*}
\scriptsize
\begin{center}
\resizebox{0.9\textwidth}{!}{%
\begin{tabular}{ l  l }
\toprule
\textbf{Class}       & \textbf{Sentences}   \\
\toprule
451  & When I go to Costco I don't use those Dyson dryers any more because I know most people end up rubbing their hands on either side \\
& of the hand slots and most don't really wash their hands properly, so god knows how much bacterial crap is on those. \\
\hline

452 & Oh just dandy? ! So now we all have to carry around our own hand towels? !... sigh\\
\hline

453 & I think recurring use of the same hand towel will give bacteria a nice \\
& environment to grow as well, using the dispensable paper towel may be the most hygienic option.\\
\hline

454 & Please don't take this to seriously but er?...Go Fuck the Fuck off you little fucking Fuck faced fucker? ! \\
\hline

455 & Trying too hard.\\
\hline

456 & A hand dryer will **never** beat paper towels in effectiveness or speed. I'm not going to argue that. \\
\hline

457 & They are WAY cheaper to operate. Hand dryers cost about 0.02-0.18 cents per dry (in electricity) vs 1 cent per sheet for paper towels. So your hotel chain \\
& can go from spending \$1 million on paper towels a year down to (potentially) less than \$100K. That's huge.\\
\hline

458 & They're much better for the environment. It's estimated that if 1 out of 2 Americans were to stop using paper towels, \\
& we would save upwards of 25,000 trees **A DAY**. Given global warming concerns, etc, I think this is an important thing to take into consideration \\
\hline

459 & So, I'd argue if you value saving money for your business and saving the environment, a hand dryer would be the best choice.\\
\hline

460 & The paper industry is one of the greenest industries in operation today. \\
\hline

461 & Paper is highly recyclable and decomposes fast, and the trees we grow it with come from paper farms, not the rainforest or \\
& other precious sources of biodiversity. I don't think the environmental angle should be considered a significant factor in this argument. \\
\hline

462 & I do agree that from the business owners perspective, the paper towel is inferior in the cost-effectiveness.\\
\hline

463 & But still, as a hand drying device, it is superior to the automatic dryer. \\
\hline

464 & I'd argue that the only people really making decisions on hand dryers are going to be businesses/hotels, etc because this \\
& isn't something we generally will have in our homes. With that said, I would argue the superior choice for them in this aspect is definitely \\
& going to always be the hand dryer because it's cheaper to operate, and results in less waste in the basket, etc.\\
\hline

465 & So, with that said, when does cost effectiveness come into play with you? Are you only arguing from \\
& the user perspective and not the business perspective? If you're solely arguing from the former, than I really can't debate you. \\
\hline

466 & You're forgetting about large, non-profit organizations: hospitals and universities. \\
& While concerned with cost and environmental impact, these organizations put public health foremost.\\
\hline

467 & It's not just altruism or civic-mindedness: even minor, not remotely life-threatening \\
& outbreaks can be massively disruptive in health and/or education outcomes and labor costs.\\
\hline

468 & These organizations choose paper towels, because they are more effective at drying hands, more likely to be used, \\
& and do not concentrate, gestate, and spew forth fecal bacteria as hand dryers have been found to do.\\
\hline

469 & Do you have evidence that shows that hand dryers are a significant health risk? I understand they can harbor bacteria, but have there been studies where (for example) people \\
& using a hand dryers get sicker more often, etc, than people who use only paper towels? I'd like to see something like that if this is to be a convincing point.\\
\hline

470 & Not so fast haha. Unfortunately you forgot your own statement of the problem. You gave the condition "in public bathrooms" and you used \\
& things such as "towel let's me not touch the door on the way out" as justification for your argument that paper towel is better. \\
&Therefore per your argument we are taking into account everything that goes into drying of hands in a public bathroom environment.\\
\hline

471 & Cost/Environmental Impact(Due to High Foot Traffic)/Restocking/Availability all need to be \\
& considered as to how effective paper towels are versus hand dryers. Your paper towel has ZERO effectiveness if it runs out for example. \\
\hline

472 & When we take these things into consideration the hand dryer is obviously superior for all intents and purposes since it never runs out and is cheaper. \\
\hline

473 & If you want to modify your statement to, given a paper towel and a hand dryer the paper towel drys my hands faster and more effectively then yes you are probably correct.\\
\hline

474 & I'm pretty sure that's what my whole assertion was. It's in the last sentence of the OP \\
\hline

475 & That is incorrect. You're whole assertion definitely included more then just the last sentence and several other supporting points.  \\
& I can confirm by scrolling up the page! But now I am just being cheeky. \\
\hline

476 & There's a bathroom that I have to visit when we play baseball on the weekends.  Of course, the bathroom is nothing fancy. \\
& It's in a ballpark.  It's probably swept out once a year as far as cleaning goes.  But the absolute most disgusting thing in that bathroom \\
& is the spot on the painted concrete wall directly below the hand dryer that is literally caked in brown sludge.  Probably 3 square feet of \\
& the nasty that came off peoples hands *after* they washed them.  Papertowels are not an option here, so I wipe my hands on my clothes every time.  \\
\hline

477 & Also maintenance to pick up all the used towels. \\
\hline

478 & But if you value sanitation paper towels are *far* better.\\
\hline

479 & In fact, using a hand dryer will probably cause you to end up with more bacteria on your hands than were present before you washed your hands! \\
\hline

480 & There's a reason why in food service institutions, hand dryers are not allowed at employee handwash stations. \\
\hline

481 & I find this a valid argument, but do we have evidence showing that perhaps people who frequent a \\
& hand dryer location gets sick/harmed more often than those only exposed to paper towels?\\
\hline

482 & The study I linked in my previous post found that fecal bacteria were found deposited on the hands of people who used hand dryers (particularly hot air dryers, \\
& which are the worst offenders), and labeled the use of hand dryers as a public health threat as a result, as significant infections occur through the fecal-oral route.\\
\hline

483 & Where did you get the cent per sheet figure? \\
\hline

484 & Thank you for the source!\\
\hline

485 & 1. Go to any bulk ordering website (like alibaba) and you'll see that the unit price for paper towel rolls is \\
& about the same as the highest cost of one hand dry as you suggest. So paper towels win there.\\
\hline

486 & Each year Americans plant 1.6 billion trees. Going by your math, we would only be saving about 9 million trees a year. \\
& A difference of 5\% won't do much in the grand scheme when talking about global warming or the environment. So yeah, paper towels still reign supreme.\\
\hline

487 & I don't understand the argument here. You one time buy a hand dryer and it quickly makes up its costs over the course of 10-15 years, during which time you need \\
& to be constantly replenishing and buying paper towels and disposing of those paper towels with waste management services (that you need to pay for) \\
\hline

488 & The OP already stated the environmental argument isn't going to change his view, so anything to do with this angle is moot. \\
& But I agree it would be a relative drop in the bucket from a tree perspective.\\
\hline

489 & Where do reusable towels, as commonly found in continental Europe, fit in terms of financial and environmental cost?\\
\hline

490 & That's a good question. Would have to consider thought the cost of running washing machines to \\
& constantly clean the towels (water, electricity). I'm not sure what that comes out to.  \\
\hline

491 & Hands dryers blow water droplets around and aerosolize them. Not everyone washes thoroughly.\\
\hline

492 & It only takes ones person spreading shit-borne disease through the air to get a few people sick and wash out those economic and environmental savings.\\
\hline

493 & To know where the balance lies we would need to calculate the cost of the spread disease and the cost of the paper used.\\
& I bet the cost of the disease just from the doorknobs would be worse, with a hand dryer you can't use the paper towel to avoid touching the doorknob.\\
\hline

494 & Paper towels can run out, whereas hand dryers don't. \\
\hline

495 & If you use paper towels in a bathroom that isn't checked on often, it could be very possible that someone will not be able to dry their hands.\\
\hline

496 & I've been to bathrooms without paper towels, but never one with a broken hand dryer.\\
\hline

497 & That's a good point and while I do agree that this is a possible downside of the paper towel, in \\
& the event that both are available, the paper towel is always more effective. Edit: a word\\
\hline

498 & Nah his argument is terrible. Running out isn't a paper towel issue, it's a management issue. Paper towels are still superior in this regard.\\
\hline

499 & I have no real evidence available but you could look at germs associated with paper towels compared to hand blowers. So you may be dry but not clean? Just food for thought.\\
\hline

500 & The fact paper towels need to be 'managed' makes it a paper towel issue. You don't need someone 'managing' the hand blower, it's just there. \\

\bottomrule
\end{tabular}
}
\caption{Processed textual dataset part 10}
\end{center}
\end{table*}

\begin{table*}
\scriptsize
\begin{center}
\resizebox{0.9\textwidth}{!}{%
\begin{tabular}{ l  l }
\toprule
\textbf{Class}       & \textbf{Sentences}   \\
\toprule
501  & Of course you do need someone managing the hand blower. It's called repair when some idiot think they \\
& can put their weight on the hand blower and attempt to dry any other parts than their hands. (i.e. foot) \\
\hline

502 & Also you need routine maintenance for hand blowers.\\
\hline

503 & When a hand blower is down it's down for extended periods of time until a technician comes in. When paper towel runs out it takes 2 minutes to pull a new stack from the cupboard.'
\\
\hline

504 & Repairs does not equal management. Some idiot can rip the paper towel dispenser off the wall. The argument is symmetric, therefore moot to the point in question.\\
& Also hand dryers need maintenance a lot less frequently than paper towels need refilling. \\
\hline

505 & Repair does not equal management Who's job do you think it is then to call/schedule for a mechanic to repair a hand dryer then?
\\
\hline

506 & How often do you see an idiot ripping a paper power dispenser off the wall in comparison to people breaking hand dryers? A broken paper towel dispenser \\
& can still function as a manual paper towel dispenser unless it's purposely broken in a way the opening is sealed.  \\
\hline

507 & In that case you can still put the paper towels on the counter and let people grab it manually.\\
\hline

508 & Paper towels does not require any device for dispensing, but hand dryers absolutely  \\
& need a machine to deliver it's function, this is another way paper towels are superior to hand dryers!\\
\hline

509 & Point taken about hand dryers need a lot less maintenance than hand dryers; though the need for specific skills for hand dryer maintenance/repair means it \\
& has more downtime once it's down, whereas any idiot can open the cupboard and grab fresh paper towels themselves, so it's +1 for hand dryer and +1 for paper towels.\\
\hline

510 & I've seen plenty of towel dispenser ripped open or torn off the wall because they were empty but \\
& some idiot found it empty, but thought there was one towel left and wrenched it open.
\\
\hline

511 & I've never encountered a broken hand dryer.\\
\hline

512 & If we separate maintenance from management, then hand dryers need less maintenance, no management (aka refilling) and less often repairs.\\
\hline

513 & Ugh, you're just repeating your past points and haven't addressed any of my replies...hand dryers may have one or two points \\
& that are better but paper towels are still have far more advantages that outweighs a hand dryer. It's still not enough to change my \\
& view enough I want to have hand dryers replacing paper towels in my office...\\
\hline

514 & Which replies didn't I address. Your point seemed to be: * While towels need refilling, hand dryers need regular maintenance, \\
& to which I replied that you need to maintain a hand dryer a *lot* less than towels need refilling \\
\hline

515 & Hand dryers break down, to which I replied towel dispensers also get broken. I also added my \\
& anecdotal accounts of how I've seen loads of broken dispensers, but never encountered a broken dryer.\\
\hline

516 & But you said I didn't address your replies. This response instead implies I simply address your replies multiple times  \\
\hline

517 & Repeating someone's point isn't addressing it.  Unless you say you're conceding and affirming my points, well then, fine, I'll take that. \\
\hline

518 & 1. Hand dryers do not need refilling, whereas towels do\\
\hline

519 & Hand dryers do require maintenance, which could be considered equivilent to refilling, but is much less frequent \\
\hline

520 & In my own personal experience, hand dryers are broken much less often that towel dispensers. \\
\hline

521 & Both do need repairs, but hand dryers, it appears to me, less often. That's my stance.\\
\hline

522 & Concession to this, but the refilling process is such a simplistic task it doesn't require a \\
& technician to do its advantage outweighs the lack of refill/maintenance a hand dryer has.\\
\hline

523 & Also concession to this, but the same point about refilling being a task so simplistic that it still outweights the advantage of a hand dryer.\\
\hline

524 & My personal experience is opposite to this, but the ratio should not matter as a paper towels \\
& really do not require a dispenser to function, which makes it a plus 1 for paper towels\\
\hline

525 & Thanks, it's been fun. Sorry for being a little frustrating in the middle.\\
\hline

526 & We got better so it's a net good in the end :)\\
\hline

527 & Dryers can break down and take longer to be fixed than it does for management to replace a roll of paper towels.\\
\hline

528 & Those dyson hand air blade things are actually very dirty and need to be cleaned on a daily basis. If management doesn't clean them, \\
& fuck even just waving your hands in the air till they dry is a better option lol\\
\hline

529 & At this point we'd need data on how much time hand dryers spend out of actions vs how much total time towel dispensers spend empty. But I'd assume that \\
& an empty dispenser probably gets 'ignored' for longer than a broken dryer (which would likely be raised immediately and happens less often) \\
\hline

530 & As for the air blades, I assume you're talking about the version 1 with the U-shaped design? Sure, that's a design flaw, which is why the version 2 do away with that.\\
\hline

531 & Yeah, but I have yet to see any places use those. The version 1's haven't broke down yet lol\\
\hline

532 & lot of places around me use the ver2, having replaced non-airblades. Also, that would then support low break down rate of air dryers\\
\hline

533 & Also, it probably should be noted that cleaning down hand dryers would be part of regular bathroom cleaning *anyway,* so wouldn't \\
& be exceptional maintenance specific to the device. Whereas refiling paper towels would be exceptional activity.\\
\hline

534 & Yes, but while a management issue can screw up an air dryer, it has a far smaller chance of doing so than with paper towels.\\
\hline

535 & If he presented you with a new perspective that made you concede your original view even a little bit, you should award him a delta.\\
\hline

536 & The bathroom running out of towels is meaningless to the discussion. You can't use the absence of A when comparing A to B, \\
& because there is nothing for B to be compared to. OP is contending that A is superior to B, the absence of A doesn't come into the equation. \\
\\
\hline

537 & Paper towels are superior to dryers REGARDLESS if they run out because in terms of drying capabilities towels are in fact the better option \\
\hline

538 & Furthermore there have been occasions where the dryer doesn't work and I've had to walk to another bathroom to dry my hands.\\
\hline

539 & The possibility of running out is an inherent characteristic of paper towels, which is negative. Air dryers don't have this characteristic, and are therefore better in this regard.\\
\hline

540 & If the speed of dispensing the "drying material" is relevant, even though it's a property of the plastic dispenser, then this argument is relevant too.  \\
\hline

541 & Because running out of something is an inherent characteristic of **everything** because we exist in a universe with entropy. \\
&It is irrelevant to refer to the absence of a thing as though that in some way says something negative about it. \\
&If I say I prefer lemonade over cola and then someone tells me that I might run out of lemonade, that doesn't change the fact that I still prefer lemonade. \\
&Of course towels get used up, that's what happens when you use disposable paper towels. If you're going to bring up the possibility of running out of paper towels, \\
&you ought to bring up the possibility of the hand dryer being broken. If X occurs between two objects (A and B) but you only use X in the instance \\
& of A and not B to make B look better than A, you're clearly not being fair.   \\
\hline

542 & Towels need to be replaced at least once a day (maybe upwards of ten if you're in a busy place).\\
\hline

543 & The dryer could break, but with existing technology, it's unlikely that it will require maintenance several times a day.\\
\hline

544 & The towel dispenser is guaranteed to require this, and is therefore more likely to be inconvenient, no? \\
\hline

545 & Yeah but context is everything. What happens more often? Towels running out or the dryer being broken? \\
& Both are certainly possible, but this particular question has to do with probability.\\
\hline

546 & The average user of a public bathroom would say towels are vastly better at drying hands than a dryer is.\\
\hline

547 & The guy who gets paid to stock the towel dispenser, and the guy who pays for the towels, the other man's salary and \\
& the electricity bill of the building would argue that dryers are less costly and need less maintenance therefore are better than towels\\
\hline

548 & If you can't take the *just slightly* larger picture into account, should we consider you to have any meaningful input on the issue? \\
& One would hope that administrators are taking effectiveness into account along with cost and employee morale, \\
& and that the end user would have some inclination to avoid waste and some consideration for staff."
 \\
\hline

549 & removed \\
\hline

550 & 1. They're quicker by a very small margin. You're arguing over seconds, and it's not like these are seconds right at your job or whatever.  If you leave a desk \\
& to use the bathroom, your hands will dry by the time you need them again. You're really arguing over a few seconds more of having dry hands for little to nothing \\

\bottomrule
\end{tabular}
}
\caption{Processed textual dataset part 11}
\end{center}
\end{table*}

\begin{table*}
\scriptsize
\begin{center}
\resizebox{0.9\textwidth}{!}{%
\begin{tabular}{ l  l }
\toprule
\textbf{Class}       & \textbf{Sentences}   \\
\toprule
551  & Effectiveness is measured by time really, and that's part of 1. Dry is dry though. It's not like your hands are going to retain water like a towel itself. \\
\hline

552 & Why are you cleaning sinks like that? And, after using the auto dryer, I still have to touch the door to exit.\\ 
&This is a commonly held but bad belief. The door isn't germ city any more than the rest of the bathroom. And when you do make it out of the bathroom \\
& by doing that, you still have to touch absolutely everything else that people have also touched with the same hands.\\
\hline

553 & Also, if you used an air dryer, you hosed your hands down with the combined fecal bacteria of everyone who used that bathroom before you, so I really wouldn't worry about the door. \\
\hline

554 & Removed \\
\hline

555 & And in all scenarios dryers are extremely loud, which scares my kids.\\
\hline

556 & hey often jump if someone else engages the dryer and when they were younger they refused to use them because of noise.\\
\hline

557 & Also a big + for flexibility of the towel - a need to wipe dry my kids face after they indulged on an ice cream \\
& and I cleaned the face with water is quite common. It is very hard to do it with a hand dryer. \\
\hline

558 & I feel because of this all businesses that cater to children (like malls for example) should at least offer an option of a towel. But sadly a lot of them don't.\\
\hline

559 & removed\\
\hline

560 & paper towels are not a poop cannon \\
\hline

561 & no way. i could glide my hands through those for longer than a minute and my hands are still at least kind of moist, they never *completely* dry the hand \\
\hline

562 & The paper towels dry it all pretty much instantly\\
\hline

563 & OP missed the main strength of paper towels: they are the most hygienic option. \\
\hline

564 & Air dryers breed and expel fecal bacteria.\\
\hline

565 & I will disagree with you on quickness. I've used the air blade dryer and while it's a cool piece of technology, \\
& I don't believe my hands have become sufficiently dry after just two passes. Basically all of the water accumulates at the ends of my fingers. \\
\hline

566 & Plus, in thinking about my experience using paper towels in the place I use them most (my office restroom), I tear the towel off the dispenser \\
& and am immediately turning around and walking to the exit while simultaneously drying my hands. By the time I get to the exit where the trash can is, my hands are dry. \\
\hline

567 & Yes, I like paper towels more, but if you run a business and people keep throwing paper towels in your toilet, you will continually have plumbing issues to deal with. \\
\hline

568 & How many signs do you see in restrooms reminding people to throw paper towels in the trash, not the toilet? \\
& These business owners have had to call the plumber on occasion. That's why automatic hand dryers are better for business owners.\\
\hline

569 & Time for a restroom remodel. Even a simple wall divider between sink and toilet will radically \\
& reduce that behavior, and you won't have an e.coli breeding facility bolted to your wall. \\
\hline

570 & One major problem with paper towels is that you have to empty the paper basket *all the time*.\\
\hline

571 & I don't think the door argument holds either, since you 1) Will have to touch at least one more door to get out of the building in most cases,\\
& and 2) Can use a part of your sleeve, shirt, or basically any piece of clothing to open the door. \\
& Also, in some cases you can use you elbow to push the handle, and push you back against the door. \\
\hline

572 & If you haven't encountered one of the more modern hand dryers then paper towels are definitely superior but they do make them *powerful* now. \\
\hline

573 & Why THE FUCK would you choose paper over air?! Does no one care about the environment?\\
\hline

574 & I know this argument isn't about the environment but if nothing else makes you choose a hand dryer over paper towels, the environment SHOULD. \\
\hline

575 & As culb77 already pointed out, air dryers were a bad idea from the start. They are poop cities complete with an airport with direct flights to your hands.\\
\hline

576 & deleted \\
\hline

577 & Paper towels are much *more* hygienic. \\
\hline

578 & When you rub your hands with them, they physically remove bacteria and other contaminants from your hands.\\
\hline

579 & In contrast, bacteria love reproducing inside of hand drying machines because of the moisture, warm air, and darkness.\\
\hline

580 & Well then, I stand corrected. Thanks \\
\hline

581 & Environmentally friendly would be the hand dryer, even though it's using electricity that might have been produced by coal.\\
\hline

582 & The paper towel by using them your using: The tree that was cut to make the paper\\
& The factory, machines and chamicals used to chop, clean, pulp the paper and then process it into the towels.\\
&The fuel used to transport them from t he place they where bought to the place they are going to be used\\
\hline

583 & The dryer you are using: The ore that was mined from the ground, the trucks, drills, machinery that was used to dig them out, transport them to a refinery.\\
\hline

584 & Just as an addendum, having worked in multiple places that use paper towels, that shit gets thrown into a landfill, no recycling about it. \\
& Then consumers come along, use it a thousand times, maybe before it is discarded\\
\hline

585 & Anyway, most people undo their hand washing as soon as they open the door when they hold the handle \\
& that everyone else has held, along with the people who didn't clean their hands.\\
\hline

586 & Electricity is being produced all the time, for everything, Paper towels are being produced for a specific use and reason. \\
& Take away paper towels, you break a huge chain, take away hand dryers, the use of everything is still there.\\
\hline

587 & Because hand dryers just materialize out of midair and have no chain associated with their manufacture.\\ 
& None at all. Glad to see there's such critical thought put into your response.\\
\hline

588 & Just in case you didn't know, and regardless of what turns out to be more environmentally friendly, \\
& there's another thing you can do that makes a big difference: vigorously flick water off your hands first.\\
\hline

589 & I always take a handkerchief with me, so I usually use that. If that is dirty I will use paper towels. If paper towels aren't available I use my undershirt.\\
\hline

590 & I was at Mc Donalds once in the bathroom and washed my hands and wiped them on my shirt. This woman looked at me with disgust \\
& and I said 'those hand dryers blow so many germs on your hands you might as well just rinse them in the toilet before flushing'.  \\
\hline

591 & She went 'oh, really?' and I said, yeah... that's why I don't use the blowers. She washed her hands and wiped them on her shirt. Lol.\\
\hline

592 & Yeah, that huge outbreak of disease due to hand dryers...\\
\hline

593 & The risk of the dryer comes from the fact it circulates the water droplets off people's hands. Meaning, just walking past it, or being near it, puts you "at risk".\\
\hline

594 & And that's how you get bladder infections. There's being careful, and then there's being too careful. \\
& Also, since having a kid, if I don't go pee the moment I feel that urge, I'm at serious risk of flat out pissing myself. \\
\hline

595 & Also, since having a kid, if I don't go pee the moment I feel that urge, I'm at serious risk of flat out pissing myself.  \\
& Hahahaha! Oh geez! That's gotta subside, right? Are there exercises you can do or something? \\
\hline

596 & It's been seven years. It's not going away. I do kegals but it does fuck all. When I have to pee I have to pee now.\\
\hline

597 & THIS!!!!\\
\hline

598 & I would expect a hand drier. Especially depending on where the energy of it comes from\\
\hline

599 & Considering the paper towels already exist and will be used by someone eventually even if you didn't use them, \\
& it proably dosnt matter but if you made your own paper towels each time you needed to dry your hands then a hand dryer would be more efficient.\\
\bottomrule
\end{tabular}
}
\caption{Processed textual dataset part 12}
\end{center}
\end{table*}

\section{More Examples from PAPYER}

To demonstrate more details of our dataset, we pick 1 example from each super category.

\begin{center}
\includegraphics[width=\columnwidth]{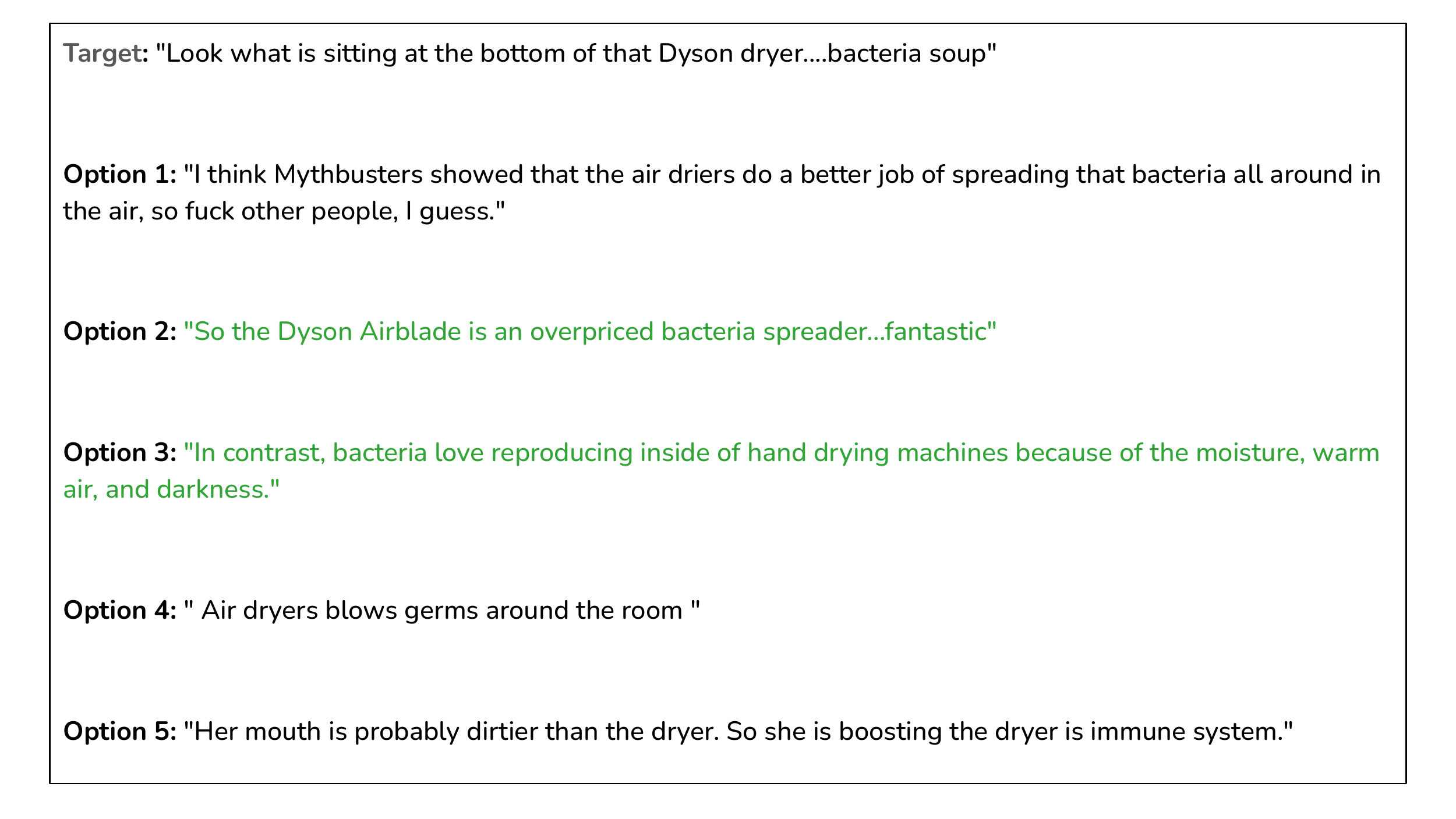}
\captionof{figure}{Example with super category: \textit{For paper towels}}
\end{center}

\begin{center}
\includegraphics[width=\columnwidth]{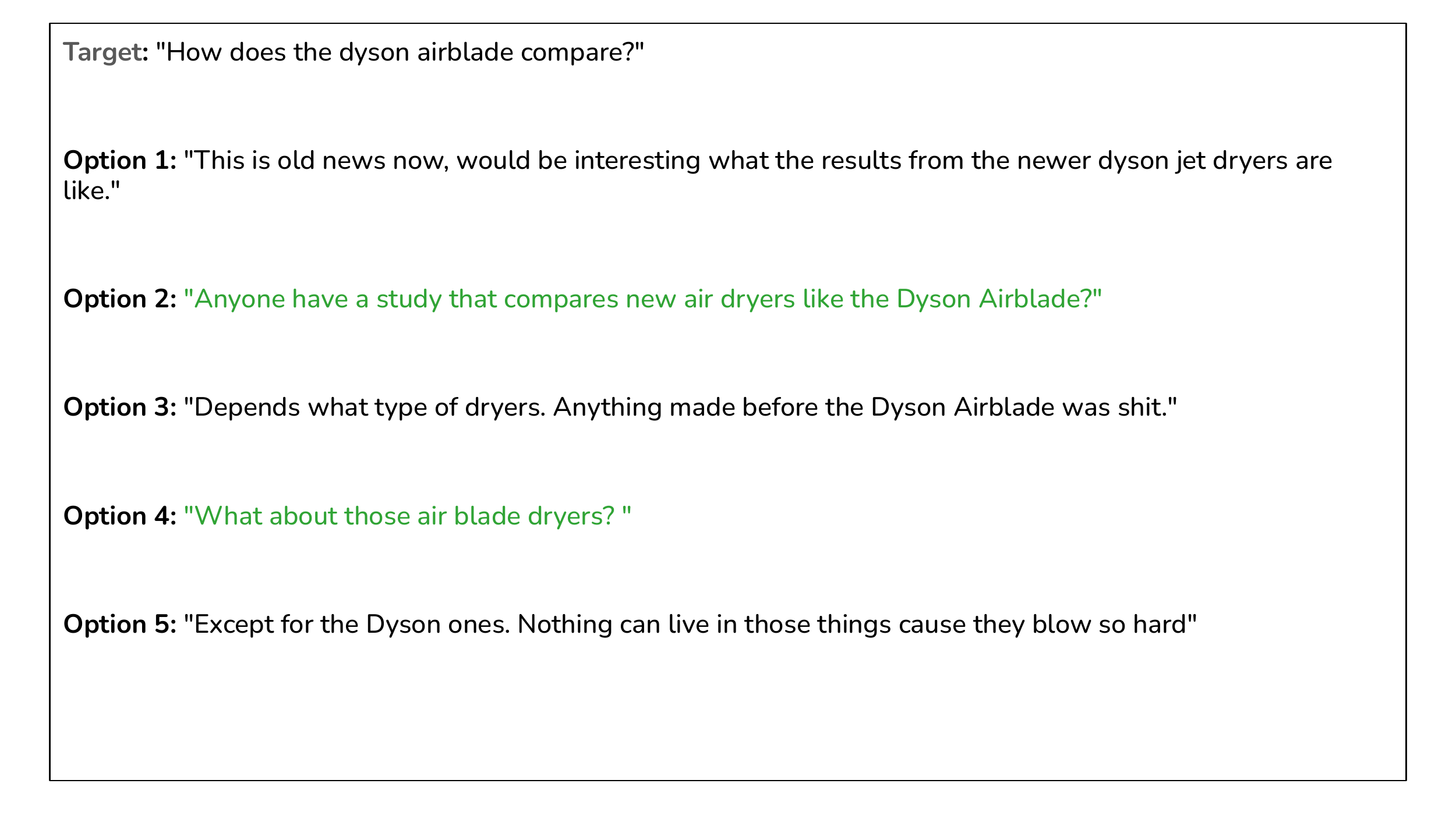}
\captionof{figure}{Example with super category: \textit{For hand dryer}}
\end{center}

\begin{center}
\includegraphics[width=\columnwidth]{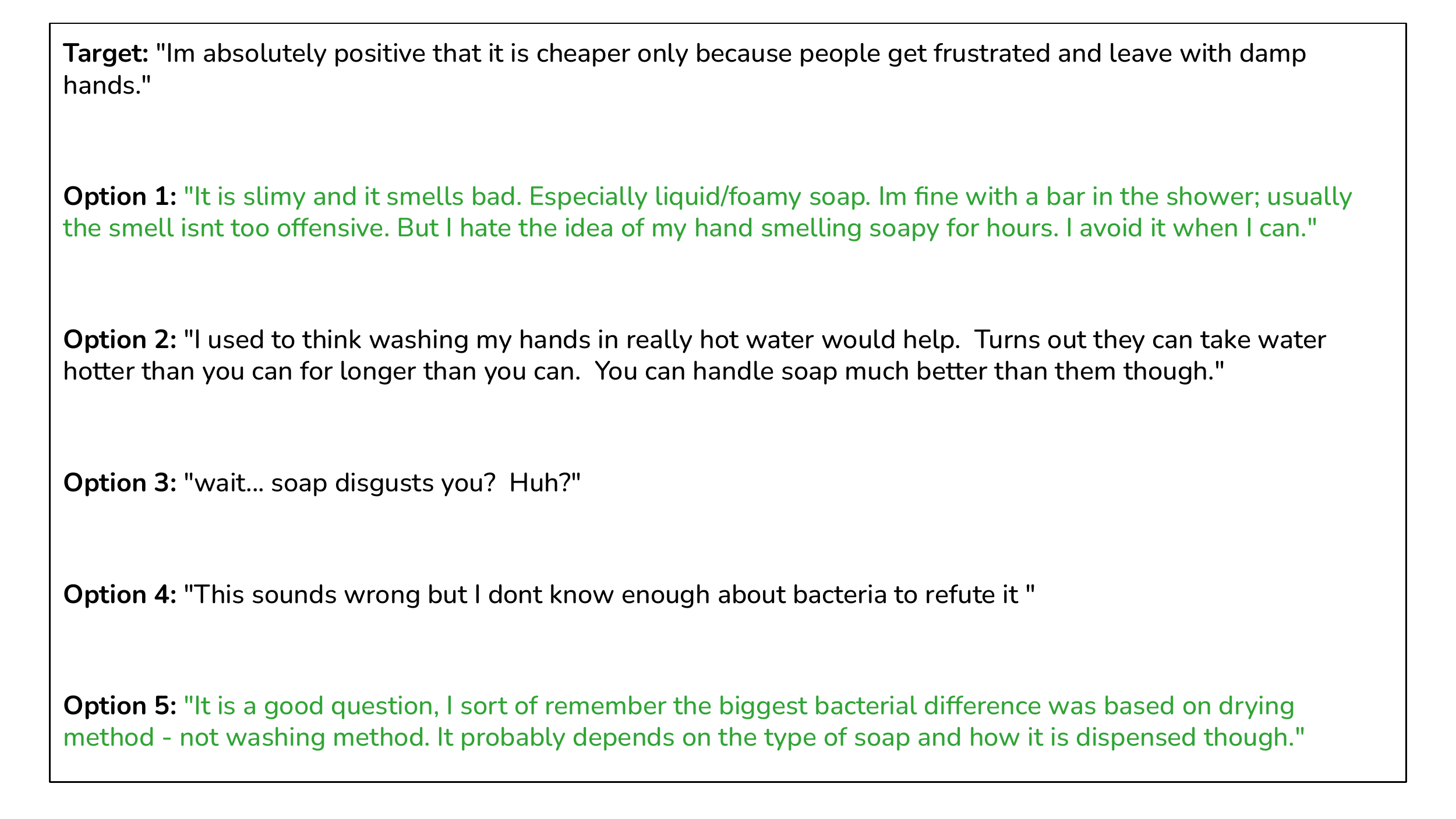}
\captionof{figure}{Example with super category: \textit{Other}}
\end{center}

\end{document}